\documentclass[preprint, a4paper]{elsarticle}
\usepackage[ansinew]{inputenc}
\usepackage{graphicx}
\usepackage{latexsym}
\usepackage{varwidth}
\usepackage{color}
\usepackage{dsfont}

\usepackage{amsmath,amsfonts,amssymb}
\usepackage{stmaryrd}

\newcommand\Omit[1]{}

\def\qed{\hspace{\fill}\rule[-.2ex]{.6ex}{1.2ex}}

\long\def\COMMENT#1\ENDCOMMENT{\message{(Commented text...)}\par}
\renewcommand{\mod}[1]{\llbracket #1 \rrbracket}
\newcommand{\Spain}{\mbox{\it Spain}}
\newcommand{\myref}[1]{(\ref{#1})}

\newcommand{\mymod}[1]{\llbracket {#1} \rrbracket}
\newcommand{\expf}{\rhd_f}
\newcommand{\Set}[1]{\{{#1}\}}
\newcommand{\ie}{{\em i.e.,\ }}

\def\expl#1#2{{#2}\rhd{#1}}
\newcommand{\Elc}[2]{{#2}\rhd^{\ell c}{#1}}
\newcommand{\Elne}[2]{{#2}\rhd^{\ell ne}{#1}}
\newcommand{\Elneu}[2]{{#2}\rhd_1^{\ell ne}{#1}}
\newcommand{\Elned}[2]{{#2}\rhd_2^{\ell ne}{#1}}
\newcommand{\nElne}[2]{{#2}\not\!\!\rhd^{\ell ne}{#1}}

\newcommand{\nElned}[2]{{#2}\not\!\!\rhd_2^{\ell ne}{#1}}
\newcommand{\iffdef}{\quad\!\!\stackrel{def}{\Leftrightarrow}\!\!\quad}

\newcommand{\LLE}{\mbox{\sf LLE}}
\newcommand{\LLEs}{\mbox{\sf LLE$_{\scriptscriptstyle\Sigma}$}}

\def\nms{\vdash_{\scriptscriptstyle{\Sigma}}}
\def\RLE{\mbox{\sf RLE}}
\def\RLEs{\mbox{\sf RLE$_{\scriptscriptstyle\Sigma}$}}

\def\Econ{\mbox{\sf E-Con$\boldmath {}_{\scriptscriptstyle\Sigma}$}}
\def\ECM{\mbox{\sf E-CM}}
\def\EWCM{\mbox{\sf E-W-CM}}
\def\EDR{\mbox{\sf E-DR}}

\def\RA{\mbox{\sf RS}}
\def\ECC{\mbox{\sf E-C-Cut}}
\def\EWCC{\mbox{\sf E-W-C-Cut}}
\def\ERC{\mbox{\sf E-R-Cut}}

\def\ERW{\mbox{\sf ROR}}

\def\LOR{\mbox{\sf LOR}}
\def\ERef{\mbox{\sf E-Reflexivity}}

\newtheorem{defi}{Definition}
\newtheorem{theo}{Theorem}
\newtheorem{proposition}{Proposition}

\newtheorem{example}{Example}

\def\qed{\hfill {\scriptsize{$\blacksquare$}}\break}
\def\beginproof{\noindent {\it Proof:~}}
\def\endproof{\hfill \qed}
\def\iff{if and only if }

\def\p{{\sf P}}
\def\np{{\sf NP}}
\def\npc{{\sf NP}-complete}

\def\conpc{{\sf coNP}-complete}

\def\sat{{\sc sat}}
\def\unsat{{\sc unsat}}
\def\bhdeux{$\mbox{{\sf BH}}_2$ }
\newcommand{\mypost}[3]{\noindent {\bf (#1)}\hspace{2mm}{#2}\hfill{\bf (#3)}\\ }
\newcommand{\mypostitem}[3]{\noindent { #1}\hspace{2mm}{#2}\hfill{\bf (#3)}\\ }
\newcommand{\mypostbis}[2]{\noindent {\bf (#1)}\hspace{2mm}{#2}\\ }

\newcommand{\myPhi}{{\it \Phi}}

\newcommand{\E}{\myPhi} % un profil
\newcommand{\IC}{\mu}
\newcommand{\arb}{\ensuremath{\Delta}}
\newcommand{\arbic}{\ensuremath{\Delta_{\IC}}}
\newcommand{\fequiv}{\equiv}
\newcommand{\K}{\varphi}
\newcommand{\I}{\omega}
\newcommand{\J}{\omega'}

\newcommand{\rit}{\mathds{R}}

\begin{document}

\begin{frontmatter}

\author{Isabelle Bloch$^1$, J\'er\^ome Lang$^2$, Ram\'on {Pino P\'erez}$^3$, Carlos Uzc\'ategui$^{4}$}
\title{Morphologic for knowledge dynamics:\\ revision, fusion, abduction}
%Morphologic for knowledge representation and reasoning: revision, fusion, abduction}
\address{(1) LTCI, T\'el\'ecom ParisTech, Universit\'e Paris-Saclay, Paris, France - isabelle.bloch@telecom-paristech.fr\\
(2) Universit\'e Paris-Dauphine, PSL Research University, CNRS, UMR 7243, LAMSADE, 75016 Paris, France - lang@lamsade.dauphine.fr\\
(3) Departamento de Matem\'aticas, Universidad de Los Andes - Merida, Venezuela - pino@ula.ve \\
(4) Escuela de Matem\'aticas, Facultad de Ciencias, Universidad Industrial de
Santander - Bucaramanga, Colombia - cuzcatea@saber.uis.edu.co}

\begin{abstract}
Several tasks in artificial intelligence require to be able to find models about
knowledge dynamics. They include belief revision, fusion and belief merging, and abduction. In this paper we exploit the algebraic framework of mathematical morphology in the context of propositional logic, and define operations such as dilation or erosion of a set of formulas. We derive concrete operators, based on a semantic approach, that have an intuitive interpretation and that are formally well behaved, to perform revision, fusion and abduction. Computation and tractability are addressed, and simple examples illustrate the typical results that can be obtained.
%We show how to associate tools of Mathematical Morphology with Logic. This gives rise to a rich area that we call Morphologic. We show that this general and appealing framework can be used successfully  in many different classes of tasks in Knowledge Representation, namely Belief Revision, Information Fusion, Abduction, Preference Representation and others.
%In particular, we introduce the computation techniques of the morphological operators on which our applications are based. We also show that our morphological operators have a good rational behavior with respect to reasoning purposes.
\end{abstract}

\end{frontmatter}

%%%%%%%%%%%%%%%%%%%%%%%%%%%%%%%%%%%%%%%%%%%%%%%%%%%%%

%\input{intro-essai}
\section{Introduction}

Several tasks in artificial intelligence require to be able to find models about
knowledge dynamics. In particular, how do beliefs change in the light of a new observation,
how can we extract a coherent source of information of many sources of information (eventually contradictory),
or how can a given observation be explained? All these questions fall more precisely under the following topics:
  belief revision, belief merging or fusion, and abduction, respectively.

%\fbox{une phrase pour definir chacun ?}

Such tasks have been formalized and axiomatized in various logics.
It is out of the scope of this paper to review the huge amount of work done in this direction, and we will rely on existing postulates, now rather widely accepted, such as AGM postulates for revision~\cite{KatsunoMendelzon91}, integrity constraints postulates for merging and fusion~\cite{KONI-98,KONI-02,KP11}, rationality postulates for abduction and explanatory relations~\cite{PINO-99,PU03}.

Here the propositional logic is considered, and propositional formulas are used to encode either pieces of knowledge (which may be generic, for instance
integrity constraints, or factual such as observations) or ``preference items'' (such as beliefs, opinions,
desires or goals). Such formulas are then used for complex reasoning or decision making tasks.

In this paper, we propose to build tools for modeling knowledge dynamics based on mathematical
morphology operators applied to propositional formulas.
Mathematical morphology is originally based on set theory. It has been
introduced in 1964 by Matheron \cite{MATH-67,MATH-75}, in order to study porous media. But
this theory evolved rapidly to a general theory of shape and its transformations,
and was applied in particular in image processing and pattern recognition \cite{SERR-82}.
Additionally to its set theoretical foundations, it also relies on topology on sets,
on random sets, on topological algebra, on integral geometry, on lattice theory.
In particular, the general algebraic framework of lattices allows developing mathematical morphology in various domains of information processing, beyond sets and functions, such as fuzzy sets, logics, graphs, hypergraphs, formal concept analysis, etc.~\cite{IB:INS-11,IB:CVIU-13,IB:LOS-07,TCIS-02,Rons90}.

The aim of this paper is to develop mathematical morphology in propositional logics, called morphologic, and to propose concrete morphological operators to perform revision, fusion and abduction, which are tractable and have an intuitive meaning. In particular we will make use of two important operations, dilations and erosions. Intuitively, when applied to a set, the effect of dilation is to expand the set while the effect of erosion is to shrink the set.

The following ideas explain intuitively why  morphologic is an adequate tool for knowledge dynamics:
\begin{itemize}
\item {\em Belief revision}: let $\varphi$ and $\psi$ be two
propositional formulas. The models of the revision $\varphi \circ \psi$
of $\varphi$ by $\psi$ are the models of $\psi$ which are closest
(with respect to a given proximity notion) to a model of $\varphi$. Intuitively, using the language of morphologic, it means that $\varphi$ has to be dilated enough to become consistent with
$\psi$.
\item {\em Belief merging}: finding the best
compromise between a finite set of formulas $\varphi_1$, ... $\varphi_n$
amounts to selecting the models which minimize the aggregation
(using some given operator) of the distances to each of the $\varphi_i$.
This amounts intuitively to dilate simultaneously all the $\varphi_i$
until they constitute a consistent set.
\item {\em Abductive reasoning}: preferred explanations of a formula are defined based on a set of axioms, several of which being closed to properties of morphological operators, in particular erosion.
\end{itemize}

An important noticeable aspect is that the framework of morphologic gives us not only natural and general notions
to deal with many tasks of knowledge dynamics, but this approach is also well behaved. Actually, the operators and relations obtained via the morphological tools enjoy good rationality properties.
Moreover, last but not least, under certain assumptions there are interesting ways of computing some of our proposed operators.

The main contribution of this work is to propose such models in the framework of morphologic, based on a semantic approach. One interesting aspect is that the proposed operators include some of existing ones, and also new ones. For each of them, the properties will be analyzed and discussed. Finally, the outcome is a toolbox of operational methods, among which a user can choose according to the required properties.

This paper is organized as follows:
 Section \ref{sec:MM} is devoted to the presentation of concepts in mathematical morphology and to introduce logical morphology (morphologic).
 Section \ref{computations} shows the general techniques of computation of the operators when the metric over the space of valuations
 %($\mathbf{2}^N$)
 is given by the Hamming distance. Section \ref{sec:Revision} is  devoted to show how well-known revision operators can be interpreted in the framework of morphologic.
   Section \ref{sec:Fusion} proposes a similar analysis
   %makes a similar work of the previous section but
   in the framework of fusion. It shows how belief merging operators can be interpreted in the framework of morphologic. Section \ref{sec:abduction} is devoted to abduction (explanatory relations) built on morphological operations aiming to capture the notion of the {\em most central part}.
Based on a common notion of pre-order relation on models, derived from morphological operators, Section \ref{sec:unify} presents a unified framework for revision and abduction. In Section \ref{sec:conclu} we finish with some concluding remarks and perspectives for future work.

%%%%%%%%%%%%%%%%%%%%%%%%%%%%%%%%%%%%
%%%%%%%%%%%%%%%%%%%%%%%%%%%%%%%%%%%%%%

%\input{mm}

\section{From mathematical morphology to logical morphology}
\label{sec:MM}

\Omit{
\fbox{TO DO IN THIS SECTION:}
\begin{itemize}
\item add a subsection (or a comment somewhere) discussing the role of the structuring element (a way to impose preferences, a parameter to be tuned for each application...). Partially done in Section 2.6.
\end{itemize}
}

In this section we recall the main concepts and tools used in mathematical morphology and their interpretation in mathematical logic. This interpretation is possible via the identification between a logical formula and a set of interpretations (its models) in the framework of finite propositional logic.

\subsection{Algebraic framework: complete lattices}

Mathematical morphology relies on concepts and tools from various
branches of mathematics: algebra (lattice theory), topology, discrete
geometry, integral geometry, geometrical probability, partial
differential equations, etc.~\cite{MATH-75,SERR-82}; in fact any mathematical theory that
deals with shapes, their combinations or their evolution, can be
brought to contribute to morphological theory.
When adopting a logics point of view, the algebraic framework is particularly relevant, and we will concentrate on it in the sequel.

The basic structure in this framework is a complete lattice $(L, \leq)$\footnote{Although mathematical morphology has also been extended to complete semi-lattices and general posets~\cite{keshet2000}, based on the notion of adjunction, in this paper we only consider the case of complete lattices.}.
We denote the supremum by $\bigvee$, the infimum by $\bigwedge$, the smallest element by $0_L$ and the greatest element by $1_L$. We have $0_L = \bigwedge L = \bigvee \emptyset$ and $1_L = \bigvee L = \bigwedge \emptyset$. The framework of complete lattices is fundamental in mathematical morphology, as explained in~\cite{HEIJ-90,RonsHeij:91,Rons90}.

All the following definitions and results are detailed in textbooks on mathematical morphology, such as~\cite{Heijmans94,NajTal10,SERR-88}. We restrict the presentation to operators from $(L, \leq)$ into itself.

An algebraic dilation is defined as an operator $\delta$ on $L$ that commutes with the supremum, and an algebraic erosion as an operator $\varepsilon$ that commutes with the infimum, i.e. for every family $(x_i)_{i \in I}$ of elements of $L$ (finite or not), where $I$ is an index set, we have:
\begin{equation}
\delta(\bigvee_{i \in I} x_i) = \bigvee_{i \in I} \delta(x_i),
\label{eq:defDil}
\end{equation}
\begin{equation}
\varepsilon(\bigwedge_{i \in I} x_i) = \bigwedge_{i \in I} \varepsilon(x_i).
\label{eq:defEro}
\end{equation}
These are the two main operators, from which a lot of others can be built.

Among the numerous examples of complete lattices, one will be particularly interesting for the extension to logics:
$({\cal P}(E), \subseteq)$, the set of subsets of a set $E$, endowed with the set theoretical inclusion. It is a Boolean lattice (i.e. complemented and distributive). The smallest and greatest elements are $0_L = \emptyset$ and $1_L = E$, respectively.

Algebraic dilations and erosions in $(L, \leq)$ satisfy the following properties:
\begin{itemize}
\item $\delta(0_L) = 0_L$ and $\varepsilon(1_L) = 1_L$,
\item $\delta$ and $\varepsilon$ are increasing with respect to the partial ordering on $L$,
\item in $({\cal P}(E), \subseteq)$,
$\delta(X) = \cup_{x \in X} \delta(\{ x \} )$.
\end{itemize}

Another important concept is the one of adjunction.
A pair of operators $(\varepsilon, \delta)$ defines an adjunction on $(L, \leq)$ if:
\begin{equation}
\forall (x,y) \in L^2, \; \delta(x) \leq y \Leftrightarrow x \leq \varepsilon(y).
\end{equation}

If a pair of operators $(\varepsilon, \delta)$ defines an adjunction, the following important properties hold:
\begin{itemize}
\item $\delta(0_L) = 0_L$ and $\varepsilon(1_L) = 1_L$,
\item $\delta$ is a dilation and $\varepsilon$ is an erosion (in the algebraic sense expressed by Equations~\ref{eq:defDil} and~\ref{eq:defEro});
\item $\delta\varepsilon \leq Id$, where $Id$ denotes the identity mapping
on $L$ (i.e. $\delta\varepsilon$ is anti-extensive);
\item $Id \leq \varepsilon\delta$ (i.e. $\varepsilon\delta$ is extensive);
\item $\delta \varepsilon \delta \varepsilon = \delta \varepsilon$ and
$\varepsilon \delta \varepsilon \delta = \varepsilon \delta$, i.e. the
composition of a dilation and an erosion are idempotent operators ($\delta \varepsilon$ is called a morphological opening and $\varepsilon \delta$ a morphological closing).
\end{itemize}

The following representation theorem holds: an increasing operator $\delta$ is an algebraic dilation iff there is an operator $\varepsilon$ such that $(\varepsilon, \delta)$ is an adjunction; the operator $\varepsilon$ is then an algebraic erosion and $\varepsilon(x) = \bigvee \{ y \in L, \; \delta(y) \leq x \} $. Similarly, an increasing operator $\varepsilon$ is an algebraic erosion iff there is an operator $\delta$ such that $(\varepsilon, \delta)$ is an adjunction; the operator $\delta$ is then an algebraic dilation and $\delta(x) = \bigwedge \{ y \in L, \; \varepsilon(y) \geq x \}$.

Finally, let $\delta$ and $\varepsilon$ be two increasing operators such that $\delta\varepsilon$ is anti-extensive and $\varepsilon\delta$ is extensive. Then $(\varepsilon, \delta)$ is an adjunction.

Further properties and derived operators can be found in seminal works such as~\cite{Heijmans94,SERR-82,SERR-88}, or in more recent ones~\cite{IB:LOS-07,NajTal10}.

In this paper, the fact that dilations and erosions are increasing operators that commute with the supremum and the infimum, respectively, will play an important role.

\subsection{Structuring element and morphological dilations and erosions}

Let us now consider the lattice $({\cal P}(E), \subseteq)$ of the subsets of $E$. We have
$\delta(X) = \cup_{x \in X} \delta(\{ x \} )$. If $E$ is a vectorial or metric space (e.g. $\mathbb{R}^n$), and if $\delta$ and $\varepsilon$ are additionally supposed to be invariant under translation, then it can be proved that there exists a subset $B$, called {\em structuring element}, such that
\begin{equation}
\delta(X) = \{ x \in E \mid  \check{B}_x \cap X \neq \emptyset \}
\label{eq:setDilation}
\end{equation}
and
\begin{equation}
\varepsilon(X) = \{ x \in E \mid B_x \subseteq X \},
\label{eq:setErosion}
\end{equation}
where $B_x$ denotes the translation of $B$ at point $x$ (i.e. $x + B$), and $\check{B}$ is the symmetrical of $B$ with respect to the origin.
The operators are then called morphological dilations and erosions. Details on these definitions and their properties can be found e.g. in~\cite{IB:LOS-07,Heijmans94,NajTal10,SERR-82}.

The structuring element $B$ defines a neighborhood that is considered at each point. This is typically the case in image processing and computer vision, where the underlying lattice is built on sets or functions of the spatial domain. It is a subset of $E$ with fixed shape and size, directly influencing the extent of the morphological operations. It is generally assumed to be compact, so as to guarantee good properties. In the discrete case (that will be considered all through this paper), we assume that it is connected, according to a discrete connectivity defined on $E$.

The general principle underlying morphological operators consists in translating
the structuring element at every position in space and checking if this translated
structuring element satisfies some relation with the original set (intersection for dilation, Equation~\ref{eq:setDilation}, inclusion for erosion, Equation~\ref{eq:setErosion})~\cite{SERR-82}.

\begin{figure}[htbp]
\begin{center}
\begin{tabular}{cccc}
\includegraphics[width=.5cm]{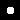} &
\includegraphics[width=3cm]{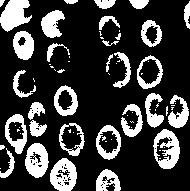} &
\includegraphics[width=3cm]{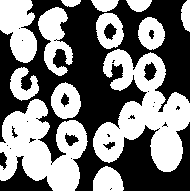} &
\includegraphics[width=3cm]{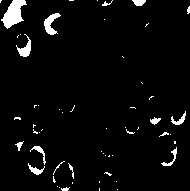} \\
(a) & (b) & (c) & (d)\\
\end{tabular}
\end{center}
\caption{(a) Structuring element $B$ (ball of the Euclidean distance). (b) Subset $X$ in the Euclidean plane (in white). (c) Its dilation $\delta_B(X)$. (d) Its erosion $\varepsilon_B(X)$.}
\label{fig:examplesMM}
\end{figure}

An example on a binary image is displayed in Figure~\ref{fig:examplesMM}.

The structuring element can also be seen as a binary relation between points~\cite{IB:LOS-07}, i.e. $y \in B_x$ iff $R(x,y)$ where $R$ denotes a relation on $E \times E$. Dilation and erosion are then expressed as follows:
\[
\delta(X) = \{ x \in E \mid \exists y \in X, R(y,x) \},
\]
\[
\varepsilon(X) = \{ x \in E \mid \forall y \in E, R(x,y) \Rightarrow y \in X \}.
\]
These formulas apply for any binary relation $R$. If $R$ is reflexive (i.e. $R(x,x)$ for all $x$), then $\delta$ is extensive ($X \subseteq \delta(X)$) and $\varepsilon$ is anti-extensive ($\varepsilon(X) \subseteq X$). These properties hold in the case illustrated in Figure~\ref{fig:examplesMM}. The objects in the original image are then expanded by dilation, to an extent that depends on the shape and the size of the structuring element, and reduced by erosion. Similar interpretations hold for any relation $R$, and these properties will also be important in the remainder of this paper.

\subsection{Lattice of formulas and morpho-logic}

The idea of using mathematical morphology in a logical framework has been first introduced in~\cite{IPMU-00b,TCIS-02}.
Let $PS$ be a finite set of propositional symbols, with $|PS|=N$. The set of formulas  (generated by $PS$ and the usual connectives) is denoted by
$\Phi$.
Well-formed formulas are denoted by Greek letters
$\varphi$, $\psi$... The set of all interpretations for $\Phi$ is denoted by $\Omega = \mathbf{2}^{|PS|}$, interpretations are denoted by $\omega$,
$\omega'$..., and
$\llbracket \varphi \rrbracket = \{\omega \in \Omega  \mid \omega \models \varphi\}$
is the set of all models of $\varphi$ (i.e.\ all interpretations for which $\varphi$ is true).

The underlying idea for constructing morphological operations on logical
formulas is to consider formulas and interpretations from a set theoretical perspective.
Since $\Phi$ is isomorphic to $\mathbf{2}^{\Omega}$ up to the syntactic equivalence, i.e., knowing a
formula defines completely the set of its models (and conversely, any set of models corresponds to a subset of $\Phi$ built of syntactic equivalent formulas), we can
identify $\varphi$ with the set of its models $\llbracket \varphi \rrbracket$, and
then apply set-theoretic morphological operations.
We recall that $\llbracket \varphi \vee \psi \rrbracket = \llbracket \varphi \rrbracket \cup \llbracket \psi \rrbracket$,
$\llbracket \varphi \wedge \psi \rrbracket = \llbracket \varphi \rrbracket \cap \llbracket \psi \rrbracket$,
$\llbracket \varphi \rrbracket \subseteq \llbracket \psi \rrbracket$ iff $\varphi \models \psi$, and $\varphi$ is consistent iff $\llbracket \varphi \rrbracket \neq \emptyset$.
 Considering the inclusion relation on $\mathbf{2}^\Omega$, $(\mathbf{2}^\Omega, \subseteq)$ is a Boolean complete lattice. Similarly a lattice (which is isomorphic to $\mathbf{2}^\Omega$) is defined on $\Phi_\equiv$, where $\Phi_\equiv$ denotes the quotient space of $\Phi$ by the equivalence relation between formulas (with the equivalence defined as $\varphi \equiv \psi$ iff $\llbracket \varphi \rrbracket = \llbracket \psi \rrbracket$). In the following, this is implicitly assumed, and we simply use the notation $\Phi$.
Any subset $\{ \varphi_i \}$ of $\Phi$ has a supremum $\bigvee_i \varphi_i$, and an infimum $\bigwedge_i \varphi_i$ (corresponding respectively to union and intersection in $\mathbf{2}^\Omega$). The greatest element is $\top$ and the smallest one is $\bot$ (corresponding respectively to $\mathbf{2}^\Omega$ and $\emptyset$).

Based on this lattice structure, it is straightforward to define a dilation as an operation that commutes with the supremum and an erosion as an operation that commutes with the infimum, as in Equations~\ref{eq:defDil} and~\ref{eq:defEro}. They naturally inherit all general properties of the algebraic framework.

\subsection{Morphological dilation and erosion of logical formulas}

Using the previous equivalences, we propose to define morphological dilation and erosion
of a formula with a structuring element as follows, according to the preliminary work in~\cite{IPMU-00b,TCIS-02}. The underlying lattice is $(\Phi_\equiv, \models)$, or equivalently $(\mathbf{2}^\Omega, \subseteq)$. Since these two lattices are isomorphic, we will use the same notations for morphological operations on each of them.

\begin{defi}
A morphological dilation of a formula $\varphi$ with a structuring element $B$ ($B \in \mathbf{2}^\Omega$) is defined through its models as:
\begin{equation}
\llbracket \delta_B(\varphi)\rrbracket = \delta_B(\llbracket \varphi \rrbracket) = \{ \omega \in \Omega \mid \check{B}_\omega \wedge \varphi \; consistent \}.
\label{eq:formDilation}
\end{equation}
Similarly, a morphological erosion is defined as:
\begin{equation}
\llbracket \varepsilon_B(\varphi)\rrbracket = \varepsilon_B(\llbracket \varphi \rrbracket) = \{ \omega \in \Omega \mid B_\omega \models \varphi \}.
\label{eq:formErosion}
\end{equation}
\end{defi}

In these equations, the structuring element $B$ represents a relationship
between worlds, i.e. $\omega' \in B_\omega$ iff $\omega'$ satisfies some
relationship with $\omega$. The condition in Equation
\ref{eq:formDilation} expresses that the set of worlds in relation to
$\omega$ should be consistent with $\varphi$. The condition in
Equation \ref{eq:formErosion} is stronger and expresses that all worlds
in relation to $\omega$ should be models of $\varphi$. Note that in this paper we only consider symmetrical structuring elements.

\medskip

There are several possible ways to define structuring elements in the context
of formulas. We suggest here a few ones. The relationship can be any relationship between worlds and defines a
``neighborhood'' of worlds. If it is symmetrical, it leads to symmetrical
structuring elements. If it is reflexive, it leads to structuring elements
such that $\omega \in B_\omega$, which leads to interesting properties,
as will be seen later. For instance, this relationship can be an
accessibility relation as in normal modal logics \cite{HughesCreswell68} (see~\cite{JANCL-02} for its use to define modalities as morphological operators).

An interesting way to choose the relationship is to base it on distances
between worlds. This allows defining sequences of increasing structuring
elements defined as the balls of a distance.
From any distance $d$ between worlds ($d: \Omega \times \Omega \rightarrow \mathbb{R}^+$), a distance from a world
to a formula is derived as a distance from a point to a set:
$d(\omega, \varphi) = \min_{\omega' \models \varphi} d (\omega, \omega')$.
The most commonly used distance between worlds in knowledge representation
(especially in belief revision \cite{DALA-88}, belief update \cite{KatsunoMendelzon91},
merging \cite{KONI-98} or preference representation \cite{LAFA-00b})
is the Hamming distance $d_H$ where $d_H(\omega,\omega')$ is the
number of propositional symbols that are instantiated differently in both
worlds. By default, we take $d$ to be $d_H$, and this is the distance we will use in most of the examples developed in this paper. In this case, the distance takes values in $\mathbb{N}$. The extension of what follows to distances taking values in $\mathbb{R}^+$ is straightforward.
Note that all what follows applies for general dilations, not necessarily derived from $d_H$.

Then dilation and erosion of size $n$ are defined from
Equations \ref{eq:formDilation} and \ref{eq:formErosion} by using the
distance balls of radius $n$ as structuring elements (i.e. $B_\omega^n = \{ \omega' \mid d(\omega, \omega') \leq n \}$):
\begin{equation}
\llbracket \delta^n(\varphi)\rrbracket = \{ \omega \in \Omega \mid \exists \omega' \in \Omega , \omega' \models \varphi \; and \; d(\omega, \omega') \leq n \} = \{ \omega \in \Omega \mid d(\omega, \varphi) \leq n \},
\label{eq:distDilation}
\end{equation}
\begin{equation}
\llbracket \varepsilon^n(\varphi)\rrbracket = \{ \omega \in \Omega \; | \; \forall \omega' \in \Omega, d(\omega, \omega') \leq n \Rightarrow \omega' \models \varphi \} = \{ \omega \in \Omega \; | \; d(\omega, \neg\varphi) > n \}.
\label{eq:distErosion}
\end{equation}

Note that we have $\delta^0(\varphi) = \varepsilon^0(\varphi) = \varphi$. By convention, when there is no ambiguity, we will set $\delta(\varphi) = \delta^1(\varphi)$ and $\varepsilon(\varphi) = \varepsilon^1(\varphi)$. More generally, whatever the operator $f$, we define $f^1 (\varphi) = f(\varphi)$ and $f^n(\varphi) = f(f^{n-1}(\varphi))$ for $n>1$.

From operations with the unit ball we define the external (respectively
internal) boundary of $\varphi$ as $\delta^1(\varphi) \wedge \neg \varphi$
(respectively $\varphi \wedge \neg \varepsilon^1(\varphi)$), corresponding to the
 worlds that are exactly at distance 1 of $\varphi$ (respectively of
$\neg \varphi$).

\medskip

\begin{figure}[htbp]
\centerline{\hbox{
\includegraphics[height=3.5cm]{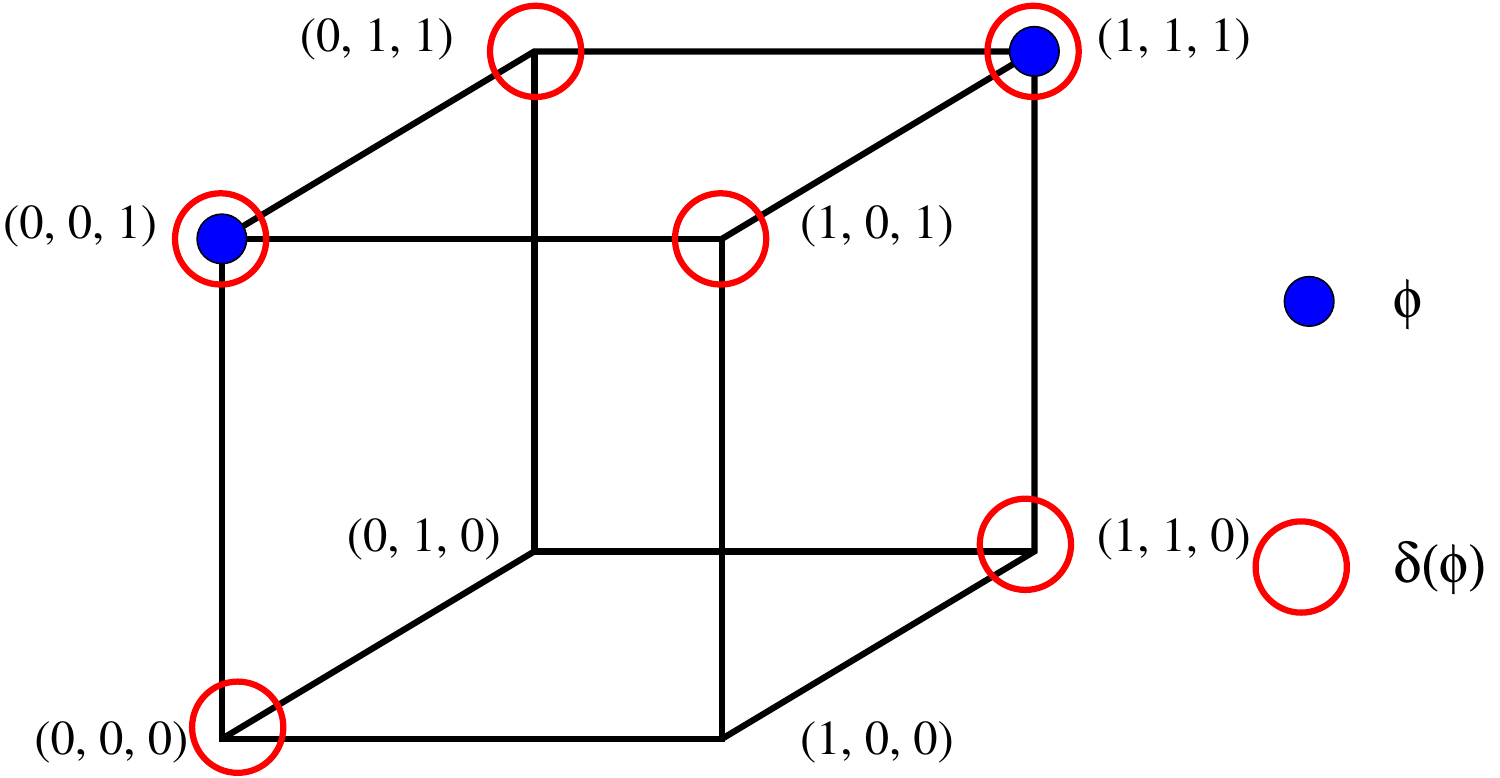}
}}
\caption{Example of a dilation of size 1: $\varphi = (a \wedge b \wedge c) \vee (\neg a \wedge \neg b \wedge c)$ and $\delta(\varphi) = (\neg a \vee b \vee c) \wedge (a \vee \neg b \vee c)$. Note that in all figures, the models of the formulas are represented.}
\label{fig:exampleDil}
\end{figure}

\begin{figure}[htbp]
\centerline{\hbox{
\includegraphics[height=3.5cm]{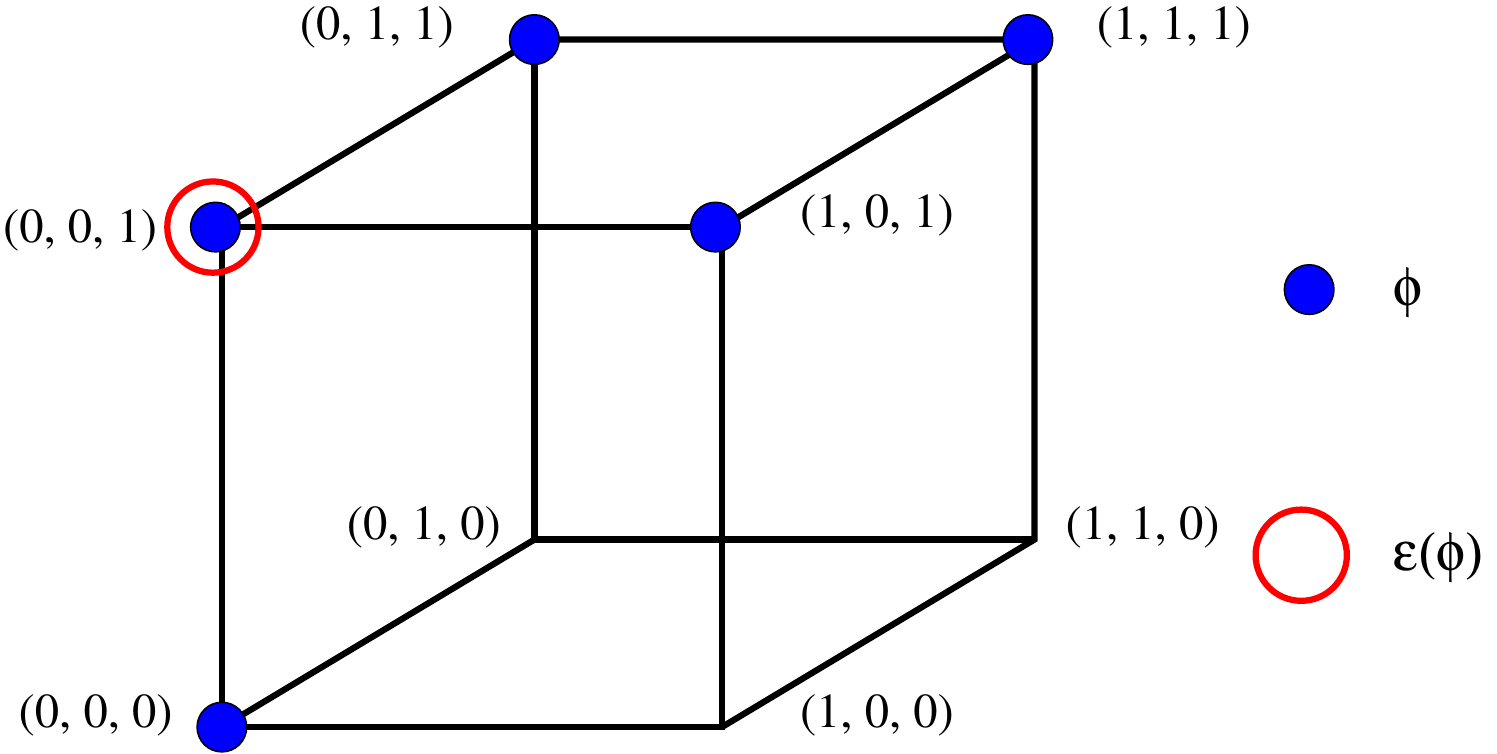}
}}
\caption{Example of an erosion of size 1: $\varphi = (a \wedge b \wedge c) \vee (\neg a \wedge b \wedge c) \vee (a \wedge \neg b \wedge c) \vee (\neg a \wedge \neg b \wedge c) \vee (\neg a \wedge \neg b \wedge \neg c)  (=c \vee ( \neg a \wedge \neg b))$ and $\varepsilon(\varphi) = \neg a \wedge \neg b \wedge c$.}
\label{fig:exampleEro}
\end{figure}

As an illustrative example, let us consider the case where we have three propositional symbols $a$, $b$ and $c$. The set of worlds $\Omega$ has then 8 elements, which can be represented as the vertices of a cube. In this example, we consider the unit cube of $\mathbb{R}^3$ (for $N$ propositional symbols, this generalizes to the hypercube of $\mathbb{R}^N$). For the sake of simplicity, we assimilate a formula formed by a simple conjunction of symbols with its corresponding model. For instance $a \wedge b \wedge c$ is assimilated to the corresponding world in $2^\Omega$, represented by the point $(1,1,1)$ in the unit cube.
%, i.e. $\mathbf{2}^N$ as in the introduction.
The edges link two worlds differing by one instantiation of a propositional symbol (i.e. at a Hamming distance of 1). For instance vertices representing $a \wedge b \wedge c$ and $\neg a \wedge b \wedge c$ are linked by an edge (we have $d(a \wedge b \wedge c, \neg a \wedge b \wedge c) = 1$). This is a convenient representation for graphically illustrating the morphological operations, as shown in Figures~\ref{fig:exampleDil} and~\ref{fig:exampleEro}. The balls of the Hamming distance are used as structuring elements. In Figure~\ref{fig:exampleDil}, we consider a formula $\varphi = (a \wedge b \wedge c) \vee (\neg a \wedge \neg b \wedge c)$. Its dilation (of size 1, i.e. by a ball of radius 1) is then $\delta(\varphi) = \neg((a \wedge \neg b \wedge \neg c) \vee ( \neg a \wedge b \wedge \neg c)) = (\neg a \vee b \vee c) \wedge (a \vee \neg b \vee c)$. The dilation of size one just amounts to add to the vertices representing $\varphi$ the vertices linked by an edge to them.
In Figure~\ref{fig:exampleEro}, an example of erosion is illustrated, for $\varphi = (a \wedge b \wedge c) \vee (\neg a \wedge b \wedge c) \vee (a \wedge \neg b \wedge c) \vee (\neg a \wedge \neg b \wedge c) \vee (\neg a \wedge \neg b \wedge \neg c) = c \vee ( \neg a \wedge \neg b)$. The erosion of size 1 is then $\varepsilon(\varphi) = \neg a \wedge \neg b \wedge c$. It amounts to keep in the result only the vertices having all their neighbors (according to the graph defined by the cube) in $\varphi$.

\medskip

The main properties of dilation and erosion, which are satisfied in
mathematical morphology on sets, hold also in the logical setting proposed
here. They are summarized below. The proofs are not given here,
 but they
are straightforward based on set/logic equivalences.

%\begin{proposition}
The dilations and erosions defined in Equations~\ref{eq:formDilation}, \ref{eq:formErosion}, \ref{eq:distDilation}, and~\ref{eq:distErosion} have the following properties:
\begin{description}
\item{Adjunction relation:} $(\varepsilon_B, \delta_B)$ is an adjunction, i.e. $\delta_B(\psi) \models \varphi$ iff $\psi \models \varepsilon_B(\varphi)$, for any structuring element $B$. This shows that the proposed definitions are a particular case of general algebraic dilations and erosions.

\item{Commutativity with union or intersection:} Dilation commutes with
union or disjunction (this is a fundamental property of dilation as mentioned in the general algebraic framework, and is derived from the adjunction property): for any family
$\varphi_1, ... \varphi_m$ of formulas, we have:
$\delta_B(\vee_{i=1}^m \varphi_i) = \vee_{i=1}^m \delta_B(\varphi_i)$.
Erosion on the other hand commutes with intersection or conjunction.
Note that this property is taken as definition in case of a general algebraic dilation or erosion.

In general, dilation (respectively erosion) does not commute with intersection
(respectively union), and only an inclusion relation holds:
$\delta_B(\varphi \wedge \psi) \models \delta_B(\varphi) \wedge \delta_B(\psi)$.

\item{Monotonicity:} Both operators are increasing with respect
to $\varphi$, i.e. if $\varphi \models \psi$, then
$\delta_B(\varphi) \models \delta_B(\psi)$ and $\varepsilon_B(\varphi) \models \varepsilon_B(\psi)$,
for any structuring element $B$. Dilation is increasing with respect to the structuring element, while erosion is decreasing, i.e. if $\forall \omega \in \Omega, B_{\omega} \subseteq B'_{\omega}$, then $\delta_B(\varphi) \models \delta_{B'}(\varphi)$ and $\varepsilon_{B'}(\varphi) \models \varepsilon_B(\varphi)$.

\item{Extensivity and anti-extensivity:} Dilation is extensive
($\varphi \models \delta_B(\varphi)$) if and only if $B$ is derived from a reflexive
relation (as is the case for distance based dilation, since if
$\omega \models \varphi$, then $d(\omega, \varphi) = 0$), and erosion
is anti-extensive ($\varepsilon_B(\varphi) \models \varphi$) under the same
conditions. We will always assume extensive dilations and anti-extensive erosions in the following.

\item{Iteration:} Dilation and erosion satisfy an
iteration property:
\[
\forall B, B', \forall \varphi, \delta_B(\delta_B')(\varphi) = \delta_{\delta_B(B')}(\varphi) \; \; \; \varepsilon_B(\varepsilon_B')(\varphi) = \varepsilon_{\delta_B(B')}(\varphi).
\]
For instance for distance based operations, for a distance satisfying the betweeness property\footnote{Let $d$ be a discrete metric on a set $M$. We say that $d$ has the
  \emph{betweenness property} if for all $x, y \in M$ and all
  $k \in \{0, 1, \dots, d(x, y)\}$ there exists $z \in M$ such that
  $\delta(x, z) = k$ and $\delta(z, y) = d(x, y) - k$. The Hamming distance has this property.}, this property can be expressed as:
\[
\delta^{n+n'}(\varphi) = \delta^{n'}[\delta^n(\varphi)] = \delta^n[\delta^{n'}(\varphi)],
\]
\[
\varepsilon^{n+n'}(\varphi) = \varepsilon^{n'}[\varepsilon^n(\varphi)] = \varepsilon^n[\varepsilon^{n'}(\varphi)].
\]
This means that the effect of these operations increases with the size of
 the structuring element, and that the computation can be done either by
successive applications of ``small'' structuring elements or directly by
the sum of the structuring elements.

\item{Duality:} Dilation and erosion
are dual operators with respect to the negation:
$\varepsilon_B(\varphi) = \neg \delta_B(\neg \varphi)$
which allows deducing properties of an operator from those of its dual
operator.

\item{Relations to distances:} Equation \ref{eq:distDilation} shows how
to derive a dilation from a distance. Conversely, from Equation~\ref{eq:distDilation} we have:
$d(\omega, \varphi) = \min \{ n \in \mathbb{N} \mid \omega \models \delta^n(\varphi) \}$,
and similarly, we have
$d(\omega, \neg \varphi) = \min \{ n \in \mathbb{N} \mid \omega \models \neg
\varepsilon^n(\varphi) \}$.

Distances between formulas can also be derived from dilation, as minimum distance and Hausdorff distance\footnote{Note that, in constrast to the Hausdorff distance, the minimum distance is improperly called distance since it does not satisfy all the properties of a true metric.}. For instance the minimum distance is expressed as:
$d_{min}(\varphi, \psi) = \min_{\omega \models \varphi, \omega' \models \psi} d_H(\omega, \omega') = \min \{ n \in \mathbb{N} \mid \delta^n(\varphi) \wedge \psi \neq \emptyset \; and \; \delta^n(\psi) \wedge \varphi \neq \emptyset \}$.
This means that the minimum distance is attained for the minimum size of dilation of both formulas such that they become consistent. The Hausdorff distance is defined as: $d_{Haus}(\varphi, \psi) = \max(\max_{\omega \models \varphi}d(\omega, \psi), \max_{\omega' \models \psi}d(\omega', \varphi))$. It can be computed from dilation by $d_{Haus}(\varphi,\psi) = \min \{ n \in \mathbb{N} \mid \varphi \models \delta^n(\psi) \mbox{ and } \psi \models \delta^n(\varphi) \}$.
\end{description}
%\end{proposition}

These properties will be used intensively in the applications of these operators for knowledge representation and reasoning.

\subsection{Some derived operators}
\label{sec:derivedOperators}

\paragraph{Conditional dilation and erosion and reconstruction}

In a number of problems and applications, we may want to restrict the result of an operation to stay within some domain, or to satisfy a particular formula. This is typically the case for instance if a result has to satisfy a theory, or a set of integrity constraints.
This idea calls for geodesic distances, from which structuring elements are derived, as the balls of this distance. Using these structuring elements in the definitions of dilation and erosion (Equations~\ref{eq:formDilation} and~\ref{eq:formErosion}) leads to the notion of geodesic, or conditional, operators. In the discrete case, that we consider here, the expression of these operators is very simple:
\begin{equation}
\delta_\psi^n(\varphi) = [\delta^1(\varphi) \wedge \psi ]^n,
\label{geodDilation}
\end{equation}
where $\psi$ denotes the conditioning formula, $n$ is the size of the structuring element, $\delta^1$ denotes the dilation using a ball of radius 1 (not geodesic) and the superscript $n$ means that the succession of dilation of size 1 and conjunction has to be performed $n$ times. This equation is a short writing for the following sequence of operations:
\begin{tabbing}
{\tt begin}\\
$\varphi_0 := \varphi \wedge \psi$;\\
{\tt For $i=1...n$} \\
\hspace{1cm} \= $\varphi_i := \delta^1(\varphi_{i-1}) \wedge \psi$;\\
{\tt end for}\\
{\tt Return $\varphi_n = \delta_\psi^n(\varphi)$}
\end{tabbing}

Similarly the geodesic erosion of $\varphi$ conditionally to $\psi$ can be computed as:
\begin{equation}
\varepsilon_\psi^n(\varphi) = [\varepsilon^1(\varphi) \vee \psi ]^n.
\label{geodErosion}
\end{equation}

\begin{figure}[htbp]
\centerline{\hbox{
\includegraphics[height=3.5cm]{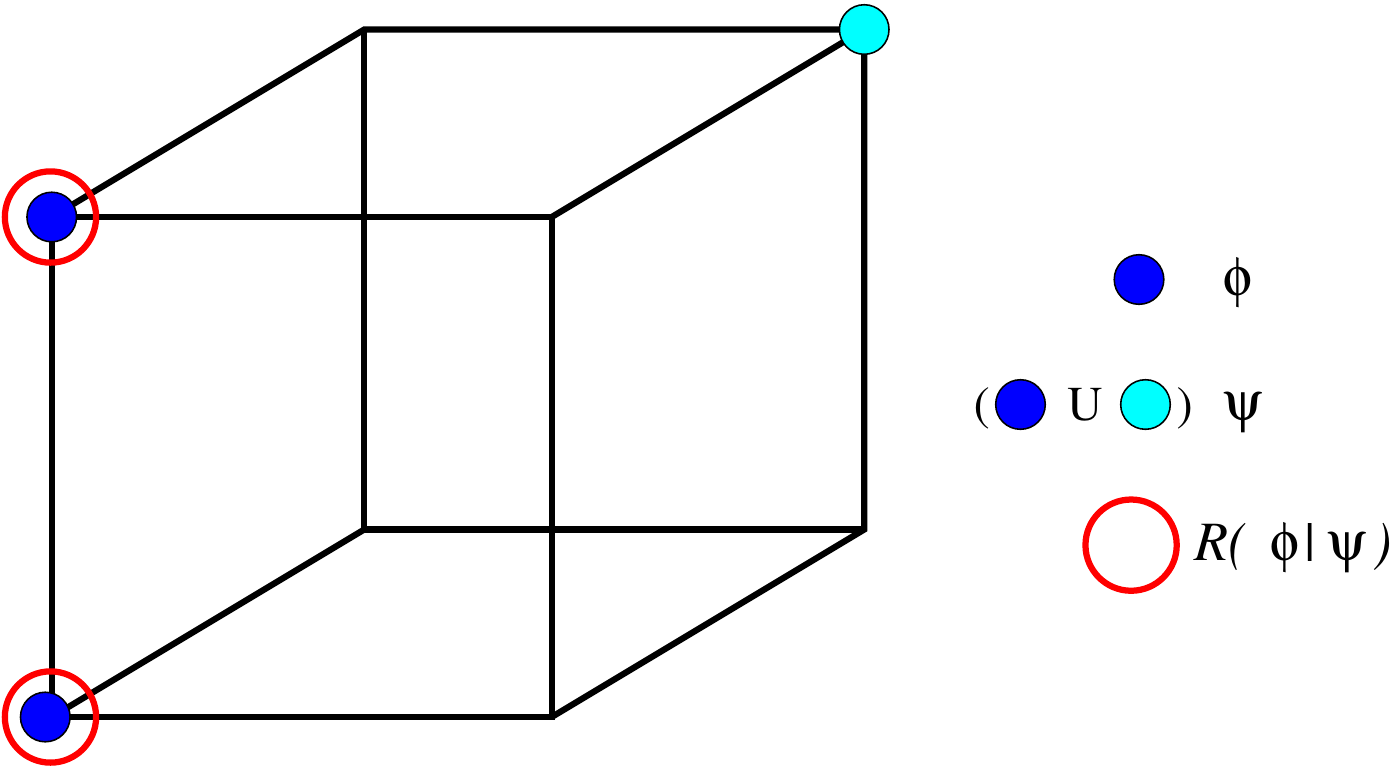}
}}
\caption{Reconstruction: only the connected component of $\psi$ which is ``marked'' by $\varphi$ is reconstructed.}
\label{fig:recons}
\end{figure}

If the conditional dilations are iterated until convergence, then the result is called reconstruction, and is denoted by $R(\varphi \mid \psi)$:
\begin{equation}
R(\varphi \mid \psi) = [\delta^1(\varphi) \wedge \psi ]^\infty.
\end{equation}
Note that in practice this sequence converges in a finite number of steps, when we consider a finite discrete space, as is the case in this paper. An example is illustrated in Figure~\ref{fig:recons}, with the same type of representation as in the previous figures. The reconstruction results in the only connected component of $\psi$ ``marked'' by $\varphi$.

\paragraph{Searching for the most central models satisfying a formula}

In some problems, it might be interesting to find the most relevant worlds that are models of a formula. This problem is solved in \cite{LAFA-00b} by taking the absolute maximum of the internal distance function (i.e. the function that associates to each world its distance to $\neg \varphi$). Mathematical morphology offers other tools that could also be interesting:
\begin{description}
\item [Ultimate erosion] is one of them. It consists in eroding iteratively $\varphi$ and, at each step $n$, keeping the connected components of $\varepsilon^n(\varphi)$ that disappear in $\varepsilon^{n+1}(\varphi)$. It corresponds exactly to the regional maxima of the internal distance (i.e. the function that assigns to each model of $\varphi$ the distance to its closest model of $\neg \varphi$). This approach may provide several components, which represent all parts of $\varphi$, belonging to different
 connected components, or connected by narrow sets of worlds. This notion can be formalized using the reconstruction operator (Definition~\ref{def:UE}).
 %It is an appropriate tool for solving Example~\ref{recommender} for instance, and $\left(doc \wedge short \wedge (recent \vee new)\right) \bigvee  \left( (I \vee F) \wedge comedy \wedge ov \wedge new \right)$ is exactly the ultimate erosion of $\varphi$.

\item [Last-non empty erosion] only keeps track of the largest component. Erosions are iterated and the last result before the erosion becomes empty is the final result. The result is then more restrictive than with ultimate erosion, and some component of $\varphi$ may not be represented. Definition~\ref{def:lastErosion} formalizes this idea.

\item [Morphological skeleton] is another approach to represent a formula in a compact and ``central'' way. It is defined as the union of the centers of maximal balls included in the initial formula (see~\cite{SERR-82} for definitions on sets and corresponding properties). This approach will not be further investigated in this paper.
\end{description}

\begin{defi}
\label{def:UE}
The ultimate erosion is expressed using the reconstruction operator as:
\begin{equation}
UE(\varphi) = \cup_{n \in \mathbb{N}} \left(\varepsilon^n(\varphi) \setminus R\left(\varepsilon^{n+1}(\varphi) \mid \varepsilon^n (\varphi)\right)\right).
\end{equation}
\end{defi}
Again in the finite discrete case, the iterative erosion process stops in a finite number of steps.

\begin{defi}
\label{def:lastErosion}
The last erosion of a formula $\varphi$, denoted by $\varepsilon_\ell(\varphi)$, is the erosion of $\varphi$ of the
largest possible size such that the set of worlds where
$\varepsilon_\ell(\varphi)$ is satisfied is not empty or the smallest size of erosion leading to a fixed point:
\begin{equation}
\varepsilon_\ell(\varphi) = \varepsilon^n(\varphi) \Leftrightarrow \left \{
\begin{array}{l l}
\varepsilon^n(\varphi) \not\vdash \bot, \\
\mbox{and } \forall m > n, \; \varepsilon^m(\varphi) \vdash \bot \mbox{ or } \varepsilon^m (\varphi) = \varepsilon^n(\varphi),
\end{array}
\right.
\end{equation}
with $n$ the smallest value for which this holds, and $\varepsilon^0(\varphi) = \varphi$.
\end{defi}

In the example of Figure~\ref{fig:exampleEro}, the first erosion is also the last non-empty erosion.

It is interesting to note that the idea of successive erosions is related to the notions of supermodels \cite{Ginsberg98} and of preferred explanations \cite{PINO-99}. For instance, it is easy to prove that $\omega \models \varepsilon^k(\varphi)$ iff $\omega$ is a $(k,0)$-supermodel of $\varphi$. The application to preferred explanations will be further investigated in Section~\ref{sec:abduction}.

\paragraph{Opening and closing}

Two other important operators are opening and closing. An
algebraic opening is an operator that is increasing, idempotent and anti-extensive, and an algebraic closing is an operator that is increasing, idempotent and extensive. Typical examples are $\delta\varepsilon$ and $\varepsilon\delta$ where $(\varepsilon, \delta)$ is an adjunction, as seen in the general algebraic framework. An important property if that any disjunction of openings is an opening, and any conjunction of closings is a closing.
Opening and closing
of a formula $\varphi$ by a structuring element $B$ are defined respectively as: $O_B(\varphi) = \delta_{B}(\varepsilon_B(\varphi))$, and
$C_B(\varphi) = \varepsilon_{B}(\delta_B(\varphi))$.

These two basic morphological filters can be seen as approximation operators, since they ``simplify'' formulas by either suppressing some irregularities for opening, or adding some parts of $\neg \varphi$ for closing.
Families of filters can be built from these two ones. For instance, granulometry \cite{SERR-82} consists in applying successively openings with structuring elements of increasing size, such decomposing a formula in parts of different characteristic sizes. Another example is alternate sequential filters \cite{SERR-88}, which consist in building sequences of opening/closing (or closing/opening), with structuring elements of increasing size. Such transformations are increasing and idempotent, and allow filtering progressively parts of $\varphi$ and $\neg \varphi$.

Note that $\varepsilon_\ell$ is an anti-extensive and idempotent operator, but it is not increasing (and hence not an opening). The same applies for ultimate erosion.

\subsection{Morphological ordering}
\label{sec:ordering}

\begin{figure}[htbp]
\centerline{\hbox{
\includegraphics[width=4cm]{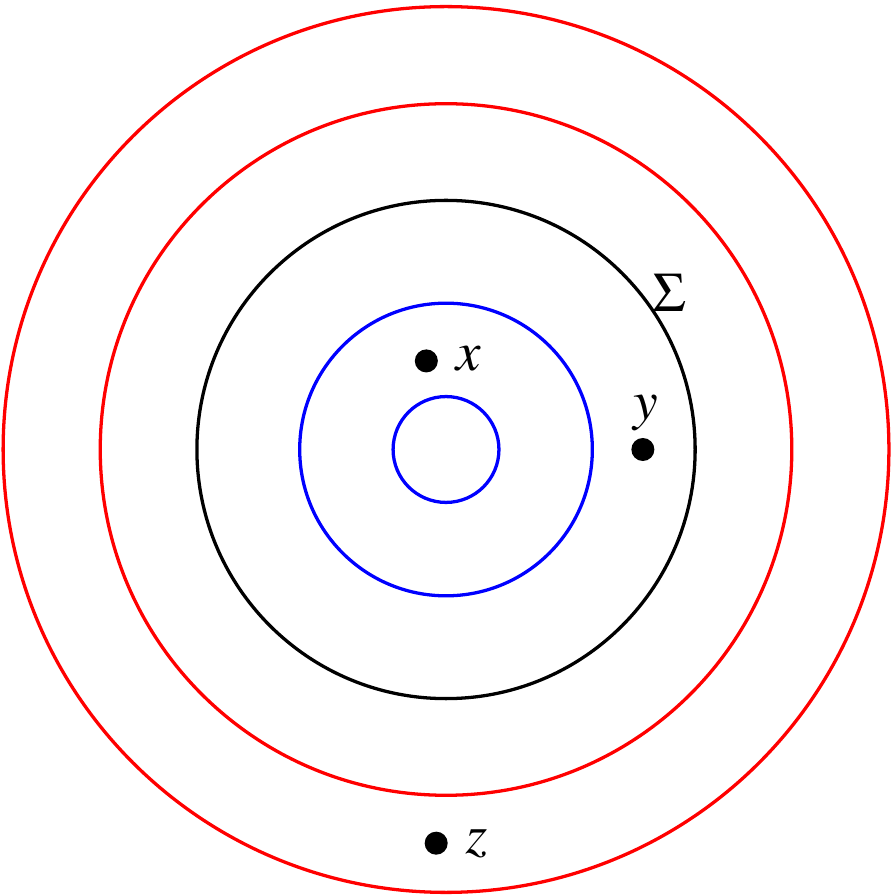}
}}
\caption{Illustration of a natural partial ordering derived from successive erosions (in blue) and dilations (in red) of $\Sigma$. We have $x \preceq_f y \preceq_f z$ in this example.}
\label{fig:orderSet}
\end{figure}

Given a formula, a natural ordering can be derived from the sequence of its successive erosions and dilations, for a given elementary structuring element (of size 1). This idea is illustrated on sets in Figure~\ref{fig:orderSet}. This will be particularly interesting in the following, when considering a theory, and for defining a partial order on the models satisfying this theory (by identifying a theory with an equivalent formula). We call it morphological ordering.

\begin{defi}
\label{def:fundamentalOrder}
Let $\Sigma$ be a theory (represented by a formula) or a formula. Let $n$ be the maximal size of dilation and $m$ the size of the last non-empty erosion, i.e.:
\[
\varepsilon^m(\Sigma) = \varepsilon_\ell(\Sigma),
\]
\[
\delta^n(\Sigma) = \delta_\ell(\Sigma),
\]
where $\delta_\ell$ is defined in a similar way as the last erosion (and $\delta_\ell(\Sigma)$ can be either $\top$ or a fixed point).
Then we define the fundamental sequence $(T_i)$ of subsets of $\Omega$ associated with $\Sigma$,
from $i=0$ to $i=n+m$, as follows:
\[
T_i=\left\{\begin{array}{lcl}
\llbracket \varepsilon^{m-i}(\Sigma) \rrbracket &\quad&\mbox{if}\;\; i\leq m\\
\llbracket \delta^{i-m}(\Sigma) \rrbracket &\quad&\mbox{if}\;\; i>m
\end{array}\right.
\]
The morphological total pre-order associated to $\Sigma$ is then defined by:
\begin{equation}\label{order3}
\omega \preceq_f \omega' \iffdef \forall k \;\; (\omega'\in T^k \Rightarrow \omega\in T^k).
\end{equation}
\end{defi}
The fact that this defines a pre-order is easy to check.
Note that this ordering depends on the choice of the elementary structuring element.

\begin{figure}[htbp]
\centerline{\hbox{
\includegraphics[height=3.5cm]{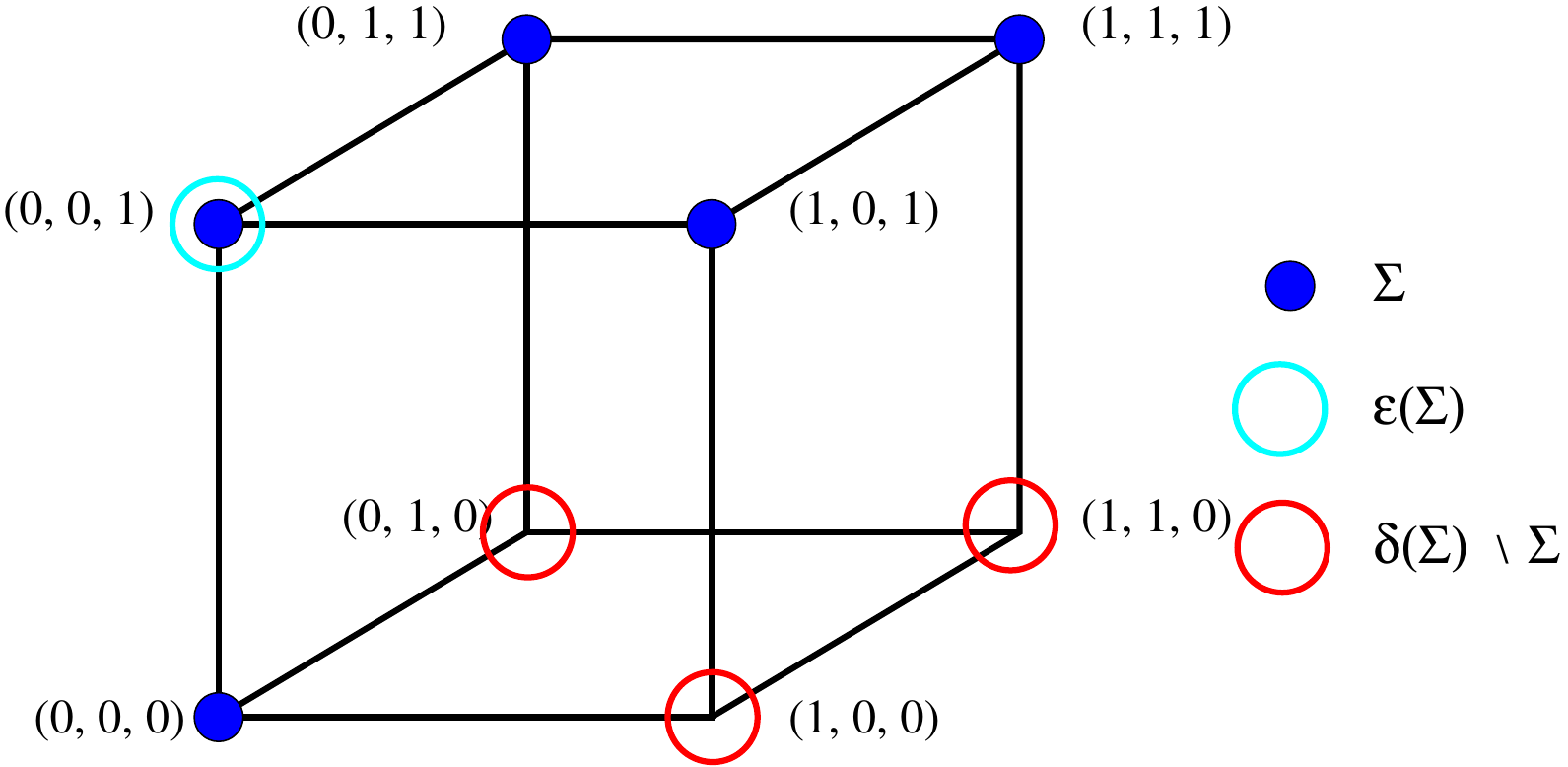}
}}
\caption{Illustration of the morphological ordering (here $n = 1, m = 1$).}
\label{fig:fundamentalOrder}
\end{figure}

As an example, let us consider again three propositional symbols, with the same representation as in Figures~\ref{fig:exampleDil} and~\ref{fig:exampleEro}, and $\Sigma = \{ a \rightarrow c, b \rightarrow c \}$ (represented by the same formula $\varphi$ as in the example of Figure~\ref{fig:exampleEro}). The models of $\Sigma$ are $\Omega \setminus \{ a \wedge b \wedge \neg c , a \wedge \neg b \wedge \neg c, \neg a \wedge b \wedge \neg c\}$. We have $\llbracket \delta(\Sigma) \rrbracket = \Omega$, $\llbracket \varepsilon(\Sigma) \rrbracket = \{ \neg a \wedge \neg b \wedge c\}$, and $\llbracket \varepsilon^2(\Sigma) \rrbracket = \emptyset$, as illustrated in Figure~\ref{fig:fundamentalOrder}.

This provides a stratification of the elements of $\Omega$, as given in Table~\ref{tab:stratification}.

\begin{table}[htbp]
\begin{center}
\begin{tabular}{| c | c |}\hline
0 & $\neg a \wedge \neg b \wedge c$\\
1 & $\neg a \wedge \neg b \wedge \neg c, a \wedge \neg b \wedge c, \neg a \wedge b \wedge c, a \wedge b \wedge c$\\
2 & $a \wedge \neg b \wedge \neg c, \neg a \wedge b \wedge \neg c, a \wedge b \wedge \neg c$\\\hline
\end{tabular}
\caption{Stratification of the elements of $\Omega$ according to the morphological ordering associated with $\Sigma = \{ a \rightarrow c, b \rightarrow c \}$.}
\label{tab:stratification}
\end{center}
\end{table}

Note that in case the last dilation yields a fixed point different from $\top$, the rank of the models in $\Omega \setminus \llbracket \delta_\ell(\Sigma) \rrbracket$ is set to $+\infty$ by convention. This amounts to ordering only $\llbracket \delta_\ell(\Sigma) \rrbracket$.

%Another example for three propositional symbols is given in Figure~\ref{fig:orderCube}.
%
%\begin{figure}[htbp]
%\centerline{\hbox{
%\includegraphics[height=3.5cm]{Figures/orderCube.pdf}
%}}
%\caption{Illustration of the fundamental ordering (here $n = m = 1$). We have $T^0 = \varepsilon(\Sigma)$, $T^1 = \Sigma$, $T^2 = \delta(\Sigma)$, and $\omega_1 \preceq_f \omega_2 \preceq_f \omega_3$.}
%\label{fig:orderCube}
%\end{figure}

\begin{proposition}
The following properties hold:
\begin{itemize}
\item The subsets $T_i$ of $\Omega$ are nested, i.e. $\forall i \in [0 ... (n+m-1)], T_i \subseteq T_{i+1}$ for the considered dilations and erosions (with structuring elements such that $\omega \in B_\omega$).
\item The relation $\preceq_f$ is reflexive and transitive, i.e. a pre-order, which is moreover total.
\item Let $R_e$ be the relation defined on $\mathbf{2}^\Omega$ by $R_e (\omega, \omega')$ iff $\max \{ k \in [0 ... (n+m)] \mid \omega \in T^k \} =  \max \{ k \in [0 ... (n+m)] \mid \omega' \in T^k \}$. This relation is an equivalence relation and the ordering induced by $\preceq_f$ on the quotient space $\mathbf{2}^\Omega / R_e$ is a total ordering.
\end{itemize}
\end{proposition}

\begin{figure}[htbp]
\centerline{\hbox{
\includegraphics[height=3.5cm]{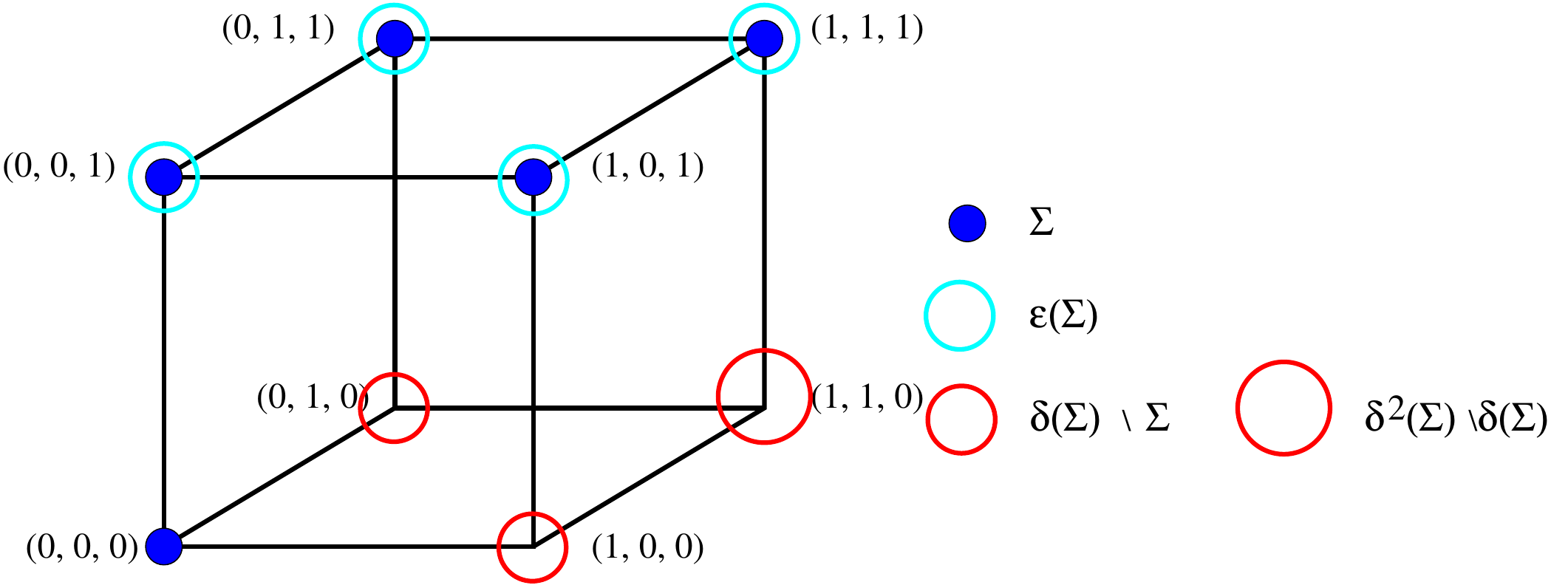}
}}
\caption{Illustration of the morphological ordering (here $n = 2, m = 1$), using $B^{ab}$ as structuring element.}
\label{fig:fundamentalOrder2}
\end{figure}

Let us briefly comment on the choice of the structuring element used in the morphological operations. When it is taken as a ball of the Hamming distance, as in all examples in this section so far, then the neighborhood it defines is isotropic and all variables are taken into account in the same way. However, different structuring elements could be used, and their choice is a way to impose preferences, for instance on some variables over other ones. As an example, let us consider the following structuring element, defining the neighborhood of any world $\omega \in \Omega$:
\[
B^{ab}_\omega = \{ \omega' \in B_\omega \mid \omega(c) = \omega'(c) \},
\]
where $B$ denotes the ball of radius 1 of the Hamming distance, and $\omega(c) = \omega'(c)$ means that $c$ is instantiated in the same way in $\omega$ and in $\omega'$. With this structuring element, $c$ is not handled in the same way as variables $a$ and $b$. Note that when performing successive erosions (respectively dilations) with such a structuring element, we may not end up with $\bot$ (respectively $\top$), but we may converge towards a fixed point (a subset of $\Omega$). Figure~\ref{fig:fundamentalOrder2} illustrates the effect of this structuring element on the same example as in Figure~\ref{fig:fundamentalOrder}. The derived morphological ordering and the corresponding stratification of $\Omega$ is now given in Table~\ref{tab:stratification2}.

\begin{table}[htbp]
\begin{center}
\begin{tabular}{| c | c |}\hline
0 & $\neg a \wedge \neg b \wedge c, a \wedge \neg b \wedge c, \neg a \wedge b \wedge c, a \wedge b \wedge c$\\
1 & $\neg a \wedge \neg b \wedge \neg c$\\
2 & $a \wedge \neg b \wedge \neg c, \neg a \wedge b \wedge \neg c $ \\
3 & $a \wedge b \wedge \neg c$\\\hline
\end{tabular}
\caption{Stratification of the elements of $\Omega$ according to the morphological ordering associated with $\Sigma=  \{ a \rightarrow c, b \rightarrow c \}$, using $B^{ab}$ as structuring element.}
\label{tab:stratification2}
\end{center}
\end{table}

As another way to handle variables differently, let us note that $\Omega$ does not need to be ``isotropic'', i.e. the cube in our illustrations could be a parallelepiped, with different lengths of the edges, representing the elementary distances between worlds. A distance between two worlds can then be defined as the length of a shortest path in this weighted graph. Structuring elements can be defined as balls of this distance. However, in general this distance does not satisfy the betweenness property, which makes is less interesting for our purpose.
%In this paper, we will keep the isotropic assumption, for the sake of simplicity.

It is important to note that the ordering of the elements of $\Omega$ depends on both $\Sigma$ and the definition of erosion and dilation, in particular the choice of the structuring element.

This morphological ordering will be used to unify several reasoning tasks, in particular abduction and revision, in Section~\ref{sec:abduction}.

%%%%%%%%%%%%%%%%%%%%%%%%%%%%%%%%%%%%%%%%%%%%%%%%%%
%%%%%%%%%%%%%%%%%%%%%%%%%%%%%%%%%%%%%%%%%%%%%%%%%%

%\input{computation}

\section{Computational issues}
\label{computations}

Unless stated otherwise, for all the operators considered here we assume that the structuring element is the ball of radius 1 for the Hamming distance.

\subsection{Dilation}

The commutativity of dilation with disjunction, along with the iteration
property, allows us to recover results of \cite{LAFA-00b}. In particular, the following result holds.

\begin{proposition}
\label{prop:dilConjLit}
Let $\varphi$ be a consistent conjunction of literals,
i.e. $\varphi = l_1 \wedge l_2 \wedge ... \wedge l_n$, then
\[
\delta^1(\varphi) = \vee_{j=1}^n ( \wedge_{ i \neq j} l_i).
\]
Similarly, if $\varphi$ is a disjunction of literals, i.e. $\varphi = l_1 \vee ... \vee l_m$, then the erosion is expressed as:
\[
\varepsilon^1(\varphi) = \wedge_{j=1}^m ( \vee_{i \neq j} l_i).
\]
In these equations $\delta^1$ (respectively $\varepsilon^1$) denotes the dilation (erosion) using as structuring element a ball of radius 1 of the Hamming distance.
\end{proposition}

This property, together with the commutation of dilation
with disjunction, gives the following result \cite{LAFA-00b}:
if $k$ is a fixed integer, then the dilation of size $k$
$\delta^k(\varphi)$ of a DNF formula $\varphi$ can be computed
in time ${\cal O}(n^k)$ -- thus in polynomial time.
In a similar way, erosion commutes with intersection and can be computed
in polynomial time from a CNF formula.

When $\varphi$ is not under DNF, computing $\delta^k(\varphi)$ directly
from $\varphi$ (without rewriting $\varphi$ under DNF first) is a
difficult problem.

However, we can prove a slightly general result:
\begin{proposition}\label{decompo}
If $\phi_1,\ldots, \phi_n$ are such that for all $i,j$, $\phi_i$ and $\phi_j$ do not share variables, then $\delta(\phi_1 \wedge \ldots \wedge \phi_n) =  \bigvee_{j=1}^n \left( \delta(\phi_j) \wedge \bigwedge_{k\neq j}\phi_k\right)$.
\end{proposition}

\beginproof
%We have the following chain of equivalences:\\
For every interpretation $\omega$ let $\omega_i = \omega^{\downarrow Var(\varphi_i)}$ be the projection of $\omega$ on the language of $\varphi_i$ ($Var(\varphi_i)$). We have
$\omega \models \delta(\varphi_1 \wedge \ldots \wedge \varphi_n)$ \iff \\
(1) there exists $\omega'$ such that $\omega' \models \varphi_1 \wedge \ldots \wedge \varphi_n$ and $d(\omega,\omega') \leq 1$.\\
 Now, $d(\omega,\omega') = \sum_{i= 1,\ldots,n}d(\omega_i,\omega_i')$ (since the $\varphi_i$ have no variable in common).
Therefore, $d(\omega,\omega') \leq 1$ \iff there exists a $j$, $j \leq n$, such that: (a) $d(\omega_j,\omega_j') \leq 1$, and (b) for every $k \neq j$, $\omega_k = \omega_k'$.
From this we get that (1) is equivalent to:\\
 (2) there exists a $j$, $j \leq n$, such that $\omega_j \models \delta(\varphi_j)$ and for every $k \neq j$, $\omega_k \models \varphi_k$. \\
Now, $\delta(\varphi_j)$ is equivalent to a formula on the language $Var(\varphi_i)$, therefore $\omega \models \delta(\varphi_j)$ iff $\omega_j \models \delta(\varphi_j)$,
Moreover, $\omega \models\varphi_k$ iff $\omega_k \models \varphi_k$.
Therefore, $\omega \models \delta(\varphi_1 \wedge \ldots \wedge \varphi_n)$ \iff there exists a $j$, $j \leq n$, such that $\omega \models \delta(\varphi_j) \wedge \bigwedge_{k\neq j}\varphi_k$, from which the result follows.
\endproof

In particular:
\begin{itemize}
\item  if $Var(\varphi) \cap Var(\psi) = \emptyset$, then $\delta(\varphi \wedge \psi) = (\varphi \wedge \delta(\psi)) \vee (\delta(\varphi) \wedge \psi)$;
\item if $\varphi_1,\ldots, \varphi_n$ are literals whose associated variables are all different, then we recover the identity $\delta(l_1 \wedge \ldots \wedge l_n) = \bigvee_{j=1}^n \bigwedge_{k\neq j}l_k$.
\end{itemize}

Now, how hard is it to compute dilations (respectively erosions) when $\varphi$ is not under DNF (respectively CNF)? First of all we have the following complexity results.

\begin{proposition}~
\begin{enumerate}
\item Given an interpretation $\omega$ and a formula $\varphi$, deciding whether $\omega \models \delta (\varphi)$ is \npc.
\item Given an interpretation $\omega$ and a formula $\varphi$, deciding whether $\omega \models \varepsilon (\varphi)$ is \conpc.\end{enumerate}
\end{proposition}

\beginproof
In both cases membership is straightforward. For hardness for point 1 we consider the following reduction from \sat: we map every formula $\alpha$ to $\langle \varphi, \omega\rangle$ where $\varphi = p \wedge \alpha$ with $p \neq Var(\alpha)$, and $\omega$ being any interpretation satisfying $p$. Using Proposition \ref{decompo} we have $\delta(p \wedge \alpha) \equiv (p \wedge \delta(\alpha)) \vee (\delta(p) \wedge \alpha)$, which is equivalent to $\alpha \vee (p \wedge \delta(\alpha))$. Now, if $\alpha$ is satisfiable, then so is $\delta(\alpha)$. Therefore, $\omega \models \alpha \vee (p \wedge \delta(\alpha))$. If  $\alpha$ is unsatisfiable, then  so are $\delta(\alpha)$ and $\alpha \vee (p \wedge \delta(\alpha))$. Therefore $\omega \not \models \alpha \vee (p \wedge \delta(\alpha))$.  The reduction from \unsat\ for point 2 is similar.
\endproof

This shows that, {\em a fortiori}, computing erosion or dilation in the general case is hard. Moreover, the size of $\varepsilon(\varphi)$ and $\delta(\varphi)$ is not polysize, except if $\p = \np$. It is not sure that there is a way of computing erosion (dilation) being more efficient than first rewriting $\varphi$ under CNF (DNF).

Note that inference from the dilation of a formula is
(theoretically) not harder than inference from the formula itself.
Namely, given any two formulas $\varphi$ and $\psi$ and any integer $k$,
determining whether $\delta^k(\varphi) \models \psi$ is {\sf coNP}-complete.
Obviously, a similar result holds for inference from erosion.

However, interesting results can be obtained for erosion by decomposing a formula into its connected components. Based on the graph interpretation used all through this paper, a connected component is classically defined as a connected component in the graph: we say that $\psi$ is a {\em connected component} of $\varphi$ if $\llbracket \psi \rrbracket$ is a connected component of the graph associated with $\varphi$ (whose set of vertices  is $\llbracket \varphi \rrbracket$) and whose set of edges is defined by $(\omega,\omega')$ whenever $d(\omega,\omega') \leq 1$).

\begin{proposition}\label{lemmasep}
If $d(\varphi, \psi) \geq 2$, for $d$ being the minimum distance between formulas, then $\varepsilon(\varphi \vee \psi) \equiv \varepsilon(\varphi) \vee \varepsilon(\psi)$.
\end{proposition}

\beginproof
Assume $d(\varphi, \psi) \geq 2$.
We already know that $\varepsilon(\varphi) \vee \varepsilon(\psi) \models \varepsilon(\varphi \vee \psi)$, so it remains to be proven that $\varepsilon(\varphi \vee \psi) \models \varepsilon(\varphi) \vee \varepsilon(\psi)$. Let $\omega \models \varepsilon(\varphi \vee \psi)$. This implies  $\omega \models \varphi \vee \psi$ if the erosion is anti-extensive (which is the case in this paper). Without loss of generality, assume $\omega \models \varphi$. Because $d(\varphi, \psi) \geq 2$, we have $d(\omega, \psi) \geq 2$. Now, assume that $\omega \not \models \varepsilon(\varphi)$, i.e., $d(\omega,\neg\varphi) \leq 1$; this means that there exists a $\omega'$ such that $\omega' \models \neg \varphi$ and $d(\omega,\omega') = 1$ ($d(\omega,\omega') = 0$ is impossible because $\omega \models \varphi$ and $\omega' \models \neg \varphi$). Now, we must have $\omega' \models \psi$; otherwise we would have $\omega' \models \neg \varphi \wedge \neg \psi$, hence $d(\omega, \neg \varphi \wedge \neg \psi) \leq 1$, which contradicts $\omega \models \varepsilon(\varphi \vee \psi)$. Therefore, $d(\varphi,\psi) \leq d(\omega,\omega') \leq 1$, which contradicts the assumption that $d(\varphi, \psi) \geq 2$.
\endproof

\begin{proposition}\label{lemmasep2Ero}
Let $\varphi_1,\ldots, \varphi_p$ be the connected components of $\varphi$. Then we have:
\[
\varepsilon(\varphi) \equiv \bigvee_{i=1}^p \varepsilon(\varphi_i).
\]
\end{proposition}

\beginproof
For any two distinct connected components $\varphi_i$, $\varphi_j$ of $\varphi$ we have $d(\varphi_i,\varphi_j) \geq 2$, therefore, $\varepsilon(\bigvee_{i=1}^p\varphi_p) \equiv \bigvee_{i=1}^p \varepsilon(\varphi_i)$; the fact that $\varphi \equiv \bigvee_{i=1}^p\varphi_p$ enables us to conclude that $\varepsilon(\varphi) \equiv \bigvee_{i=1}^p \varepsilon(\varphi_i)$.
\endproof

Now, we have to find a way of (a) computing the connected components of $\varphi$ and (b) computing $\varepsilon(\varphi)$. The first step is easy when $\varphi$ is under DNF. We first note the following fact:

\begin{proposition}\label{sep}
Let $\varphi = \psi_1 \vee \ldots \vee \psi_q$ be a DNF formula. For any $i, j \in \{1,\ldots, q\}$, $d(\psi_i,\psi_j)$ is equal to the number of disagreeing literals between $\psi_i$ and $\psi_j$.
\end{proposition}

For instance, we have $d(a\wedge \neg b \wedge c, b \wedge \neg c \wedge d) = 2$, $d(a\wedge \neg b \wedge c, b \wedge c \wedge d) = 1$, and $d(a\wedge \neg b \wedge c, c \wedge d) = 0$.

\begin{proposition}\label{ccdnf}
Let $\varphi = \psi_1 \vee \ldots \vee \psi_q$ be a DNF formula. Let $G_{\varphi}$ be the undirected graph defined by its set of vertices $\llbracket \varphi \rrbracket$, which can be grouped into subsets $\{a_1,\ldots,a_q\}$ where $a_i = \llbracket \psi_i \rrbracket$, and containing an edge $\{a_i,a_j\}$ iff $d(\psi_i,\psi_j) \leq 1$. Then the connected components of $G_\varphi$ correspond to the connected components of $\varphi$, and $\{a_i , i \in I \subseteq \{ 1,\ldots, q\} \}$ is a connected component of $G_\varphi$ iff $\bigvee_{i \in I}\psi_i$ is a connected component of $\varphi$.
\end{proposition}

\begin{figure}[htbp]
\centerline{\hbox{
\includegraphics[height=3.5cm]{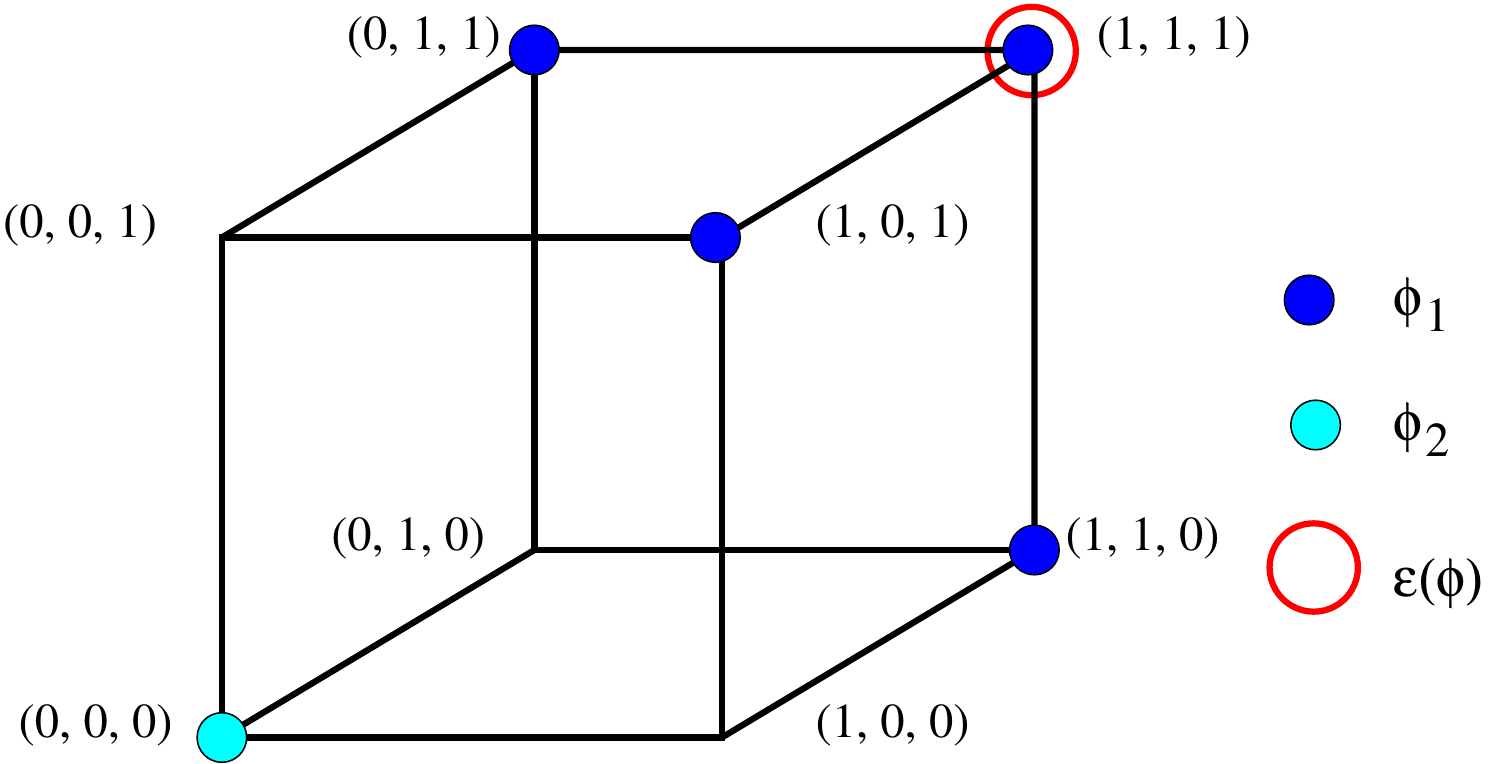}
}}
\caption{Decomposition of $\varphi$ into two connected components $\varphi_1$ and $\varphi_2$, and its erosion (only $a$, $b$ and $c$ are considered in this representation).}
\label{fig:exCC}
\end{figure}

\begin{example}\label{ex-ue}~
Let us consider $\varphi = (a \wedge b) \vee (a \wedge c) \vee (b \wedge c) \vee (\neg a \wedge \neg b \wedge \neg c \wedge \neg d)$ (Figure~\ref{fig:exCC}). The graph $G_\varphi$ has 8 vertices, grouped into 4 subsets $a_i$, and its edges are $\{a_1,a_2\}$, $\{a_1,a_3\}$, $\{a_2,a_3\}$, plus the reflexive edges $\{a_1,a_1\}$,  $\{a_2,a_2\}$, $\{a_3,a_3\}$, $\{a_4,a_4\}$. $G_\varphi$ has two connected components: $\{a_1,a_2,a_3\} = \{ (0, 1, 1), (1, 1, 1), (1, 0, 1), (1, 1, 0) \}$ and $\{a_4\} = \{ (0, 0, 0) \}$ (the valuation of $d$ is not represented here), therefore $\varphi$ has two connected components: $\varphi_1 = (a \wedge b) \vee (a \wedge c) \vee (b \wedge c)$ and $\varphi_2 = \neg a \wedge \neg b \wedge \neg c \wedge \neg d$, from which we have $\varepsilon(\varphi) =  \varepsilon(\varphi_1) \vee \varepsilon(\varphi_2) = (a \wedge b \wedge c) \vee \bot = a \wedge b \wedge c$.
\end{example}

\subsection{About last erosion and ultimate erosion}

Let us consider the last erosion (Definition~\ref{def:lastErosion}). Denote by $\ell(\varphi)$ the number of iterations to reach the last non-empty erosion of $\varphi$.

\begin{proposition}\label{bound1}
If $\not\models \varphi$ and $\varphi \not\equiv \top$ then $\ell(\varphi) \leq N-1$, where $N$ is the number of propositional symbols in the language.
\end{proposition}

\beginproof Let $k = \ell(\varphi)$. We have $\omega \models \varepsilon^k (\varphi)$ if for all $\omega' \models \neg \varphi$ we have $d(\omega,\omega') > k$. Therefore, $k < N$, because it can never be the case that $d(\omega,\omega') > N$.
%, we have that there is no  $\omega'$ such that $\omega' \models \neg \varphi$, that is, $\varphi$ is valid.
\endproof

Actually, we can find a better bound for $\ell(\varphi)$:

\begin{proposition}
\label{lemma6}
If $\not\models \varphi$  and $\varphi \not\equiv \top$ then $\ell(\varphi)$ is less than the length of the shortest prime implicate of $\varphi$ (the set of prime implicates being denoted by $PI(\varphi)$).
\end{proposition}

\beginproof
The result follows easily from $\varphi \equiv \bigwedge PI(\varphi)$, from the fact that erosion commutes with conjunction, and from the following expression of the erosion of a disjunction of literals:
\[
\varepsilon(l_1 \vee ... \vee l_m) = \wedge_{j=1}^m ( \vee_{i \neq j} l_i),
\]
this result being obtained by duality from Proposition~\ref{prop:dilConjLit} (or directly by induction on $m$).
\endproof

For instance let us consider $\varphi = (a \leftrightarrow b)$. We have $PI(\varphi) = \{a \vee \neg b, \neg a \vee b\}$, i.e., every prime implicate of $\varphi$ is of length 2; $\varepsilon^1(\varphi) = \bot$, therefore $\ell(\varphi) = 0$. This example shows that $\ell(\varphi)$ can be strictly lower than the bound expressed in Proposition~\ref{lemma6}.

Proposition \ref{bound1} enables us to say that deciding whether $\omega \models \varepsilon_\ell(\varphi)$ is in \bhdeux in the Boolean hierarchy of {\sf NP} sets.

Let us now consider ultimate erosion (Definition~\ref{def:UE}). The following result directly follows from Proposition~\ref{lemmasep2Ero}.

\begin{proposition}\label{lemmasep2UE}
Let $\varphi_1,\ldots, \varphi_p$ be the connected components of $\varphi$. Then we have: $UE(\varphi) \equiv \bigvee_{i=1}^p UE(\varphi_i)$.
\end{proposition}

Using Proposition \ref{lemmasep2UE}, the following algorithm computes the ultimate erosion of $\varphi$.

\begin{tabbing}
$UE(\varphi)$:\\
{\tt begin}\\
decompose $\varphi$ into its connected components $\varphi_1,\ldots,\varphi_p$;\\ %using $M_\varphi$;\\
{\tt if} $p = 1$\\
{\tt then} \= {\tt if} $\varepsilon(\varphi) \equiv \bot$ \\
\> {\tt then}  {\tt return} $\varphi$\\
\> {\tt else}  {\tt return} $UE(\varepsilon(\varphi))$\\
\> {\tt endif} \\
{\tt else} {\tt return} $UE(\varphi_1) \vee \ldots \vee UE(\varphi_n)$\\
{\tt endif}\\
%{\tt end}
\end{tabbing}

\subsection{About opening and skeleton}

A morphological opening is the composition of an erosion followed by a dilation: $O(\varphi) = \delta(\varepsilon(\varphi))$.
Computing $O(\varphi)$ is not an easy task. If $\varphi$ is in CNF, then $\delta(\varphi)$ is computable in polynomial time, and expressible as a polysize CNF, but then $\delta(\varepsilon(\varphi))$ is not (and can be exponentially long). If $\varphi$ is in DNF, then $\varepsilon(\varphi)$ is not polynomially computable (and can be exponentially long).
Proposition \ref{lemmasep} gives a hint on how to compute $O(\varphi)$, when $\varphi$ is under DNF.

\begin{proposition}\label{lemmasep2Open}
Let $\varphi_1,\ldots, \varphi_p$ the connected components of $\varphi$. Then we have:
$O(\varphi) \equiv \bigvee_{i=1}^p O(\varphi_i)$.
\end{proposition}

This results directly follows from Proposition~\ref{lemmasep2Ero}.

\medskip

Let us now consider the skeleton $Sk(\varphi)$. It is defined as the centers of maximal balls of the Hamming distance included in $\varphi$. In the finite discrete case, it
can be computed by the following algorithm:

\Omit{
\begin{tabbing}
{\tt begin}\\
$\varepsilon := \bot$;\\
{\tt While} $\varphi \not\equiv \bot$ {\tt do}\\
\hspace{1cm} \= $\psi := O(\varphi) (= D(E(\varphi))$;\\
\> {\tt if} $\psi \not\equiv \varphi$ \\
\> {\tt then} \\
\> \hspace{1cm} \= $\varepsilon := \varepsilon \vee (\varphi \wedge \neg \psi)$;\\
\>\> $\varphi := \psi$\\
\> {\tt end if}\\
{\tt end while}\\
{\tt Return $\varepsilon$}
\end{tabbing}

or, in a more concise and equivalent way:
}

\begin{tabbing}
{\tt begin}\\
$Sk(\varphi) := \varphi \wedge \neg O(\varphi)$; $\psi = \varphi$\\
{\tt While} $\psi \not\equiv \bot$ {\tt do}\\
\hspace{1cm} \= $Sk(\varphi) := Sk(\varphi) \vee (\varepsilon(\psi) \wedge \neg O(\varepsilon(\psi)))$;\\
\> $\psi := \varepsilon(\psi)$\\
{\tt end while}\\
{\tt Return $Sk(\varphi)$}
\end{tabbing}

We note that the number of iterations performed by this algorithm is equal to $\min \{ i, \varepsilon^i(\varphi) \equiv \bot\}$ and therefore is no larger than $N$.

\begin{example}~

Let us consider again $\varphi = (a \wedge b) \vee (a \wedge c) \vee (b \wedge c) \vee (\neg a \wedge \neg b \wedge \neg c)$, as in Figure~\ref{fig:exCC}. We have:
\begin{itemize}
\item $O(\varphi) = (a \wedge b) \vee (a \wedge c) \vee (b \wedge c)$ and $\varphi \wedge \neg O(\varphi) = (\neg a \wedge \neg b \wedge \neg c)$ which is the center of a maximal ball of radius 0;
\item $\varepsilon(\varphi) = a \wedge b \wedge c$, $O(\varepsilon(\varphi)) = \bot$, and $\varepsilon(\varphi) \wedge \neg O(\varepsilon(\varphi)) = a \wedge b \wedge c$, which is the center of a maximal ball of radius 1;
\item the next erosion provides $\bot$, so we stop here and return $Sk(\varphi) = (\neg a \wedge \neg b \wedge \neg c) \vee (a \wedge b \wedge c)$.
\end{itemize}
This is illustrated in Figure~\ref{fig:skel}.

\begin{figure}[htbp]
\centerline{\hbox{
\includegraphics[height=3.5cm]{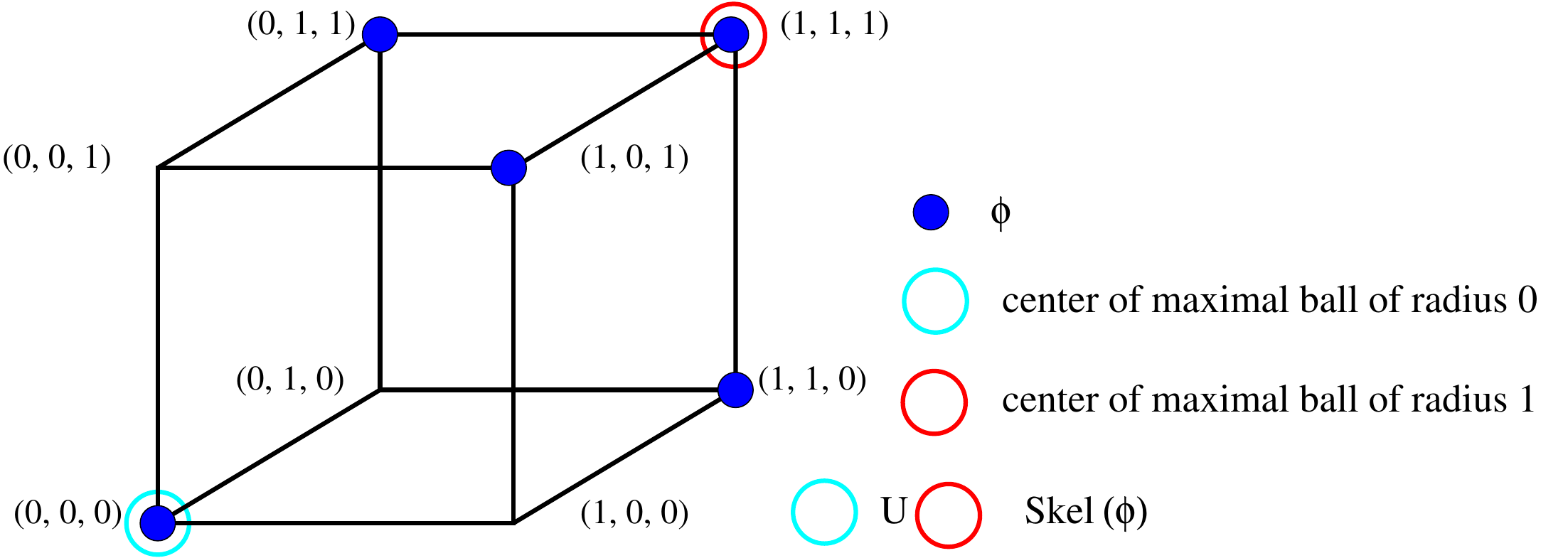}
}}
\caption{$Skel(\varphi)$: it is composed of the centers of maximal balls of radius 0 and 1.}
\label{fig:skel}
\end{figure}
\end{example}

We see that computing $Sk(\varphi)$ heavily relies on computing $O(\varphi)$.
Using the previous results on erosions and openings, we have:

\begin{proposition}\label{lemmasep2}
Let $\varphi_1,\ldots, \varphi_p$ the connected components of $\varphi$. Then we have:
$Sk(\varphi) \equiv \bigvee_{i=1}^p Sk(\varphi_i)$.
\end{proposition}

\section{Belief revision}
\label{sec:Revision}

In this section, we briefly survey some existing revision operators, and show that they can be equivalently expressed using morphological dilations. This establishes a first link between the proposed morpho-logic formalism and some reasoning tools developed for addressing aspects of knowledge dynamics. The morphological expressions will prove useful in Section~\ref{sec:unify} when proposing a unified framework for several reasoning tasks, using both erosions and dilations, and exploiting the morphological ordering introduced in Section~\ref{sec:ordering}.

%\subsection{Revision}

%\fbox{TODO: mieux montrer les etapes de la construction ?}

We start with some basics about belief revision. The aim of belief revision is to model how to incorporate in a coherent way a piece of information to a corpus of beliefs. In the most studied model, the AGM model \cite{AGM-85}, the corpus of beliefs is represented by a logical theory $K$ and the (new) piece of information by a formula $\psi$. The result of incorporating $\psi$ to $K$, i.e. the revision of $K$ by $\psi$, is denoted by $K\star\psi$. We give here a very simple presentation of this model in finite propositional logic due to Katsuno and Mendelzon \cite{KatsunoMendelzon91} in which the (old) beliefs $K$ are indeed represented by a formula $\varphi$ (that is, $K=Cn(\varphi)$) and the revision of $\varphi$ by $\psi$ is denoted $\varphi\circ\psi$. Note that $\circ$ is a function mapping a couple of formulas into a formula. This kind of function is called a revision operator\footnote{It is is easy to see that we can define an AGM operator $\star$ starting from $\circ$, by letting $K\star\psi=Cn(\varphi\circ\psi)$ where $\varphi$ satisfies $K=Cn(\varphi)$.} when it satisfies the following rationality postulates:
\\[1mm]
\mypost{R1} {$\varphi  \circ   \psi \vdash \psi$}{Success}
\mypost{R2} {If $\varphi \wedge \psi \nvdash \bot$ then $\varphi  \circ  \psi \equiv \varphi \wedge \psi$}{Minimality}
\mypost{R3} {If $\psi \nvdash \bot$ then $\varphi  \circ  \psi \nvdash \bot$}{Coherence}
\mypost{R4} {If $\varphi_1 \equiv \varphi_2$ and $\psi_1 \equiv \psi_2$ then $\varphi_1  \circ  \psi_1 \equiv \varphi_2  \circ  \psi_2$}{Syntax independence}
\mypost{R5} {$(\varphi  \circ  \psi) \wedge \theta \vdash \varphi  \circ  (\psi \wedge \theta)$}{Superexpansion}
\mypost{R6} {If  $(\varphi  \circ  \psi) \wedge \theta \nvdash \bot$ then $\varphi  \circ  (\psi \wedge \theta) \vdash (\varphi  \circ  \psi) \wedge \theta$}{Subexpansion}

A very powerful tool in order to construct revision operators is the representation theorem \cite{KatsunoMendelzon91}, based on the notion of faithful assignment.
A {\em faithful assignment}
 is a mapping which associates to each formula
$\varphi$ a total pre-order $\leq_{\varphi}$ on $\Omega$ such that the following conditions hold:\\[1mm]
\mypostbis{1}{if $\omega \models \varphi$ and $\omega' \models \varphi$ then
%$I \sim_{\varphi} J$;
$\omega \sim_{\varphi} \omega'$;}
\mypostbis{2} {if $\omega \models \varphi$ and
$\omega' \models \neg \varphi$ then
%$I <_{\varphi} J$;
$\omega <_{\varphi} \omega'$;}{}
\mypostbis{3} {if $\models \varphi_1 \leftrightarrow \varphi_2$ then
$\leq_{\varphi_1} = \leq_{\varphi_2}$.}{}

The representation theorem proven by Katsuno and Mendelzon~\cite{KatsunoMendelzon91} is the following one:
\begin{theo}\label{thm-repre}
An operator $\circ$ is a revision operator $\circ$, i.e. that satisfies {\bf R1-R6},
 iff there exists a faithful assignment
that maps each formula $\varphi$ to a total pre-order $\leq_{\varphi}$
such that for every propositional formula $\psi$ we have\footnote{The notation $\min(A,\leq)$ where $\leq$ is a total pre-order,
stands for $\{\omega \in A \mid \forall \omega' \in A, \
\omega \leq \omega'\}$.}
$$\llbracket \varphi \circ \psi \rrbracket = \min( \llbracket \psi \rrbracket ,\leq_{\varphi})$$%
\end{theo}

Intuitively, the pre-order $\leq_{\varphi}$ is  a qualitative way to express the distance of a world $\omega$
to $\varphi$, \ie $\omega \leq_{\varphi} \omega'$ means
that $\omega$ is closer to $\varphi$ than $\omega'$. Actually,
a faithful assignment can be defined from a distance $d$ from a world to a formula
in the following way: $\omega \leq_{\varphi} \omega'$ iff
$d(\omega,\varphi) \leq d(\omega',\varphi)$, where  $d(\omega,\varphi)$ is defined as
$\min\Set{d(\omega,\omega'') \mid \omega''\models \varphi}$. In particular, the revision
operator induced by the choice of the distance $d_H$ is known as Dalal's
revision operator.

\medskip

Now, let us consider the morphological dilation $\delta$ defined using as structuring element the ball of radius one of the distance $d$. It can be easily seen that we have
\[
\varphi \circ \psi = \delta^n(\varphi) \wedge \psi,
\]
with
$n = \min \{ k \in \mathbb{N} \mid \delta^k(\varphi) \wedge \psi
\mbox{ is consistent\}}$.

This approach is very natural since it corresponds to a principle of minimal change.
The following example illustrates in a precise manner the behavior of this operator.

\begin{example}[Revision]\label{linda}
John knew Linda\footnote{This story is inspired by a famous example in Cognitive Psychology  of an experiment by Tversky  and Kahneman
 \cite{TK83}. } when both of them were PhD students in  Philosophy in a very prestigious university. He remembers Linda's activism in feminism, her brilliant record and her great beauty.
Both obtained their PhD degree at the same time. Since then, five years after, John has no news from Linda. However, he thinks that Linda is for sure an activist
 in feminism, that she occupies  an excellent position in a Philosophy Department of one prestigious university and she maintains her beauty. John meets Peter, a common classmate, who
 says him that, surprisingly, Linda is now a bank teller. With this new piece of information John revises his beliefs and he thinks now that Linda is a bank teller
 who keeps her feminist activism and keeps her beauty.

 In this problem we code by the atoms $a$, $b$ and $c$ the facts Linda is a feminist activist, Linda is beautiful and Linda is a Professor respectively, and by $\neg c$ the fact that Linda is not a
 Professor (for instance the fact that Linda is a bank teller). The formula $\varphi:=a\wedge b\wedge c$ codes the beliefs of the agent (John) and the formula $\psi:= \neg c$ codes the new information. Then, following the previous definition of the revision operator $\circ$, we have $\varphi\circ\psi= \delta^1(\varphi)\wedge \neg c$. That is  because
 $\varphi\wedge\psi$ is inconsistent  and $\delta^1(\varphi)\wedge\psi$ is consistent. We have $\delta^1(\varphi)\wedge \neg c=a\wedge b\wedge \neg c$, that is
  Linda keeps her feminist activism, her beauty and she is a bank teller.
\end{example}

\begin{figure}[htbp]
\centerline{\hbox{
\includegraphics[height=3.5cm]{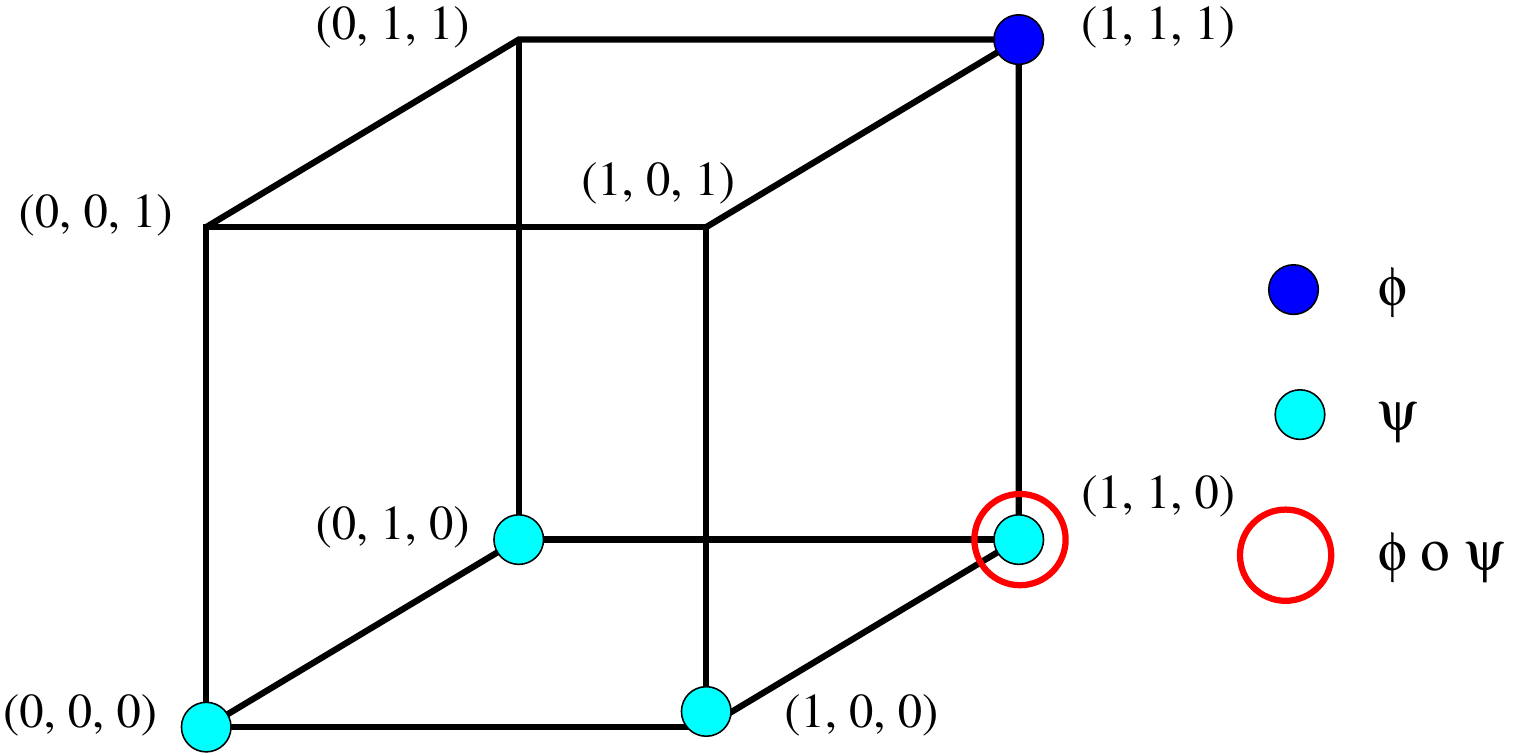}
}}
\caption{Example of revision $\varphi \circ \psi$, obtained here for a dilation of size $n=1$.}
\label{fig:revision}
\end{figure}

This example is illustrated in Figure~\ref{fig:revision}, using the same conventions as in Section~\ref{sec:MM}.

It is important to point out that within the previous approach, using as structuring element the standard ball of radius 1 (with respect to the Hamming distance in the example), there always exists $n$ such that  $\delta^n(\varphi)\equiv \top$ (when $\varphi$ is consistent).
This is essentialy the reason why  $\varphi \circ \psi$ is consistent when $\varphi$ is consistent. Also it is the reason why the so called {\em success postulate} in belief revision ($\varphi \circ \psi\vdash \psi$)  holds.

We have also remarked that there are some cases (with special structuring elements) in which we have a fixed point for the dilation, which is not necessary $\top$. For instance, we can have $\varphi$ and $n$ such that $\delta^n(\varphi)=\delta^{n+1}(\varphi)$ and $\delta^n(\varphi)\not\equiv\top$. What is interesting is that even in such a case we can define interesting and more general revision operators, namely credibility-limited revision operators \cite{BFKP12,HFCF01}. The precise way to do that is as follows:

\[
\varphi \circ \psi=
\left\{
\begin{array}{ccc}
  \delta^n(\varphi) \wedge \psi & & \mbox{where }
n = \min \{ k \in \mathbb{N} \mid \delta^k(\varphi) \wedge \psi
\mbox{ is consistent\}}\\
\varphi & & \mbox{if there is no $k$ such that }  \delta^k(\varphi) \wedge \psi \not\vdash \bot
\end{array}
\right.
\]

What is interesting to note is that in this general case, we can encode the credible worlds (see \cite{BFKP12}) as
$\mymod{ \delta^n(\varphi)}$, where $n$ is the least integer such that $\delta^n(\varphi)=\delta^{n+1}(\varphi)$.

Let us now consider the more general case, where $\delta$ is not necessarily a dilation defined from a distance. We have the following result:
%, the proof of which is straightforward.

\begin{proposition}
Let $\delta$ be an extensive and exhaustive operator (i.e. satisfying the following fillingness property: $\forall \varphi, \exists n \in \mathbb{N}, \delta^n(\varphi) \equiv \top$) on the lattice of propositional formulas. Then the operator $\circ$ defined by:
\[
\forall \varphi, \psi, \varphi \circ \psi = \delta^n(\varphi) \wedge \psi
\]
with $n = \min \{ k \in \mathbb{N} \mid \delta^k(\varphi) \wedge \psi \mbox{ is consistent} \}$ (the existence of $n$ is guaranteed by the fillingness property), $\delta^0(\varphi) = \varphi$ and $\delta^k(\varphi) = \delta(\delta^{k-1}(\varphi))$ for $k \geq 1$, is a revision operator satisfying the postulates R1-R6.
\end{proposition}

The proof of the previous proposition is based on Theorem \ref{thm-repre}. Actually, the mapping which associates $\varphi$ to $\leq_\varphi$ defined by:
\[
\forall \omega, \omega', \omega \leq_\varphi \omega' \Leftrightarrow \forall n \in \mathbb{N}, \omega' \in \mymod{\delta^n(\varphi)} \Rightarrow \omega \in \mymod{\delta^n(\varphi)}
\]
is a faithful assignment and it is not hard to see that for all $\psi$, $\mymod{\varphi \circ \psi} = \min( \mymod{\psi}, \leq_\varphi)$,
which by Theorem~\ref{thm-repre} says that $\circ$ is a revision operator.

Typically, $\delta$ can be any extensive and exhaustive dilation, but this proposition is slightly more general since it does not require $\delta$ to commute with the supremum, nor to be increasing.

The minimality property of revision operators has been widely discussed in the literature (see e.g.~\cite{KMP10,RW14,RWFA13}). Although it is not easy to define in any context in a general way, let us note that, in the particular case of propositional logic, the proposed morphological definition of revision provides a natural way to achieve this minimality in the sense that the set of models is minimally enlarged, which corresponds to the meaning of minimal change in~\cite{KatsunoMendelzon91}. The proposed approach also provides sound and precise tools to compute minimal revisions.

%%%%%%%%%%%%%%%%%%%%%%%%%%%%%%%%%%%%%%%%%%%%%%%%%%%%%%%%%%%%%
%%%%%%%%%%%%%%%%%%%%%%%%%%%%%%%%%%%%%%%%%%%%%%%%%%%%%%%%%%%%%

%\input{fusion}

\section{Belief merging}
\label{sec:Fusion}

In this section, we briefly survey some existing belief merging operators, and show the link with  morphological dilations.

%\fbox{TODO : pourrait \^etre mieux organis\'e ?}

We now recall some basics about belief merging\footnote{In knowledge dynamics the fusion of pieces of information having a logical representation is usually called belief merging \cite{KONI-98,KONI-02,KP11}.}.
%as defined by Konieczny and Pino-P\'erez \cite{KONI-98}.
Belief merging   %\cite{Rev93,Rev97,BKM91,BKMS92,Lin94,Lin95,Lin96,LibSha97,KONI-98,KONI-02,KP11}
\cite{KONI-98,KONI-02,KP11}
aims at  combining several pieces of information when there is no strict precedence between them. The agent faces several conflicting pieces of information coming from several sources of equal reliability\footnote{Actually the sources can have different reliabilities, but we will focus on the case where all the sources have the same reliability; there is already a lot to say in this case.}, and he has to build a coherent description of the world from them.

More precisely the inputs of a merging problem are a profile $\myPhi=\Set{\varphi_1,\dots,\varphi_n}$, defined as a multi-set of propositional formulas encoding
the different sources of information, and the integrity constraints encoded by a propositional formula $\mu$. The result  of merging $\myPhi$ under the constraint $\mu$ is a propositional formula which will be denoted $\arb_{\IC}(\E)$ (when $\mu\equiv\top$, we will write simply $\arb(\E)$ instead of $\arb_{\top}(\E)$). Thus, the  merging model is based on the study and construction of well behaved functions $\Delta$
mapping a couple  $(\E,\IC)$ into a formula $\arb_{\IC}(\E)$. Such functions are called merging operators. More precisely, an integrity constraint merging operator  (an IC merging operator for short) is a  function $\arb$ satisfying the following  rationality postulates:\\[2mm]
\mypostbis{IC0}{$\arb_{\IC}(\E) \vdash \IC$}
\mypostbis{IC1}{If $\IC $ is consistent, then $\arb_{\IC}(\E) $ is consistent}
\mypostbis{IC2}{If $\bigwedge \E$ is consistent with $\IC$,  then  $\arb_{\IC}(\E) \fequiv \bigwedge \E \land \IC$}
\mypostbis{IC3}{If $\E_1 \fequiv \E_2$ and $\IC_1 \fequiv \IC_2$, then  $\arb_{\IC_1}(\E_1) \fequiv  \arb_{\IC_2}(\E_2)$}
\mypostbis{IC4}{If $\K_1 \vdash \IC$ and $\K_2 \vdash \IC$, then $\arb_{\IC}(\{\K_1, \K_2\}) \land \K_1 $ is consistent if and only\\
\hspace*{1.2cm}if $\arb_{\IC}(\{\K_1, \K_2\}) \land \K_2 $ is consistent}
\mypostbis{IC5}{$\arb_{\IC}(\E_1) \land \arb_{\IC}(\E_2) \vdash \arb_{\IC}(\E_1 \sqcup \E_2)$}
\mypostbis{IC6}{If $\arb_{\IC}(\E_1) \land \arb_{\IC}(\E_2)$ is consistent, then  $\arb_{\IC}(\E_1 \sqcup \E_2) \vdash
\arb_{\IC}(\E_1) \land \arb_{\IC}(\E_2)$}
\mypostbis{IC7}{$\arb_{\IC_1}(\E) \land \IC_2 \vdash \arb_{\IC_1 \land \IC_2}(\E)$}
\mypostbis{IC8}{If $\arb_{\IC_1}(\E) \land \IC_2 $ is consistent, then  $\arb_{\IC_1 \land \IC_2}(\E)
\vdash \arb_{\IC_1}(\E)$}

\vspace{-5pt}
\noindent where $\bigwedge \E$ denotes the conjunction of all the formulas of $\E$; $\E_1 \fequiv \E_2$ means that there is a bijection $f$ from $\E_1$ into
$\E_2$ such that for any formula $\varphi\in\E_1$, we have $\varphi\equiv f(\varphi)$ (in particular, $\E_1$ and $\E_2$ have the same cardinality as multisets); the symbol $\sqcup$ stands for the multiset union.

For a detailed explanation of these postulates, see \cite{KONI-02}. However, let us make a comment  about Postulate  (IC4), known as the fairness postulate. As a matter of fact, this is a very restrictive postulate. Indeed, the only operators satisfying all the postulates are the operators built from distance and aggregation functions (see \cite{KP11}). Very natural operators fail to satisfy (IC4). In Section 5.2 of \cite{KONI-02} there are interesting results around this problem.

 An operator $\arb$ is called an {\em IC quasi-merging operator} if it satisfies all the previous postulates except (IC6), but instead of this postulate it satisfies the following one:\\[1mm]
\mypostbis{IC6'}{If $\arb_{\IC}(\E_1) \land \arb_{\IC}(\E_2)$ is consistent, then  $\arb_{\IC}(\E_1 \sqcup \E_2) \vdash
\arb_{\IC}(\E_1) \vee \arb_{\IC}(\E_2)$}

  \vspace{-5pt}

In order to establish a representation theorem we need to introduce the notion of \emph{syncretic assignment}. This is a function mapping each profile $\E$ to a total pre-order $\leq_{\E}$ over interpretations such that for any profiles $\E, \E_1, \E_2$ and for any belief bases $\K,\K'$   the following conditions hold:\\[1mm]
\mypostbis{1}{If $\I \models \E$ and $\J \models \E$, then $\I \simeq_{\E} \J$ }
\mypostbis{2}{If $\I \models \E$ and $\J \not\models \E$, then $\I <_{\E} \J$ }
\mypostbis{3}{If $\E_1 \fequiv \E_2$, then $\leq_{\E_1}=\leq_{\E_2}$ }
\mypostbis{4}{$\forall \I \models \K $ $\exists \J \models \K'$ $\J \leq_{\K \sqcup \K'} \I$ }
\mypostbis{5}{If $\I \leq_{\E_1} \J$ and $\I \leq_{\E_2} \J$, then $\I \leq_{\E_1 \sqcup \E_2} \J$ }
\mypostbis{6}{If $\I <_{\E_1} \J$ and $\I \leq_{\E_2} \J$, then $\I <_{\E_1 \sqcup \E_2} \J$}
\vspace{-5pt}

When the condition (6) is replaced by
the following condition\\[1mm]
\mypostbis{6'}{If $\I <_{\E_1} \J$ and $\I <_{\E_2} \J$, then $\I <_{\E_1 \sqcup \E_2} \J$}

\vspace{-5pt}

\noindent the assignment is called a {\em quasi-syncretic assignment}, that is a function mapping each profile $\E$ to a total pre-order $\leq_{\E}$ over interpretations satisfying (1)-(5) and (6').

Now we can state the following representation theorem for merging operators:

\begin{theo}[\cite{KONI-02}]
\label{thm-repre-fus}
An operator $\arb$  is an IC merging operator
(or IC quasi-merging operator respectively)
if
and only if there exists a syncretic assignment
(or quasi-syncretic assignment respectively)
that maps each
profile $\E$ to a total pre-order $\leq_\E$ such that
$$\mod{\arbic(\E)} =\min(\mod{\IC},\leq_{\E})$$
%When this equation holds we will say that the assignment represents the operator.
\end{theo}

A very useful technique to build such operators is based on a distance (actually a pseudo-distance) between interpretations and
a numerical aggregation function. We describe how this works more precisely in what follows.

A {\em pseudo-distance}\footnote {The triangle inequality is not required.} between interpretations is a function
  $d: \Omega \times \Omega \mapsto
\rit^{+}$ such that for any $\omega$, $\omega'$ $\in \Omega$:
 $d(\omega,\omega')=d(\omega',\omega)$, and
 $d(\omega,\omega')=0$ iff $\omega=\omega'$.

An {aggregation function} $f$ is a  function  mapping for any positive integer $n$, each n-tuple of non negative reals into a positive real such that
for any
 $x_1, \ldots, x_n, x,$ $y \in \rit^{+}$:\\
%\triche
\mypostitem{$\bullet$}{   if $x \leq y$, then
  $f(x_1,\ldots,x,\ldots,x_n) \leq f(x_1,\ldots,y,\ldots,x_n)$}  %\\ \null \hfill \null
  {monotony}
\mypostitem{$\bullet$}{   $f(x_1, \ldots, x_n) = 0$ iff  $x_1 = \ldots = x_n = 0$}%  \null \hfill \null
{minimality}
\mypostitem{$\bullet$}{    $f(x) = x$}{identity}

\vspace{-5pt}

With the help of $d$ and  $f$, a distance between interpretations and an aggregation function respectively, we can construct a total pre-order $\leq_\E$ on interpretations associated with $\E=\Set{\varphi_1, ..., \varphi_n}$ in the following way. First, remember that $d(\I,\K)$ is $ \min_{\J \models \K} d(\I,\J)$. Then, define $d(\omega,\E)=f(d(\omega,\K_1)\dots,d(\omega,\K_n))$. Finally, $\omega \leq_{\E} \omega' $ iff
$ d(\omega,\E) \leq d(\omega',\E)$. This process is, actually, an assignment which is in fact a syncretic (or a quasi-syncretic) assignment when the aggregation
function has good additional properties such as symmetry, composition and decomposition (see \cite{KP11}). For instance when $f$ is the function {\em sum} or {\em leximin}, we obtain a syncretic assignment by the previous process. When $f$ is the function {\em max}, we obtain a quasi-syncretic assignment. Thus, in virtue of Theorem \ref{thm-repre-fus}, the operator defined by the equation
$\mod{\arbic(\E)} =\min(\mod{\IC},\leq_{\E})$ is an IC merging operator when the aggregation function used is the sum or leximin (Gmax) and is an IC quasi-merging operator when  the aggregation function used is the max. They are called in the literature $\arb^\Sigma$, $\arb^{Gmax}$ and $\arb^{max}$
respectively\footnote{Strictly, they are called $\arb^{d,\Sigma}$, $\arb^{d,Gmax}$ and $\arb^{d,max}$ respectively, to emphasize the chosen distance $d$.}.

\medskip

Let us now establish the links with dilations. Again we consider a dilation $\delta$ defined using the balls of the distance $d$ as structuring elements.
Then it is not hard to see the following:
\begin{equation}\label{eq:max}
\Delta^{max}_\mu(\varphi_1, ..., \varphi_m)
= \delta^n(\varphi_1) \wedge \delta^n(\varphi_2) \wedge... \wedge  \delta^n(\varphi_m)\wedge\mu,
\end{equation}
where $n = \min \{ k \in \mathbb{N} \mid \delta^k(\varphi_1) \wedge ... \wedge
\delta^k(\varphi_m)\wedge\mu \mbox{ is consistent\}}$.
 \begin{equation}\label{eq:sigma}
\Delta^{\Sigma}_\mu(\varphi_1, ..., \varphi_m)
= \bigvee_{(n_1,\dots,n_m)}\delta^{n_1}(\varphi_1) \wedge \delta^{n_2}(\varphi_2) \wedge... \wedge  \delta^{n_m}(\varphi_m)\wedge\mu,
\end{equation}
where the values $n_1,\dots,n_m$ are such that $\sum_{i=1}^m n_i$  is minimal with
$\delta^{n_1}(\varphi_1) \wedge \delta^{n_2}(\varphi_2) \wedge... \wedge  \delta^{n_m}(\varphi_m)\wedge\mu$ consistent.

\begin{figure}[htbp]
\centerline{\hbox{
\includegraphics[height=3.5cm]{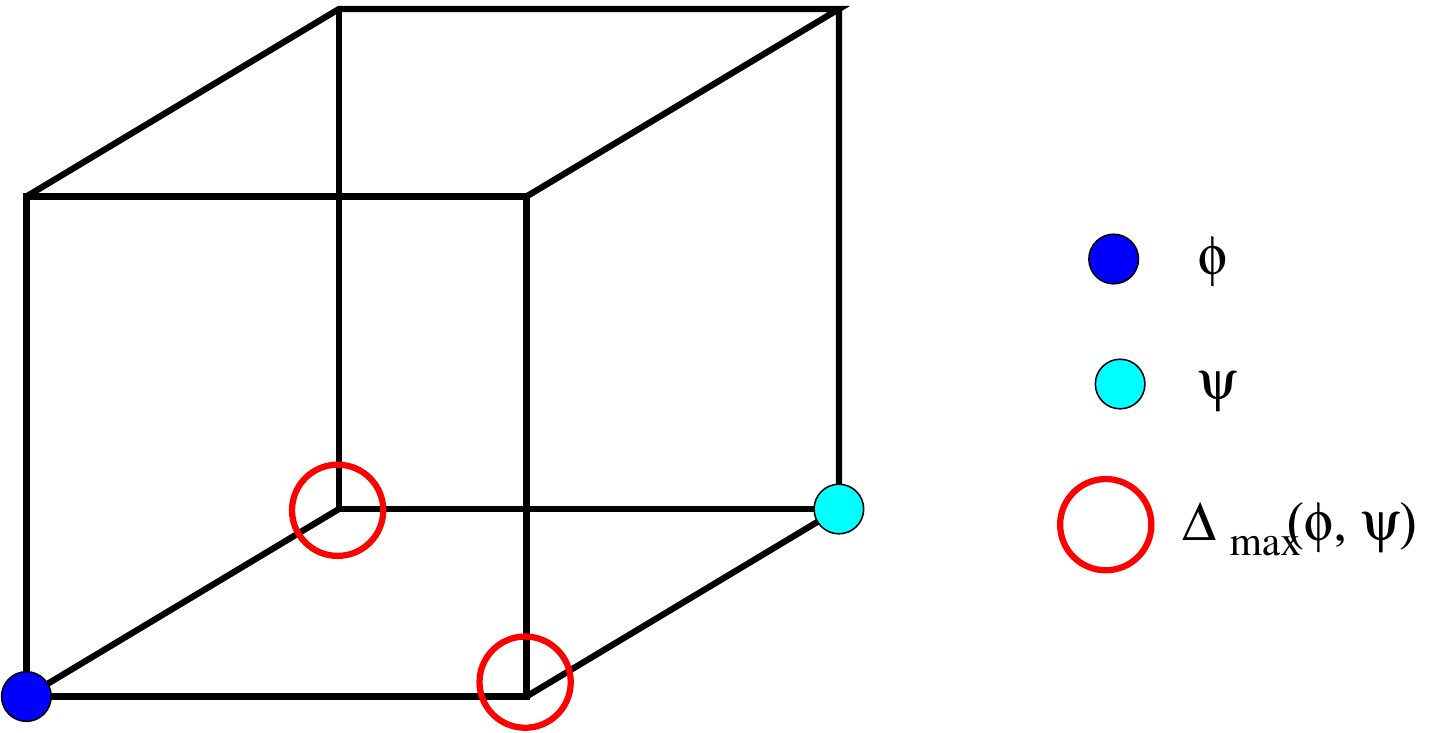}
}}
\caption{Example of fusion $\Delta^{max}(\varphi, \psi)$, obtained for a dilation of size $n=1$.}
\label{fig:fusion}
\end{figure}

An example illustrating the behavior of $\Delta^{\max}$ is displayed in Figure~\ref{fig:fusion}, with the same conventions as in Section~\ref{sec:MM} and the Hamming distance. Let us consider $\varphi = \neg a \wedge \neg b \wedge \neg c$, $\psi = a \wedge b \wedge \neg c$ and $\mu = \top$. While $\varphi \wedge \psi$ is not consistent, $\delta^1(\varphi) \wedge \delta^1 (\psi)$ is, and $\Delta^{\max}(\varphi, \psi) = \delta^1(\varphi) \wedge \delta^1(\psi) = (a \wedge \neg b \wedge \neg c) \vee (\neg a \wedge b \wedge \neg c)$ (i.e. the merging provides either $a$ or $b$, exclusively, and $\neg c$).

Next we give a less abstract example.

\begin{example}[Fusion]\label{travelers}
Let us consider two agents who want to travel together but have inconsistent preferences.
The set of propositional symbols is the set of all countries in the world. Preferences are denoted by formulas $\varphi$. In this example, we show how dilation can help reaching an agreement between agents.
Let us assume that Agent 1
prefers to travel in Spain: $\varphi_1 = \Spain$.
%(ii) has to stay in Europe: $\psi_1 = \neg (Belgium \vee France \vee \Spain \vee Portugal \vee Italy \vee Germany \vee The Netherlands \vee ... \}$.
On the other hand, Agent 2
prefers to travel in Morocco: $\varphi_2 = Morocco$.
%(ii) has to stay in a Mediterranean country: $\psi_2 = \neg (Morocco \vee \Spain \vee Italy \vee Portugal \vee ...)$.
Hence the two agents have conflicting preferences.
However, each agent is now ready to extend his preferences so that the two agents can travel together. This can be simply modeled by a dilation $\delta$, such that some neighbor countries are included in the preferences:
\[
\delta(\varphi_1) = \Spain \vee France \vee Portugal \vee Morocco
\]
\[
\delta(\varphi_2) = Morocco \vee Algeria \vee Portugal \vee \Spain
\]
Now the preferences are no more conflicting. The fusion of the agents' preferences, denoted $\Delta(\varphi_1,\varphi_2)$, can be expressed as the conjunction of the dilated preferences:
\[
\Delta(\varphi_1,\varphi_2) = \delta(\varphi_1) \wedge \delta(\varphi_2) = \Spain \vee Portugal \vee Morocco.
%( \varphi'_1 \wedge \varphi'_2, \psi_1 \vee \psi_2) = (\Spain \vee Portugal, %\neg (\bigvee Medit.\, and\, Eur.\, countries))
\]
A solution for traveling can then be found in the set of models of these formulas.

To go one step further, we can add constraints the agents have to satisfy. For instance if Agent 1 has to stay in Europe and Agent 2 has to stay in a Mediterranean country, these constraints can be taken into account by conditional dilations, thus modifying preferences as:
\[
\varphi'_1 = \delta(\varphi_1) \wedge \psi_1 = \Spain \vee France \vee Portugal,
\]
\[
\varphi'_2 = \delta(\varphi_2) \wedge \psi_2 = \delta(\varphi_2),
\]
where $\psi_1$ and $\psi_2$ encode the constraints.
Then the new set of consistent preferences is given by $\varphi' = \varphi'_1 \wedge \varphi'_2 = \Spain \vee Portugal$.

Now suppose that the integrity constraints are encoded by a formula $\mu$, which establishes the fact that one and only one country can be visited except Spain and Morocco.
In this case, the fusion of $\varphi_1$ and $\varphi_2$ under the constraint $\mu$, denoted $\Delta_\mu(\varphi_1,\varphi_2)$ is
exactly $\delta(\varphi_1)\wedge\delta(\varphi_2)\wedge\mu$, \ie $$\Delta_\mu(\varphi_1,\varphi_2)=Portugal$$
\end{example}

%\textcolor{red}{Other examples: ``directional'' dilation (along specific dimensions only); dilation specific to each agent -- j'en ajoute ?}

Equations \ref{eq:max} and \ref{eq:sigma} allow defining more general merging operators when $\delta $ is an extensive and exhaustive operator
congruent with logical equivalence, i.e. if $\varphi_1\equiv\varphi_2$ then $\delta(\varphi_1)\equiv\delta(\varphi_2)$. We are going also to consider the following symmetry property for $\delta$, related to the fairness postulate:
(IC4):\\[1.3mm]
\mypostbis{sym}{$\delta^n(\varphi)\wedge\varphi'\not\vdash\bot$ iff $\delta^n(\varphi')\wedge\varphi\not\vdash\bot$}

\vspace{-5pt}
\noindent In particular we have the following results:

\begin{proposition}
Let $\delta$ be an extensive and exhaustive operator which is congruent with logical equivalence on the lattice of propositional formulas. Then $\Delta_\mu^{\max}$ defined by:
\[
\Delta^{\max}_\mu(\varphi_1, ..., \varphi_m)
= \delta^n(\varphi_1) \wedge \delta^n(\varphi_2) \wedge... \wedge  \delta^n(\varphi_m)\wedge\mu,
\]
where $n = \min \{ k \in \mathbb{N} \mid \delta^k(\varphi_1) \wedge ... \wedge
\delta^k(\varphi_m)\wedge\mu \mbox{ is consistent\}}$ (the existence of $n$ being guaranteed by the fillingness property), is a merging operator
satisfying (IC1-IC3), (IC5), (IC6') and (IC7-IC8).
Moreover it satisfies (IC4) iff $\delta$ satisfies (sym). Thus, if $\delta$ is an extensive and exhaustive operator which is congruent with logical equivalence and satisfies (sym), the operator $\Delta^{\max}$ is an IC quasi-merging operator.
\end{proposition}

\beginproof Define $d(\omega,\varphi)=n$ where $n=\min \Set{k \mid \omega \in\mod{\delta^k(\varphi}}$.
This function $d$ is well defined because of exhaustivity of $\delta$.
Define $d(\omega,\myPhi)=\max(d(\omega,\varphi_1), \dots, d(\omega,\varphi_n)))$ where $\myPhi=\Set{\varphi_1,\dots,\varphi_n}$.
Now let $\omega\leq_\myPhi\omega'$ iff $d(\omega,\myPhi)\leq d(\omega',\myPhi)$.
Finally let $\Delta_\mu(\E)$ be a formula satisfying the following equation:
$\mod{\arbic(\E)} =\min(\mod{\IC},\leq_{\E})$. This is well defined because $\delta$ is congruent with logical equivalence.
It is easy to see that $\Delta^{\max}_\mu(\E)=\arbic(\E)$.
By the hypothesis about $\delta$ and the fact that the aggregation function taken is the max function, it is also easy to check that
the assignment $\E\mapsto \leq_\E$ is a quasi-syncretic assignment (property (4) is indeed equivalent to property (sym)).
Thus, by virtue of Theorem~\ref{thm-repre-fus}, $\Delta^{\max}$ is an IC quasi-merging operator.
\endproof

\begin{proposition}
Let $\delta$ be an extensive and exhaustive operator which is congruent with logical equivalence on the lattice of propositional formulas. Then $\Delta_\mu^{\Sigma}$ defined by:
\[
\Delta^{\Sigma}_\mu(\varphi_1, ..., \varphi_m)
= \bigvee_{(n_1,\dots,n_m)}\delta^{n_1}(\varphi_1) \wedge \delta^{n_2}(\varphi_2) \wedge... \wedge  \delta^{n_m}(\varphi_m)\wedge\mu,
\]
where the numbers $n_1,\dots,n_m$ are such that $\sum_i n_i$  is minimal with
$\delta^{n_1}(\varphi_1) \wedge \delta^{n_2}(\varphi_2) \wedge... \wedge  \delta^{n_m}(\varphi_m)\wedge\mu$ consistent, is a merging operator
satisfying (IC1-IC3), (IC5-IC8).
Moreover it satisfies (IC4) iff $\delta$ satisfies (sym). Thus, if $\delta$ is an extensive and exhaustive operator which is congruent with logical equivalence and satisfies (sym), the operator $\Delta^{\Sigma}$ is an IC merging operator.
\end{proposition}

\beginproof
Similar to the proof of the previous proposition but using the sum ($\Sigma$) function  instead of the max function.
\endproof

This approach has been extended in~\cite{Gorogiannis2008a} to first order logic, by combining dilation and comparison ordering operators. The merging postulates are then adapted, and conditions on these two operators are established in order to satisfy these postulates. An implementation using binary decision diagrams has furthermore been proposed in~\cite{Gorogiannis2008b}.

\section{Abduction}
\label{sec:abduction}
%Ramon: modified 05/12/2014

%\fbox{Carlos'version - January 2014 + modif in def of last consistent erosion to account for fixed point}
%
%\fbox{TO DO IN THIS SECTION:} add examples with other SE?
%
%\bigskip

The process  of inferring the best explanation of an observation
is usually known as {\em abduction}. In the logic-based approach
to abduction, the background theory is given by a consistent set
of formulas\footnote{Often in this work we will identify a finite set of formulas $\Sigma$ with
the conjunction of all its formulas and, by abuse of language, we continue to call this formula $\Sigma$. Thus, for instance,
we denote the conjunction of formulas of $\Sigma\cup\Set\alpha$ by $\Sigma\wedge\alpha$.}
 $\Sigma$. The notion of a {\em possible explanation}
is defined by saying that a formula $\gamma$ that is consistent with $\Sigma$ is an explanation of
$\alpha$ if $\Sigma\cup\{\gamma\}\vdash \alpha$ (this will be
written $\gamma\vdash_\Sigma\alpha$). An explanatory relation is a
binary relation $\expl{}{}$ where the intended meaning of
$\expl{\gamma}{\alpha}$ is ``$\gamma$ is a {\em preferred
explanation} of $\alpha$''.

In \cite{PINO-99}, a set of postulates that should be satisfied by
preferred explanatory relations was proposed and discussed.

The aim of this section is threefold: first, to propose very
natural explanatory relations using morphologic that in some cases are
computationally tractable; secondly, to examine the adequacy of
logical postulates proposed in \cite{PINO-99}, and thirdly, the
discovery of new logical properties for explanatory reasoning.

Morphologic allows us  to
define  the {\em most central part} of a formula, according to the
fundamental principles of this theory (see e.g.
\cite{SERR-82,SERR-88}, and Section~\ref{sec:MM}). Using this
notion we define two explanatory relations. The first one,
$\Elne{}{}$, has the following intended meaning: $\gamma$ is a
preferred explanation of $\alpha$ if every model of $\Sigma\cup
\{\gamma\}$ belongs to the {\em most central part} of $\Sigma\cup
\{\alpha\}$. For the second one, $\Elc{}{}$, we define a sequence
which approximates the most central part of $\Sigma$; then we say
that $\gamma$ is a preferred explanation of $\alpha$ if
$\gamma\vdash_\Sigma\alpha$ and moreover every model of $\Sigma\cup\Set\gamma$
is one of the  closest elements of the sequence which are  also
model of $\alpha$.

In this section, we mostly consider cases where $\Sigma \wedge \alpha \not\vdash \bot$.

\subsection{Explanatory relations based on erosion}
\label{sec:operators}

In this section we define precisely the
concept of {\em most central part} of a formula with the help of
the erosion operator. Then, based on this concept, we define two
explanatory relations.

\subsubsection{Using the last non-empty erosion}

In this section, we propose to exploit the idea of last erosion $\varepsilon_\ell(\varphi)$, as introduced in Definition~\ref{def:lastErosion}.

\begin{figure}[htbp]
\centerline{\hbox{
\includegraphics[height=3.5cm]{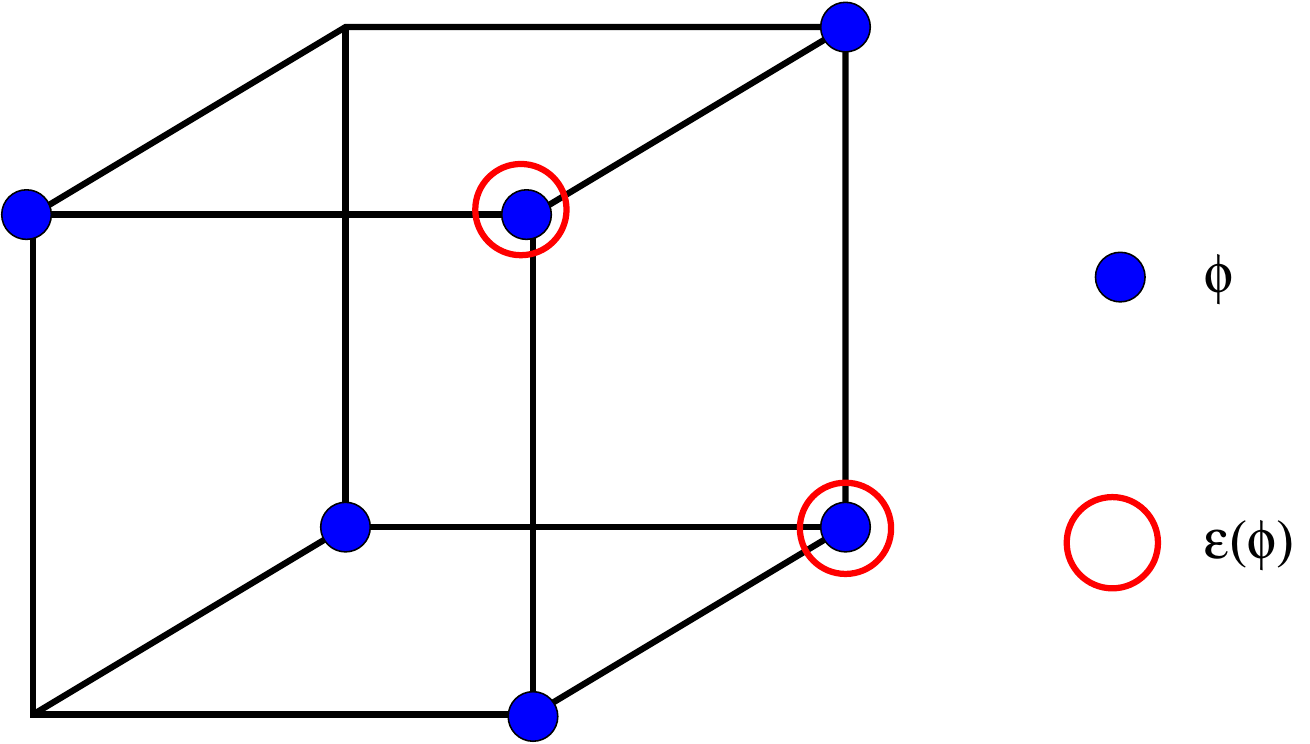}
}}
\caption{An example of $\varphi$ and its last erosion, equal to $\varepsilon(\varphi)$ in this case.}
\label{fig:last_erosion}
\end{figure}

Let us take (see Figure \ref{fig:last_erosion})
$\varphi = (a \vee \neg b \vee \neg c) \wedge (a \vee b \vee c)$, and an erosion defined using the balls of the Hamming distance as structuring elements.
Using the properties of erosion, and in particular the fact that it commutes with the conjunction, it is easy to derive:
\[
\varepsilon^1(\varphi) = (a \vee \neg b) \wedge (a \vee \neg c) \wedge (\neg b \vee \neg c) \wedge (a \vee b) \wedge (a \vee c) \wedge (b \vee c) = (a \wedge \neg b \wedge c) \vee (a \wedge b \wedge \neg c).
\]
Since $\varepsilon^2(\varphi) \vdash \bot$, we have
$\varepsilon^1(\varphi) = \varepsilon_\ell(\varphi)$ (its models
are in red in Figure~\ref{fig:last_erosion}).

A preferred explanation of $\alpha$ is then defined from this
operator applied on $\Sigma \wedge \alpha$, more precisely:

\begin{defi}
\label{def:explaLastEro}
The explanation relations derived from the last non-empty erosion are defined as follows:
\begin{equation}
\Elneu\gamma\alpha \iffdef \gamma \equiv_\Sigma \varepsilon_\ell(\Sigma \wedge
\alpha).
\label{eq:lasterosion1-unique}
\end{equation}
\begin{equation}
\Elned\gamma\alpha \iffdef \gamma \vdash_\Sigma \varepsilon_\ell(\Sigma \wedge
\alpha).
\label{eq:lasterosion1}
\end{equation}
\end{defi}

The idea of taking the last erosion of $\Sigma \wedge \alpha$ can
be interpreted in terms of robustness. An erosion of size $n$ of a
formula is a formula that can be changed while still proving the
initial formula. If at most $n$ symbols are changed in
$\varepsilon^n(\varphi)$ then $\varphi$ is always satisfied. Here,
considering $\varepsilon_\ell(\Sigma \wedge \alpha)$ means that we are
looking at the most reduced formula that satisfies $\Sigma \wedge
\alpha$, i.e. the one that can be changed the most while
satisfying $\Sigma \wedge \alpha$.

Taking $\equiv_\Sigma$ or $\vdash_\Sigma$ in Definition \ref{def:explaLastEro} is interesting because $\gamma$ could then have
models outside $\Sigma$, which may lead to more interesting
explanations from a syntactic point of view (note that the syntax
of $\Sigma$ is not taken into account in the proposed approach,
since all operations are performed on the models, at a semantic
level). However this may also add noise to the explanations. Two
possibilities can be suggested to limit this effect: (i) to use
$\equiv$ or $\vdash$, at the price of loosing meaningful explanations in some
cases from a syntactical point of view; (ii) to impose that
explanations have to be built from a user defined set of atoms.

It is interesting to note that using $\Elned{}{}$, we have for each $\gamma'$ such that $\gamma \wedge \gamma'$ is consistent $\Elned{\gamma \wedge \gamma'}\alpha$. Using $\Elneu{}{}$ avoids this very strong relations with conjunctions.

In the following we illustrate the behavior of $\Elned{}{}$ (similar illustrations can be provided for $\Elneu{}{}$). We denote by $PE_{\Elned{}{}}(\alpha) = \{ \gamma \mid \Elned\gamma\alpha \}$
the set of preferred explanations of $\alpha$. We can distinguish a subset of  $PE_{\Elned{}{}}(\alpha)$ that contains the  simpler (or purer)
preferred explanations of $\alpha$, denoted $PPE_{\Elned{}{}}(\alpha)$, defined by the following equation:
$$PPE_{\Elned{}{}}(\alpha)=\Set{\gamma \mid \gamma\vdash \varepsilon_\ell(\Sigma \wedge \alpha) \mbox{ and $\gamma$ is consistent}}$$

Actually, it is easy to see that the preferred explanations can be defined starting with the pure preferred explanations and adding a little noise. More precisely,
$PE_{\Elned{}{}}(\alpha) = \{ \gamma\vee\delta \mid \gamma\in PPE_{\Elned{}{}}(\alpha) \mbox{ and } \delta\in R  \}$, where
 $R$, the noise, is defined by
$R=\{  \delta \mid \delta\wedge\Sigma\vdash \bot   \}$.

Let us take $\Sigma=\Set{a\vee b\vee c}$ and $\alpha=\varphi$ where $\varphi$ is
defined as in the previous example (Figure
\ref{fig:last_erosion}). Note that $\Sigma \wedge \alpha = \varphi$. Thus, the pure preferred explanations of $\alpha$ are
%For Equation \ref{eq:lasterosion2}, we directly have $\gamma = (a \wedge \neg b \wedge c) \vee (a \wedge b \wedge \neg c)$.
% For Definition \ref{eq:lasterosion1}, if we denote
% $PE_{\Elne{}{}}(\alpha) = \{ \gamma \; : \; \Elne\gamma\alpha \}$
% (the preferred explanations of $\alpha$), we have:
\[
PPE_{\Elned{}{}}(\alpha) = \{ (a \wedge \neg b \wedge c), (a \wedge
b \wedge \neg c), (a \wedge \neg b \wedge c) \vee (a \wedge b
\wedge \neg c) \}.
\]

\begin{figure}[htbp]
\centerline{\hbox{
\includegraphics[height=3.5cm]{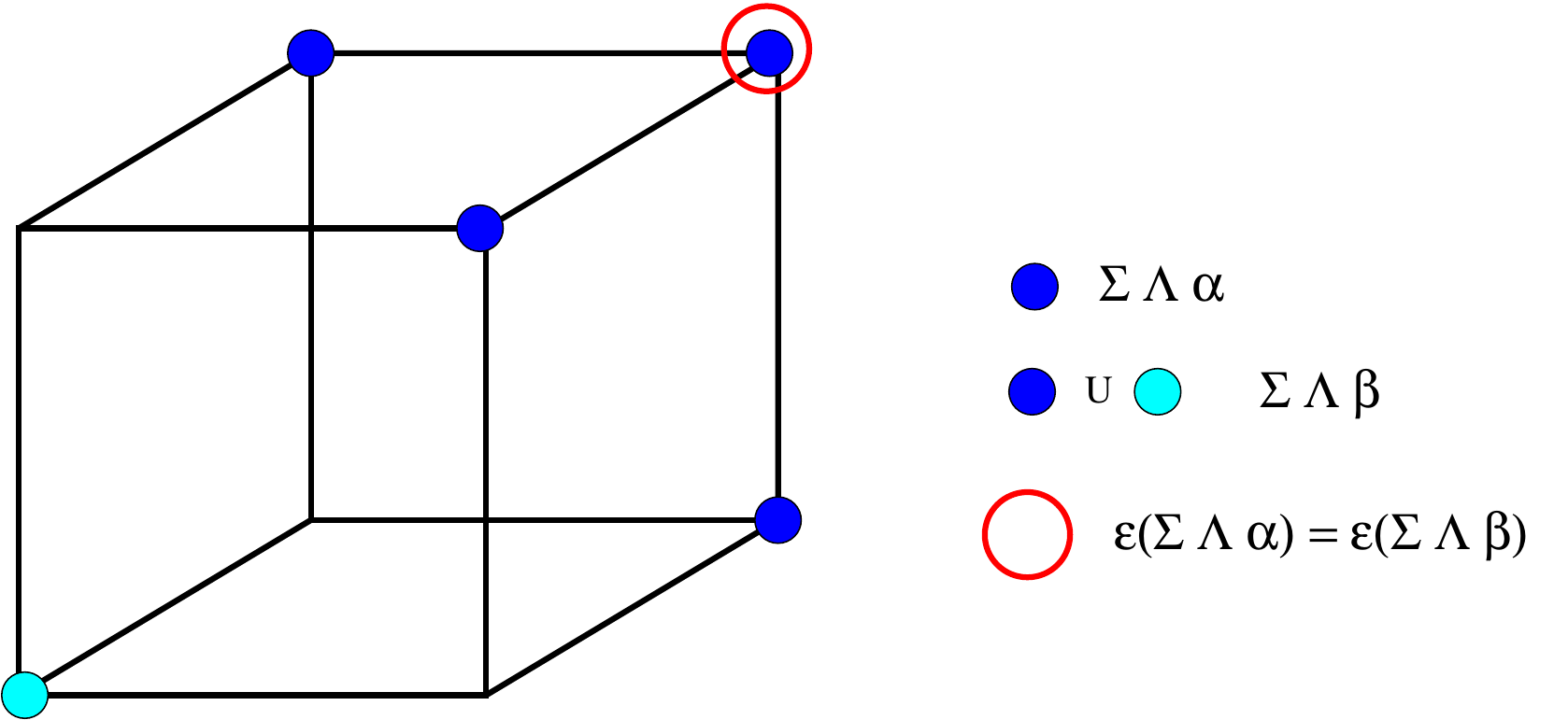}
}}
\caption{An example of $\Sigma \wedge \alpha$ and $\Sigma \wedge \beta$ that have the same last erosion. $\Sigma \wedge \beta$ has two connected components (blue models on the one hand and the cyan one on the other hand), the second one being not represented in the final result.}
\label{fig:last_erosion_2cc}
\end{figure}

Erosion does not take in account  all ``parts'' of a formula. Let
us take for instance: $\Sigma \wedge \alpha = (a \vee b) \wedge (a
\vee c) \wedge (b \vee c)$ and $\Sigma \wedge \beta = ((a \vee b)
\wedge (a \vee c) \wedge (b \vee c)) \vee (\neg a \wedge \neg b
\wedge \neg c)$ (Figure~\ref{fig:last_erosion_2cc}). Then we have:
$\varepsilon_\ell(\Sigma \wedge \alpha) = \varepsilon_\ell(\Sigma
\wedge \beta) = a \wedge b \wedge c$ and $PE_{\Elned{}{}}(\alpha) =
PE_{\Elned{}{}}(\beta)$ (as well as $PPE_{\Elned{}{}}(\alpha) =
PPE_{\Elned{}{}}(\beta)$). The set of worlds satisfying $\Sigma
\wedge \beta$ is disconnected, and the connected component
containing only $(\neg a \wedge \neg b \wedge \neg c)$ is not
represented in the explanations of $\beta$. This should not be
surprising, since any explanatory relation will select some part
of an observation as the most relevant one. However, if  this is
considered to be a problem, it can be avoided by considering the
ultimate erosion instead of the last erosion, which will select at
least one element of each connected component of an observation (see Section~\ref{sec:derivedOperators}).

\subsubsection{Using thew last consistent erosion}

Another idea consists in eroding $\Sigma$ as much as possible but
still under the constraint that it remains consistent with
$\alpha$:
\begin{equation}
\varepsilon_{\ell c}(\Sigma, \alpha)=\varepsilon^n(\Sigma)
\label{eq:LastCons}
\end{equation}
where
\[
\left\{
\begin{array}{ll}
n= \sup\{k \in \mathbb{N} \mid \varepsilon^k(\Sigma) \wedge \alpha \not\vdash \bot \} & \mbox{if } n < +\infty\\
n = \min \{k\in \mathbb{N} \mid \forall k'>k, \varepsilon^{k'}(\Sigma) = \varepsilon^k(\Sigma), \varepsilon^k(\Sigma) \wedge \alpha \not\vdash \bot \} & \mbox{otherwise}.
\end{array}
\right.
\]
From this operator, we define the following explanatory
relation:

\begin{defi}
\label{def:LastConsistentEro}
The explanation operator derived from the notion of last consistent erosion is defined as:
\begin{equation}
\Elc\gamma\alpha \iffdef \gamma \vdash_\Sigma \varepsilon_{\ell c}(\Sigma, \alpha)
\wedge \alpha.
\label{eq:consistenterosion1}
\end{equation}
\end{defi}

This definition has a different interpretation. Here we consider
erosion of $\Sigma$ alone, which means that we are looking at the
formulas that satisfy $\alpha$ while being the most in the theory,
i.e. that can be changed while remaining in the theory.
% (but not
%necessarily satisfying $\alpha$ after the changes).

As before, we denote $PE_{\Elc{}{}}(\alpha) = \{ \gamma \mid \Elc\gamma\alpha \}$
the set of preferred explanations of $\alpha$. We define the set of simpler (or purer)
preferred explanations of $\alpha$ (with respect to the relation $\Elc{}{}$), denoted $PPE_{\Elc{}{}}(\alpha)$, by the following equation:
$$PPE_{\Elc{}{}}(\alpha)=\Set{\gamma \mid \gamma\vdash \varepsilon_{\ell c} (\Sigma, \alpha)\wedge\alpha \mbox{ and $\gamma$ is consistent}}$$
Also, as in the case of last non-empty erosion, we have
$PE_{\Elc{}{}}(\alpha) = \{ \gamma\vee\delta \mid \gamma\in PPE_{\Elc{}{}}(\alpha) \mbox{ and } \delta\in R  \}$.

\begin{figure}[htbp]
\centerline{\hbox{
\includegraphics[height=3.5cm]{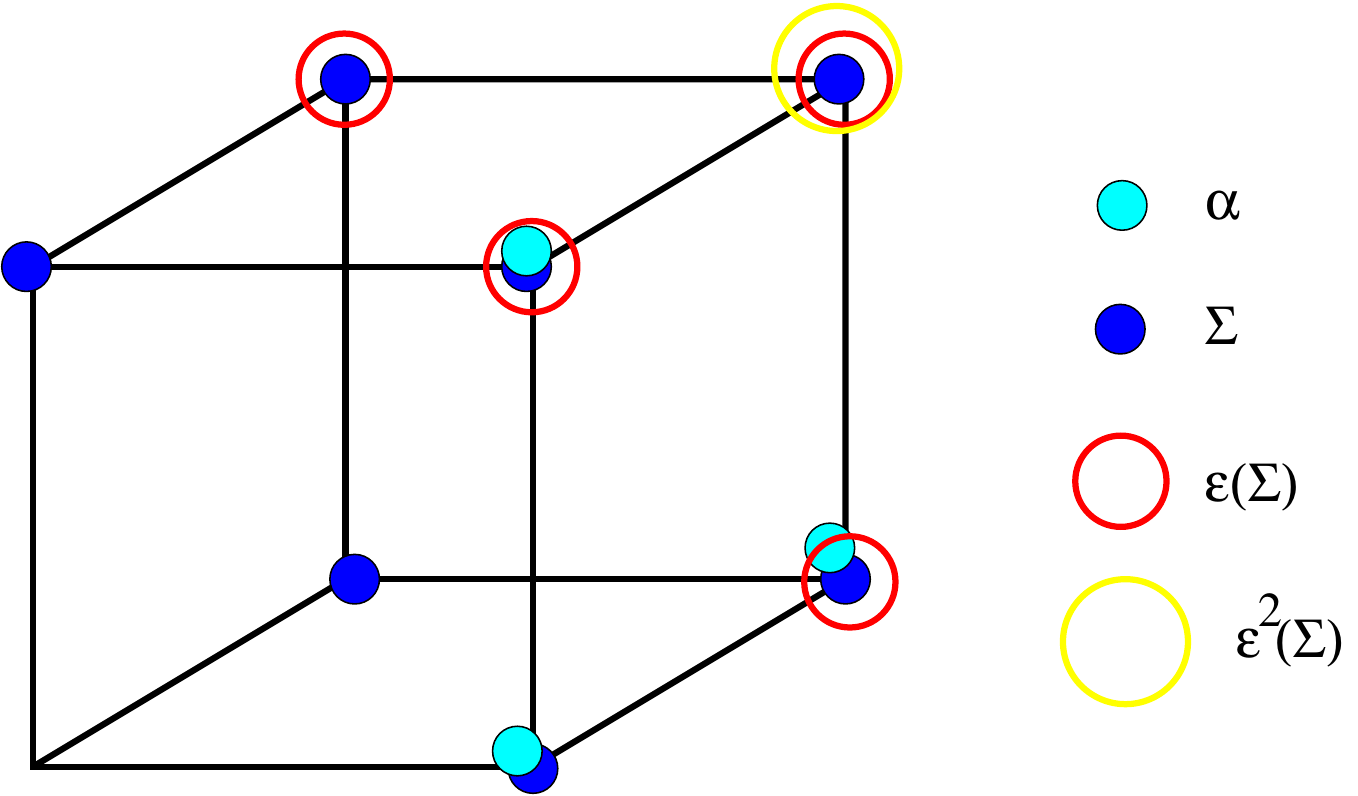}
}}
\caption{An example of last consistent erosion.}
\label{fig:consistent_erosion}
\end{figure}

Let us come back to the illustrative example, and take (see Figure
\ref{fig:consistent_erosion}):
$\Sigma = a \vee b \vee c$, and $\alpha = (a \wedge \neg b \wedge c) \vee (a \wedge b \wedge \neg c) \vee (a \wedge \neg b \wedge \neg c)$.
We have:
$\varepsilon^1(\Sigma) = (a \vee b) \wedge (a \vee c) \wedge (b \vee c)$,
$\varepsilon^2(\Sigma) = a \wedge b \wedge c$, and finally
$\varepsilon^3(\Sigma) \vdash \bot$.
Therefore:
\[
\varepsilon^1(\Sigma) \wedge \alpha = (a \wedge \neg b \wedge c) \vee (a \wedge b \wedge \neg c)
\]
and
$\varepsilon^2(\Sigma) \wedge \alpha \vdash  \bot$.
The value of $n$ in Equation
\ref{eq:LastCons} is then equal to 1.

For Definition \ref{def:LastConsistentEro}, $\gamma$ can be
anything in the set
\[
PPE_{\Elc{}{}}(\alpha) = \{ (a \wedge \neg b \wedge c), (a \wedge b
\wedge \neg c), (a \wedge \neg b \wedge c) \vee (a \wedge b \wedge
\neg c) \}.
\]
To compare $\Elc{}{}$ with $\Elne{}{}$, notice  that
$\varepsilon^1(\Sigma\wedge\alpha)=\perp$. Hence
$\Elne\gamma\alpha$ for any $\gamma\vdash_\Sigma\alpha$.  In
particular,  $\Elne{(a\wedge\neg b\wedge \neg c)}\alpha$ which
does not hold for $\Elc{}{}$.

There is an alternative way of looking at $\Elc{}{}$ which will be
particularly useful in the next section. The iteration of the
erosion operator provides a method of linearly pre-ordering the
models of $\Sigma$, according to the morphological ordering
introduced in Section~\ref{sec:MM}
(Definition~\ref{def:fundamentalOrder} and Equation~\ref{order3},
considering here only the sequence of successive erosions). It is
not difficult to verify that the following holds:
\begin{equation}
\label{repre}
\Elc\gamma\alpha \,\,\mbox{\iff} \llbracket \Sigma \wedge \gamma \rrbracket \subseteq \mbox{min}
(\llbracket \Sigma \wedge \alpha \rrbracket ,\preceq_f).
\end{equation}

% $\vdash_\Sigma$ is interesting because $\gamma$ could then have
% models outside $\Sigma$, which may lead to more interesting
% explanations from a syntactic point of view (not that the syntax
% of $\Sigma$ is not taken into account in the proposed approach,
% since all operations are performed on the models, at a semantic
% level). However this may also add noise to the explanations. Two
% possibilities can be suggested to limit this effect: (i) To use
% $\vdash$, at the price of loosing meaningful explanations in some
% cases from a syntactical point of view. (ii) To impose that
% explanations have to be built out of a user defined set of atoms.

One of the original features of the proposed approach is that minimality is obtained directly, by construction. There is no need for a second step aiming at selecting minimal explanations among hypotheses obtained in a first step.

An interpretation can be that the morphological ordering provides a kind of plausibility order among the possible explanations. The preferred explanation is then the most plausible one according to this ordering.

\subsection{Examples}
\label{abdu:examples}

We will explore some ways of defining structuring elements which
are more appropriate for the task of finding explanations. We will
analyze  the following example through different structuring elements. %again Example~\ref{ex:fundamentalExample}.

\begin{example}
\label{ex:fundamentalExample}
%To infer the best explanation of an observation
%is usually known as {\em abduction}. In the logic-based approach
%to abduction, the background theory is given by a consistent set
%of formulas $\Sigma$. This background theory helps to build the best explanations.
%Actually, he notion of a {\em possible explanation}
%is defined by saying that a formula $\gamma$ is an explanation of
%$\alpha$ if $\Sigma\cup\{\gamma\}\vdash \alpha$ (this will be
%written $\gamma\vdash_\Sigma\alpha$). An explanatory relation is a
%binary relation $\expl{}{}$ where the intended meaning of
%$\expl{\gamma}{\alpha}$ is ``$\gamma$ is a {\em preferred
%explanation} of $\alpha$''.
Let us consider the very simple theory $\Sigma_1 = \{ a \rightarrow c, b \rightarrow c\}$ (represented by the same formula $\varphi$ as the one in Figure~\ref{fig:exampleEro}), and suppose that the observation is $c$.
What are the ``good'' explanations of $c$?
We present three  different interpretations where the most natural answers would be different.
%In this paper, we propose tools to deal with such kinds of problems.
%
%\[
%\Sigma_1 = \left \{
%\begin{array}{ccc}
%a  & \rightarrow & c\\
%b & \rightarrow & c
%\end{array} \right .
%\]
%In this example $c$ is  the observation to be explained.
We
usually expect that the causes of $c$ are among $a, b$. Let us consider the following three interpretations, where different explanations may be expected:
\begin{enumerate}
\item
\[
\begin{array}{lcl}
a & = & \mbox{\em rained\_last\_night} \\
b & = & \mbox{\em sprinkle\_was\_on}\\
c & = & \mbox{\em grass\_is\_wet}
\end{array}
\]
The ``common sense cautious explanation'' of $c$ is $a\vee b$.

\item
\[
\begin{array}{lcl}
a & = & \mbox{\em low\_taxes} \\
b & = & \mbox{\em investment\_increases}\\
c & = & \mbox{\em economy\_grows}
\end{array}
\]

An explanation that enhances the chances of achieving the goal of
making the economy to grow is $a\wedge b$.

\item
\[
\begin{array}{lcl}
a & = & \mbox{\em book\_was\_left\_somewhere else} \\
b & = & \mbox{\em somebody\_took\_the book}\\
c & = & \mbox{\em book\_is\_not\_in\_the shelf}
\end{array}
\]
An explanation  based on the principle of the ``Ockham's razor'' will
select either $a$ or $b$ but not both, that is to say,
$(a\wedge\neg b)\vee(\neg a\wedge b)$.
\end{enumerate}
\end{example}

\begin{example}
\label{ejemplo} Let $Ab$ be a set of atoms (sometimes  are called
{\em abducibles}). As before, $B_\omega$ denote the ball of radius $1$
centered at $\omega$ (with respect to the Hamming distance for instance). Let
\[
B^{ab}_\omega=\{\omega'\in B_\omega \mid \omega(x)= \omega'(x)\;\mbox{for all $x\not\in
Ab$}\}.
\]
$B^{ab}_\omega$ contains those valuations in $B_\omega$ which agree with
$\omega$ outside $Ab$. Recall that in Example
\ref{ex:fundamentalExample} we consider the following domain
theory:
\[
\Sigma_1= \left \{
\begin{array}{ccc}
a  & \rightarrow & c\\
b & \rightarrow & c
\end{array} \right .
\]
In this example $c$ is  the observation to be explained. We
usually expect that the causes of $c$ are among $a, b$, so we set
$Ab$ to be $\{a,b\}$. We will work with the notion of explanation
given by $\Elc{}{}$.

\begin{enumerate}
\item If we use the standard structuring element $B_\omega$ we obtain
that  $\varepsilon^1 (\Sigma)= \neg a\wedge\neg b\wedge c$
and $\varepsilon^2(\Sigma)=\bot$. Thus a preferred
explanation of $c$ is
\[
\neg a\wedge\neg b\wedge c.
\]

\item Now we use $B^{ab}_\omega$  as structuring element. Then
$\varepsilon^1 (\Sigma)= \varepsilon^2(\Sigma)=
\Sigma\wedge c$. Thus a preferred explanation of $c$ is $c$.
\end{enumerate}
\end{example}

The preferred explanation given in the first example above seems
to be ``wrong'' because the expected causes of $c$ should be among
$a$ and $b$. And the second example says nothing about an
explanation of $c$. We will make some comments about this after
the next example.

\begin{example}
\label{ejemplo2} Let $\Sigma_1$ and $Ab$ as in Example
\ref{ejemplo}. Let
\[
\Sigma_2=\Sigma_1\cup\{a\vee b\}.
\]
Notice that $\Sigma_2$ is logically equivalent to $\{ (a\wedge c) \vee (b \wedge c) \}$. It models explicitly that $a \vee b$ is part of the theory, and then causes of $c$ can be found among $a$ and $b$.

\begin{enumerate}
\item With the standard ball $B_\omega$ we get $\varepsilon^1
(\Sigma_2)=\bot$. Thus, $\varepsilon_{\ell c} (\Sigma_2,c)=\Sigma_2$. In particular,
\[
\Elc{(a\vee b)}{c}.
\]
\item Now we use $B_\omega^{ab}$. Then $\varepsilon^1 (\Sigma_2)=a\wedge b \wedge c$ and $\varepsilon^2 (\Sigma_2)=\bot$. Thus
\[
\Elc{(a\wedge b)}{c}.
\]
Notice that  $c\not\!\!\rhd^{\ell c}(a\vee b)$.
\item Consider the
following structuring element
\[
B^{ab}_{\omega,2}=\{\omega\}\cup\{\omega'\in \Omega \mid d(\omega,\omega')=2 \mbox{ and }
\omega(x)=\omega'(x)\;\mbox{for all $x\not\in Ab$}\}
\]
where $d$ denotes the Hamming distance. Then $\varepsilon^1(\Sigma_2)=\varepsilon^2(\Sigma_2) = (\neg a\wedge b\wedge c) \vee (a\wedge \neg b\wedge c)$.
Thus,
\[
c\rhd^{\ell c}(a\wedge\neg b)\vee(\neg a\wedge  b).
\]
Notice that  $c\not\!\!\rhd^{\ell c}(a\wedge b)$.

\end{enumerate}
\end{example}

In Example \ref{ejemplo2} we get the ``expected'' solutions, as described in Example~\ref{ex:fundamentalExample}.
One way to understand it is as follows. Given $\Sigma$ and a set
of atoms $Ab$, let $AbForm$ be the set of formulas that use
only atoms from $Ab$. Given an observation formula $\alpha$, the
{\em cautious explanation} of $\alpha$ (with respect to $(\Sigma,
Ab)$) is defined by:
\[
ce(\alpha)=\bigvee \{\gamma\in Abform \mid
\Sigma\not\vdash\neg\gamma \mbox{ and }  \Sigma\cup\{\gamma\}\vdash
\alpha\}.
\]
Since the language is finite, restricting the formulas $\gamma $
appearing in the definition of $ce(\alpha)$ to be a conjunction of
literals from $Ab$, we get that  $ce(\alpha)$ is well defined. For
instance, in Example \ref{ejemplo} we have $ce(c)=a\vee b$. By
adding to $\Sigma$ the cautious explanation of the observation we
are imposing an extra constraint that helps to find some of its
``natural'' explanations. The expanded theory seems to be a useful
tool for the task of finding ``correct'' explanations. All this is
illustrated by Example \ref{ejemplo2}, where the choice of an
appropriate structuring element allows us to find the expected
explanations in the three situations  presented in
Example~\ref{ex:fundamentalExample}.

\begin{figure}[htbp]
\centerline{
\includegraphics[height=3.5cm]{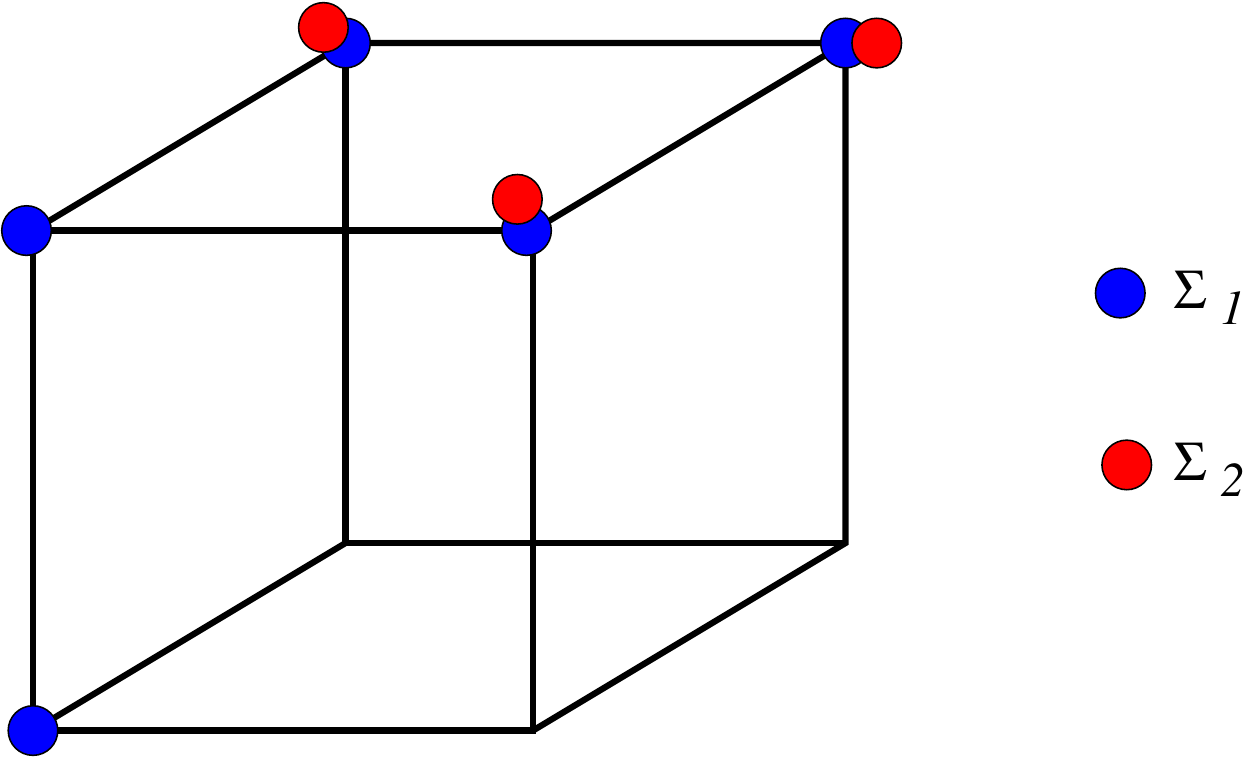} \hspace*{1cm}
\includegraphics[height=3.5cm]{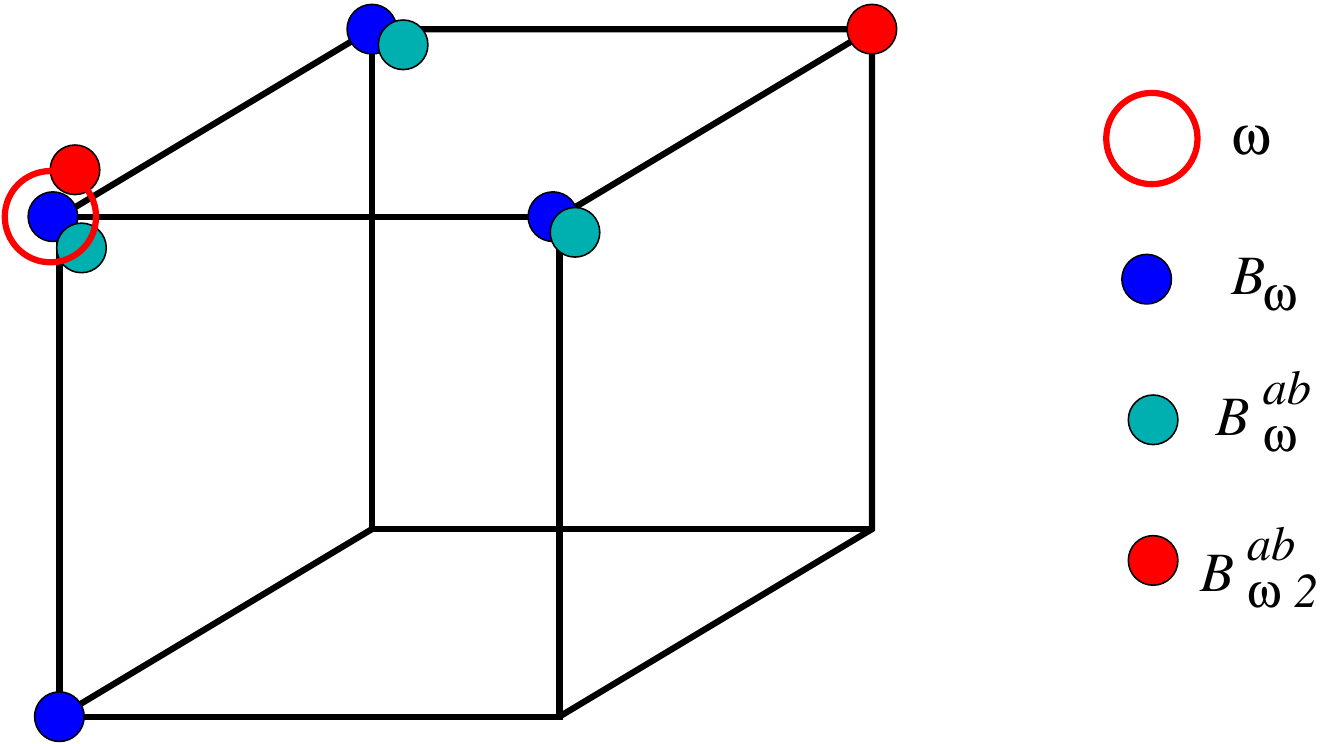}
}
\caption{Illustration of $\Sigma_1$ and $\Sigma_2$ (left) and of three different structuring elements centered at $\omega$ (right).\label{fig:ex2S3SE}}
\end{figure}

Table~\ref{tab:examples} summarizes the results for the last two examples, for $\Sigma_1$ and $\Sigma_2$ and the three considered structuring elements (Figure~\ref{fig:ex2S3SE}).

\begin{table}[htbp]
\begin{center}
\begin{tabular}{|c|c|c|}\hline
           & $\Sigma_1$ & $\Sigma_2$ \\ \hline \hline
$B_\omega$ &     $\neg a\wedge \neg b \wedge c$ & $a \vee b$\\ \hline
$B_\omega^{ab}$ &     $c$     &    $a\wedge b$        \\ \hline
$B_{\omega,2}^{ab}$ &   $c$  &  $(a\wedge \neg b)\vee(\neg a\wedge b)$   \\ \hline
\end{tabular}
\end{center}
\caption{Explanations of observation $c$ for two backgroung theories and three different structuring elements.}
\label{tab:examples}
\end{table}

These examples illustrate how different explanations can be obtained using appropriate structuring elements. Roughly speaking, if $a$ and $b$ are incompatible, then the exclusive disjunction is appropriate, and it is obtained using $B_{\omega,2}^{ab}$. If they are compatible, a parcimonious explanation is the disjunction (as required for instance in model-based diagnosis), obtained for $B_\omega$, while a more sure or constrained explanation is the conjunction, obtained for $B_\omega^{ab}$.

\subsection{Rationality postulates}
\label{sec:properties}

In this section we study the properties of the two proposed
explanatory relations according to the postulates introduced in
\cite{PINO-99}. The basic rationality postulates for explanatory
relations are the following:\\[2mm]
% (we use the notation
% $\alpha\vdash_\Sigma\beta$ instead of $\Sigma\cup\{\alpha\}$):

\begin{tabular}{lcl}
\LLEs: & & If $\nms \alpha\leftrightarrow \alpha'$ and $\expl{\gamma}{\alpha}$ then $\expl{\gamma}{\alpha'}$.\\

\RLEs: & & If $\nms \gamma \leftrightarrow \gamma'$ and $\expl{\gamma}{\alpha}$ then $\expl{\gamma' }{\alpha}$.\\

\ECM: & & If $\expl{\gamma}{\alpha} $ and $\gamma\nms\beta $ then $ \expl{\gamma}{(\alpha\wedge\beta)}$.\\

\ECC:   & & If $ \expl{\gamma}{(\alpha\wedge \beta) }$ and $\forall \delta \;[\expl{\delta}{\alpha} \;\Rightarrow \;
\delta\nms \beta\;] $ then $\expl{\gamma}{\alpha} $.\\

\RA:    & & If $ \expl{\gamma}{\alpha}$, $\gamma'\nms\gamma$ and $\gamma'\not\nms\bot $ then $ \expl{\gamma'}{\alpha}$.\\

\ERW:      && If $ \expl{\gamma}{\alpha}$ and $\expl{\delta}{\alpha} $ then $\expl{(\gamma\vee\delta)}{\alpha} $.\\

\LOR:     && If $\expl{\gamma}{\alpha} $ and $\expl{\gamma}{\beta} $ then $\expl{\gamma}{(\alpha\vee\beta)} $.\\

\EDR:     && If $ \expl{\gamma}{\alpha}$ and $\expl{\delta}{\beta} $ then $\expl{\gamma}{(\alpha\vee\beta)} $ or $\expl{\delta}{(\alpha\vee\beta)}$.\\

\ERC:   && If $\expl{\gamma}{(\alpha\wedge\beta)}$ and $\exists\delta\; [\expl{\delta}{\alpha}\;\&\;\delta\nms\beta]$ then $\expl{\gamma}{\alpha}$.\\

\ERef:       && If $\expl{\gamma}{\alpha} $  then $\expl{\gamma}{\gamma} $.\\

\Econ:      &&  $ \not\nms \neg \alpha$ iff there is $\gamma$ such that $\expl{\gamma}{\alpha}$.
\end{tabular}\\

The intended meaning and motivation for these postulates can be found in \cite{PINO-99}.

%\fbox{TODO: link with Chellas' axioms? I have to check what I meant by that...}

It is immediate from the definition of $\Elc{}{}$ and $\Elne{}{}$
that \LLEs, \RLEs, \RA, \ERW, and \Econ\ are satisfied. Moreover,
from the representation of $\Elc{}{}$ given by Equation
\ref{repre} and some general results of \cite{PINO-99} we get the
following proposition.

\begin{proposition}
$\Elc{}{}$ is a causal E-rational explanatory relation. In
particular, it satisfies \LLEs, \RLEs, \RA, \ERW, \Econ, \ECM\ and
\ERC.
\end{proposition}

From the results in \cite{PINO-99} we also know that by being
E-rational, $\Elc{}{}$ also satisfies \ECC, \ERef, \EDR\ and \LOR.
However, the situation for $\Elneu{}{}$ and $\Elned{}{}$ is quite different since
% ,
% as we will see below,
the basic postulates \ECM\ and \ECC\ do
not hold (for a proof of this claim see Appendix~\ref{app:proofs}).

\medskip

We introduce a weaker form of these postulates:\\

\begin{tabular}{lcl}
\EWCM: & & If $\expl{\gamma}{\alpha}$ and $\expl\gamma\beta$ then $\expl{\gamma}{(\alpha\wedge\beta)}$.\\

\EWCC: && If $\expl{\gamma}{(\alpha\wedge \beta) }$ and $\forall \delta \;[\expl{\delta}{\alpha} \;\Rightarrow \; \expl\delta\beta\;]$ then $\expl{\gamma}{\alpha}$.

\end{tabular}\\

% \begin{description}
% \item[\EWCM:]\hspace{4cm}
% \( \displaystyle \frac{\expl{\gamma}{\alpha} \;\; ;
% \;\expl\gamma\beta}
%  {\expl{\gamma}{(\alpha\wedge\beta)}}
% \)
% \end{description}
% \begin{description}
% \item[\EWCC:]\hspace{3cm}
% \( \displaystyle \frac{\expl{\gamma}{(\alpha\wedge \beta) }\;\;
% ,\;\; \forall \delta \;[\expl{\delta}{\alpha} \;\Rightarrow \;
% \expl\delta\beta\;]}
% {\expl{\gamma}{\alpha}} \)
% \end{description}

These new postulates might also look natural.
% than the original version \ECM\ and \ECC.
However, $\Elneu{}{}$ and $\Elned{}{}$ are the first
natural non trivial examples known in the literature that satisfy \EWCM\ and
\EWCC\ but neither \ECM\ nor \ECC \footnote{$\EWCM$ in fact was already considered by Flach \cite{Flach96} but he did not provide any example for it not satisfying already the stronger version \ECM.}.

The next proposition collects all the facts we know about
$\Elneu{}{}$ and $\Elned{}{}$.

\begin{proposition}
\label{prop:PropLNE}
The explanatory relations $\Elneu{}{}$ and $\Elned{}{}$ satisfy
\LLEs, \RLEs,
\ERW, \EWCM, and \Econ. Moreover $\Elneu{}{}$ satisfies \ERef\ and \EWCC\ but
$\Elned{}{}$ does not, and $\Elned{}{}$ satisfies \RA\ but $\Elneu{}{}$ does not.

\end{proposition}

The proof of this result can be found in Appendix~\ref{app:proofs}.

For some properties, they may be required or not, depending on the application. For instance the fact that $\Elned{}{}$ does not satisfy \ERef\ is a good point if one wants to avoid ``self-explanations'', i.e. $\gamma \rhd \gamma$.

\bigskip

We end this section by considering the postulate \LOR.
Actually, the relations  $\Elneu{}{}$ and $\Elned{}{}$ do not satisfy the postulate \LOR\
(for a counter-example see Appendix~\ref{app:proofs}).
Since \EDR\ implies \LOR\ \cite{PINO-99}, then we already know that \EDR\ fails  for $\Elneu{}{}$ and $\Elned{}{}$.

% We will
% give a counter-example of it for $\Elne{}{}$.  Again $\Sigma$ will
% be $\{\top\}$. Consider
% \[
% \alpha = (a \vee b \vee c) \wedge (a \vee \neg b \vee \neg c)
% \]
% and
% \[
% \beta = (\neg a \vee \neg b \vee c) \wedge (a \vee \neg b \vee c)
% \wedge (a \vee b \vee c).
% \]
% We have:
% \[
% \varepsilon^1(\alpha) = (a \wedge b \wedge \neg c) \vee (a \wedge
% \neg b \wedge c) = \varepsilon_\ell(\alpha),
% \]
% \[
% \varepsilon^1(\beta) = a \wedge \neg b \wedge c =
% \varepsilon_\ell(\alpha),
% \]
% \[
% \alpha \vee \beta = a \vee b \vee c,
% \]
% \[
% \varepsilon^1((\alpha \vee \beta)) = (a \vee b) \wedge (a \vee c)
% \wedge (b \vee c),
% \]
% \[
% \varepsilon^2(\alpha \vee \beta) = a \wedge b \wedge c =
% \varepsilon_\ell(\alpha \vee \beta).
% \]
% Let $\gamma = a \wedge \neg b \wedge c$. Then $\Elne\gamma\alpha$
% and $\Elne\gamma\beta$, but $\nElne{\gamma}{(\alpha \vee \beta)}$.

% Since \EDR\ implies \LOR\ \cite{PINO-99}, then we already know that \EDR\ fails  for $\Elne{}{}$.

%\subsection{Summary table}

\bigskip

Table \ref{tab:summary} summarizes the results we obtained so far.

\begin{table}[htbp]
\begin{center}
\begin{tabular}{| l || c | c |c |} \hline
Property & $\Elneu{}{}$ & $\Elned{}{}$ & $\Elc{}{}$ \\
 & (Equation \ref{eq:lasterosion1-unique}) & (Equation \ref{eq:lasterosion1}) &  (Equation \ref{eq:consistenterosion1}) \\ \hline \hline
{\LLE} & $\surd$ & $\surd$ & $\surd$  \\
{\RLE} & $\surd$ & $\surd$ & $\surd$ \\ \hline
{\ECM} & $\times$ & $\times$ &  $\surd$ \\
{\EWCM} & $\surd$ & $\surd$ & $\surd$ \\ \hline
{\ECC} & $\times$ & $\times$ & $\surd$ \\
{\ERC} & $\times$ & $\times$ & $\surd$ \\
{\EWCC} & $\surd$ & $\times$ & $\surd$ \\
{\ERef} & $\surd$ & $\times$ & $\surd$ \\ \hline
{\ERW} & $\surd$ & $\surd$ & $\surd$ \\
{\RA}  & $\times$ & $\surd$ & $\surd$ \\ \hline
{\LOR} & $\times$ & $\times$ & $\surd$ \\
{\EDR} & $\times$ & $\times$ & $\surd$ \\ \hline
{\Econ} & $\surd$ &  $\surd$ & $\surd$ \\ \hline
\end{tabular}\\

\caption{Properties of the proposed relations.}
\label{tab:summary}
\end{center}
\end{table}

%\subsection{Examples}
%
%\fbox{To do: at least graphical examples illustrating all what precedes.}
%
%\fbox{Find good examples for both versions?}

%%%%%%%%%%%%%%%%%%%%%%%%%%%%%%%%%%%%%%%%%%%%%%%%%%%%%%%%%%%%%****

\subsection{Unified view using the fundamental pre-order $\preceq_f$}
\label{sec:unify}

%\fbox{TO DO:}
%\begin{itemize}
%\item revise completely... In particular, modify definitions to include the possibility of fixed point.
%\item comment on the meaning of sigma (a fixed theory for abduction, a modifiable background knowledge for revision...)
%\item illustration with different structuring elements, and our basic example with $\Sigma_1$.
%\end{itemize}

We present in this section a unified treatment of abduction and
revision. In particular, we propose to put in the same framework some of
the results of Sections~\ref{sec:Revision} and ~\ref{sec:abduction} (and~\cite{IPMU-00b,ECSQARU-01}), using the fundamental morphological pre-order relation $\preceq_f$.

%
%\fbox{$\vdash$ should be $\vdash_\Sigma$?}

In the following we still assume anti-extensive erosions and extensive dilations.

There is an alternative way of looking at $\Elc{}{}$ which will be
particularly useful in what follows. The iteration of the
erosion operator provides a method of linearly pre-ordering the
models of $\Sigma$. %Consider the following relation among models:
We have already noted that, when $\alpha$ is consistent with $\Sigma$, we have a representation of the relation
$\Elc{}{}$ in terms of the morphological order given by the equivalence \myref{repre}.

Actually, if we take the following pre-order  over the models of $\Sigma$:
\begin{equation}\label{order1}
\omega \leq_E \omega' \iffdef \forall k \;\; (\omega'\in \varepsilon^k (\Sigma)\rightarrow \omega\in \varepsilon^k(\Sigma)),
\end{equation}
it is clear that $\leq_E$ and $\preceq_f$ coincide over $\mymod\Sigma$. Thus equivalence \myref{repre} can be rewritten as:
% is a total pre-order and it is not
% difficult to verify that the following holds:
\begin{equation}
%\label{repre}
\Elc\gamma\alpha \,\,\mbox{\iff}
\llbracket \gamma \wedge   \Sigma \rrbracket \subseteq \mbox{min}
(\llbracket \Sigma\wedge\alpha \rrbracket ,\leq_E).
\end{equation}
% Notice that from the equivalence \ref{repre} and by the results in
% \cite{PINO-99}, it is clear that $\Elc{}{}$ is, as we already
% said, a rational explanatory relation.

Let us now come back to the revision
 based on dilation. As described in Section \ref{sec:Revision} (see also \cite{IPMU-00b}), the idea is to
dilate $\Sigma$ (which is not necessarily consistent with
$\alpha$) until it becomes consistent with $\alpha$. Note that $\Sigma$ is then no more considered as a fixed theory but rather as a background knowledge, which can evolve. More
precisely, we define $\circ$ as:
\begin{equation}
\Sigma\circ\alpha =
\left\{
\begin{array}{ccl}
  \delta^n( \Sigma) \wedge \alpha & & \mbox{where }
n = \min \{ k \in \mathbb{N} \mid \delta^k(\Sigma) \wedge \alpha
\mbox{ is consistent\}}\\
\Sigma & & \mbox{if there is no $k$ such that }  \delta^k(\varphi) \wedge \psi \not\vdash \bot
\end{array}
\right.
 \label{eq:revision1}
\end{equation}

The iteration of the dilation operator provides a
method of linearly pre-ordering the models of $\mymod{\delta_\ell(\Sigma)}$. Consider the following
relation among models:
\begin{equation}\label{order2}
\omega \leq_D \omega' \iffdef \forall k \;\; (\omega'\in \delta^k (\Sigma)\rightarrow
\omega\in \delta^k(\Sigma)).
\end{equation}

Indeed, it is clear that $\leq_D$ is a total pre-order over $\mymod{\delta_\ell(\Sigma)}$; we will call
it the {\em total preorder associated with $\Sigma$ using successive
dilations}. It is not difficult to verify that the following
holds:
\begin{equation}
\label{reprerev}
\llbracket \Sigma\circ\alpha \rrbracket =
\left\{
\begin{array}{ccc}
  \min (\llbracket \alpha \rrbracket ,\leq_D). & & \mbox{if }
\alpha\wedge\delta_\ell(\Sigma)\not\vdash\bot\\
\llbracket \Sigma \rrbracket & & \mbox{if  }  \alpha\wedge\delta_\ell(\Sigma)\vdash\bot
\end{array}
\right.
\end{equation}

Indeed, it is easy to check that over the set $\mymod{\delta_\ell(\Sigma)}\setminus\mymod\Sigma$ the relations $\leq_D$ and $\preceq_f$ coincide.

By the representation theorem for credibility-limited revision operators
(see~\cite{BFKP12}), it follows from Equation~\ref{reprerev} that $\circ$ is credibility-limited revision operator~\cite{BFKP12,HFCF01},
operators that generalize the classical AGM-revision operators~\cite{AGM-85,KatsunoMendelzon91}.

The pre-order defined by Equations~\ref{order1} and~\ref{order2} can be merged in the morphological ordering $\preceq_f$ introduced in Section~\ref{sec:MM}. By the previous observations, the morphological order $\preceq_f$ is $\leq_E$ over $\mymod\Sigma$ and $\leq_D$ over the set
$\mymod{\delta_\ell(\Sigma)}\setminus\mymod\Sigma$.
% This
% preorder is the main tool for what follows.

Based on the morphological ordering, we can
associate with each observation $\alpha$ the following set of valuations:
\[
M(\alpha)=
\left\{
\begin{array}{ccl}
  \min ( \llbracket \alpha \rrbracket ,\preceq_f) & & \mbox{if }
\alpha\wedge\delta_\ell(\Sigma)\not\vdash\bot\\
\llbracket \Sigma \rrbracket & & \mbox{if  }  \alpha\wedge\delta_\ell(\Sigma)\vdash\bot
\end{array}
\right.
\]
Note that the criterion used to define $M(\alpha)$ is based on the
morphology operators $\delta$ and $\varepsilon$. The interpretation we give to
$M(\alpha)$ is that it contains those worlds that are
(morphologically) more relevant given the observation $\alpha$.
Therefore for the task of revising $\Sigma$ or explaining $\alpha$
we only look at $M(\alpha)$. This will be made precise in the
result that follows. We will denote by $C(\alpha)$ the formula
whose models are  exactly $M(\alpha)$.

\begin{theo}\label{seqfunda}
Let $\Sigma$, $\alpha$ and $\gamma$ consistent formulas.
\begin{enumerate}
\item If $\alpha$ is consistent with $\Sigma$, then
$\Elc\gamma\alpha$ iff $\gamma\vdash C(\alpha)$.

\item  If $\alpha$ is inconsistent with $\Sigma$, then
$\Sigma\circ\alpha=C(\alpha)$.
\end{enumerate}
\end{theo}

The previous result suggests the following definitions
\begin{equation}\label{seq_fundamental1}
\alpha\expf \gamma\iffdef \gamma\vdash C(\alpha)
\end{equation}
and
\begin{equation}\label{seq_fundamental2}
\Sigma\circ_f\alpha=C(\alpha)
\end{equation}
where $\alpha$ and $\gamma$ are consistent formulas.

As an example, let us consider the example in Figure~\ref{fig:fundamentalOrder} for $\Sigma = \Sigma_1$. For $\alpha = (\neg a \wedge \neg b \wedge \neg c) \vee (\neg a \wedge \neg b \wedge c) \vee (\neg a \wedge b \wedge \neg c)$, $\alpha$ is consistent with $\Sigma$ and its explanation is $\gamma \equiv_\Sigma \neg a \wedge \neg b \wedge c$, which corresponds to the rank 0 in Table~\ref{tab:stratification}. Now if $\alpha$ is reduced to $\alpha = \neg a \wedge b \wedge \neg c$, then it is no more consistent with $\Sigma$ and the revision applies.

Some comments about these definitions should be made. First of all,
even when an observation is inconsistent with the background
theory $\Sigma$ there is  a formula $\gamma$ such that
${\alpha}\expf{\gamma}$. That is to say, we can ``explain'' more
observations with $\expf$ than with $\Elc{}{}$. The interpretation
we give to this fact is that for explaining an observation it is
allowed (if necessary) to ``change'' the background theory.
% (this will be made more  precise in the section about dynamics).
Thus in the explanatory process described by $\expf$ the observation is
absolutely reliable. Notice also that $\expf{}{}$ makes it explicit that
some explanations might not be consistent with $\Sigma$.

The operator $\circ_f$ is not an  AGM revision operator for $\Sigma$ (even not a credibility-limited revision operator), since
when the observation $\alpha$ to be incorporated is consistent
with $\Sigma$ we have only $\Sigma\circ_f \alpha\vdash
\Sigma\wedge\alpha$, not the equivalence (the equivalence in the case where $\alpha$ and $\Sigma$ are consistent is just the vacuity postulate, usually denoted by K*4, which is related to the minimality R2). The reason for this is that $\circ_f$ is based
on preferences on models of $\Sigma$ , so even when $\Sigma\wedge \alpha$ is consistent,
some sort of central reason for accepting $\alpha$ has to be found. Note that the previous remark
says that $\circ_f$ does not satisfy the postulate K*4, which has been
criticized by some authors in particular  in \cite{RYAN-91}. Unlike
 Ryan's operators, which are based on ordered theory presentations, K*4 and success are  the
only   postulates which are not satisfied by $\circ_f$. However, note that $\circ_f$ satisfies the modified
version of success of credibility-limited revision operators, that is: $\Sigma\circ\alpha\vdash\alpha$ or
$\Sigma\circ\alpha\equiv\Sigma$.
% \fbox{explain this and K*4?}

%%%%%%%%%%%%%%%%%%%%%%%%%%%%%%%%%%%%%%%%%%%%%%%%%%%%%%%%
%%%%%%%%%%%%%%%%%%%%%%%%%%%%%%%%%%%%%%%%%%%%%%%%%%%%%%%%%%

%\input{conclusion-essai}

\section{Final remarks and perspectives}
\label{sec:conclu}

We have given the fundamental concepts and techniques in mathematical morphology, and have shown how to interpret these techniques
in terms of mathematical logic, namely in propositional logic. This connection has originated a new domain called morphologic. We have
used dilation operators in order to define belief revision operators and belief merging operators.

We have shown that we
can find some operators defined in the literature when the dilation operators come from a distance. Moreover we have extended
the class of belief revision operators and the class of belief merging operators by using a larger class of operators, in particular having the extensivity and exhaustivity properties.

A similar work has been done using contraction operators. These operators are used in two ways in order to define explanatory relations.
It is interesting to note that the use of different structuring elements is determinant in the way the information is structured.
The examples in Section~\ref{abdu:examples} point out in a clear way this phenomenon.

Under the assumption that the geometry comes from the Hamming distance between interpretations,
we have shown how to compute dilation, erosion, last erosion, ultimate erosion, opening and
skeleton operators over formulas. These calculations constitute the basis of our applications to
different tasks in knowledge representation.

We have proven that our general operators of revision and fusion are well behaved, in particular they satisfy the AGM postulates and the postulates
of integrity constraints belief merging. We have also proven that the explanatory relations defined using morphologic satisfied suitable structural properties.

Potential extensions would be to analyze how minimality criteria for could be expressed in the proposed framework, as the ones proposed for abduction~\cite{bienvenu2008,eiter1995,halland2012}, revision for Horn clauses~\cite{DP11,DP15,ZPZ13} or for description logics~\cite{ARXIV,QLB06,QY08,RW09,RW10,WWT10}, or more generally for institutions~\cite{AABH15} and satisfaction systems~\cite{MA:AIJ-17}.

One interesting feature that is worth to remark is the fact that morphologic allows us to give an ordered structure to the pieces of information.
That is, it allows having preferences over the formulas. It is exploited by the morphological total pre-order defined by Equation~\ref{order3}. Note that these preferences
depend on the structuring element used for defining dilations and erosions.

Finally, our approach provides a reusable framework for performing numerous operations on formulas, and includes computational and axiomatic building blocks, to be applied in different reasoning problems.

Future work will aim to apply the tools of morphologic in order to explain multiple observations and for putting dynamics in the explanatory process. We also expect to treat mediation process using the tools developed in this work.

\appendix

\section{Proofs}
\label{app:proofs}

In this appendix, we provide  proofs of certain technical claims.\\

\noindent{\bf A counter-example of \ECM\  for $\Elneu{}{}$ and $\Elned{}{}$.}\\

Note that a counter-example of \ECM\  for $\Elneu{}{}$ is also a counter-example of \ECM\  for $\Elned{}{}$.

\noindent In this example $\Sigma$ will be $\{\top\}$, so we will remove it
altogether.  Let us consider the following formulas:
% (see Figure \ref{fig:ECM}):
\[
\alpha = \neg a \vee b \vee c,\quad \beta= (\neg a \vee \neg b \vee \neg c) \wedge (\neg a \vee b \vee \neg c)
\]
\[
\alpha \wedge \beta = (\neg a \vee \neg b \vee \neg c) \wedge
(\neg a \vee b \vee \neg c) \wedge (\neg a \vee b \vee c).
\]

 \Omit{
\begin{figure}[htbp]
\centerline{\hbox{ \input{ECM1.pstex} }}
\centerline{\hbox{ \input{ECM2.pstex} }}
\caption{A counter-example for {\bf E-CM}.} \label{fig:ECM}
\end{figure}
}

Using the computation formulas for erosion of a formula under CNF
(Proposition~\ref{prop:dilConjLit}), we get:
\[
\varepsilon^1(\alpha) = (\neg a \vee b) \wedge (\neg a \vee c)
\wedge (b \vee c),
\]
\[
\varepsilon^2(\alpha) = \neg a \wedge b \wedge c =
\varepsilon_\ell(\alpha).
\]
A unique world satisfies this formula, and therefore no further
erosion can be performed ($\varepsilon^3(\alpha) \vdash \bot$).
Similarly, we have:
\[
\varepsilon^1(\alpha \wedge \beta) = \neg a \wedge b \wedge \neg c
= \varepsilon_\ell(\alpha \wedge \beta)
\]
which is the last non-empty erosion. It follows that $\Elneu{(\neg
a \wedge b \wedge c)}{\alpha}$; moreover $(\neg a \wedge b \wedge c)\vdash \beta$,
but clearly the formula  $(\neg a \wedge b \wedge c)$  is not a preferred explanation of $\alpha \wedge
\beta$.\\

%%%%%%%%%%%%%%%%%%%%%%%%%%%%%%%%%%%
%%%%%%%%%%%%%%%%%%%%%%%%%%%%%%%%%%%

\noindent{\bf A counter-example of $\ECC$ for $\Elneu{}{}$ and $\Elned{}{}$.}\\

\noindent As the same counter-example works for $\Elneu{}{}$ and $\Elned{}{}$, we omit the subscript in the notation of the relation. Again $\Sigma$ will be $\{\top\}$. Consider
\[
\alpha = a \vee b \vee c \mbox{ and }
\beta = a \vee \neg b \vee \neg c.
\]
We have then:
\[
\varepsilon^1(\alpha) = (a \vee b) \wedge (a \vee c) \wedge (b
\vee c),
\]
\[
\varepsilon^2(\alpha) = a \wedge b \wedge c =
\varepsilon_\ell(\alpha),
\]
\[
\varepsilon^1(\beta) = (a \vee \neg b) \wedge (a \vee \neg c)
\wedge (\neg b \vee \neg c),
\]
\[
\varepsilon^2(\beta) = a \wedge \neg b \wedge \neg c =
\varepsilon_\ell(\beta),
\]
\[
\alpha \wedge \beta = (a \vee b \vee c) \wedge (a \vee \neg b \vee
\neg c),
\]
\[
\varepsilon(\alpha \wedge \beta) = (a \wedge b \wedge \neg c) \vee
(a \wedge \neg b \wedge c) = \varepsilon_\ell(\alpha \wedge
\beta).
\]

Let us now set $\gamma = (a \wedge b \wedge \neg c) \vee (a \wedge
\neg b \wedge c)$, then $\Elne{\gamma}{(\alpha\wedge\beta)}$.  On
the other hand, we have that $\Elne\delta\alpha$ iff $\delta
\equiv a \wedge b \wedge c$ (in this case there is no noise because $\Sigma=\top$). Thus if $\Elne\delta\alpha$, then
$\delta \vdash_\Sigma \beta$. But it is clear that
$\nElne\gamma\alpha$.

\medskip

%%%%%%%%%%%%%%%%%%%%%%%%%%%%%%

\noindent{\bf A counter-example of $\LOR$ for $\Elneu{}{}$ and $\Elned{}{}$.}\\

\noindent  Again in this counter-example  $\Sigma$ will
be $\{\top\}$. Consider, for $\Elned{}{}$:
\[
\alpha = (a \vee b \vee c) \wedge (a \vee \neg b \vee \neg c)
\]
and
\[
\beta = (\neg a \vee \neg b \vee c) \wedge (a \vee \neg b \vee c)
\wedge (a \vee b \vee c).
\]
We have:
\[
\varepsilon^1(\alpha) = (a \wedge b \wedge \neg c) \vee (a \wedge
\neg b \wedge c) = \varepsilon_\ell(\alpha),
\]
\[
\varepsilon^1(\beta) = a \wedge \neg b \wedge c =
\varepsilon_\ell(\alpha),
\]
\[
\alpha \vee \beta = a \vee b \vee c,
\]
\[
\varepsilon^1((\alpha \vee \beta)) = (a \vee b) \wedge (a \vee c)
\wedge (b \vee c),
\]
\[
\varepsilon^2(\alpha \vee \beta) = a \wedge b \wedge c =
\varepsilon_\ell(\alpha \vee \beta).
\]
Let $\gamma = a \wedge \neg b \wedge c$. Then $\Elned\gamma\alpha$
and $\Elned\gamma\beta$, but $\nElned{\gamma}{(\alpha \vee \beta)}$.\\

%Since \EDR\ implies \LOR\ \cite{PINO-99}, then we already know that \EDR\ fails  for $\Elne{}{}$.

\begin{figure}[htbp]
\centerline{\hbox{
\includegraphics[height=3.5cm]{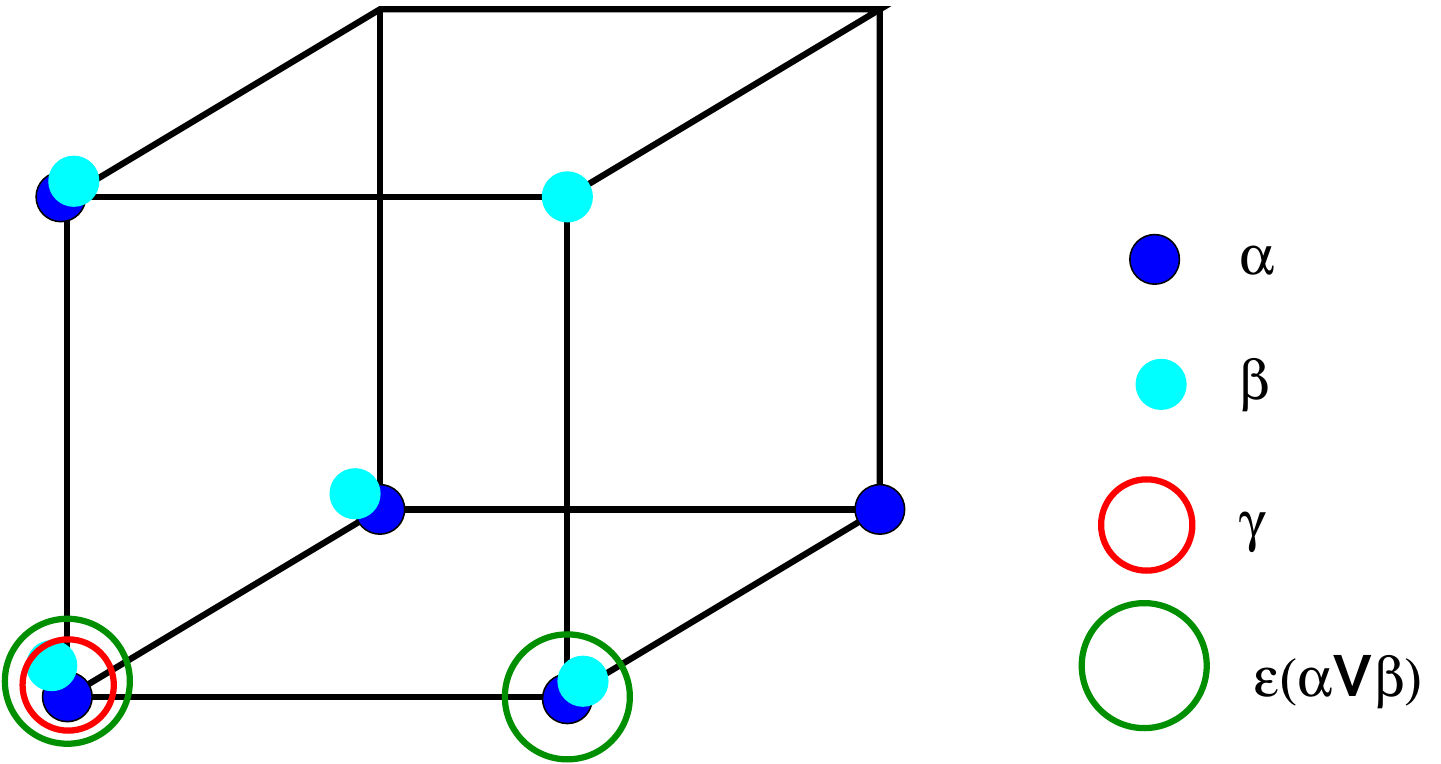}
}}
\caption{Counter-example for LOR for $\Elneu{}{}$.}
\label{fig:CEx-LOR}
\end{figure}

Now for $\Elneu{}{}$, let us consider the example in Figure~\ref{fig:CEx-LOR}. We have $\Elneu\gamma\alpha$ and $\Elneu\gamma\beta$ for $\gamma = \neg a \neg b \neg c$. But the explanations of $\alpha \vee \beta$ are $(\neg a \neg b \neg c) \vee (a \neg b \neg c)$.

%%%%%%%%%%%%%%%%%%%%%%%%%%%

\noindent {\bf Proof of Proposition~\ref{prop:PropLNE}.}\\

In what follows, we detail \EWCM, \EWCC, and \ERef\ for $\Elneu{}{}$ and $\Elned{}{}$. The other properties are straightforward. In particular it is clear that  $\Elneu{}{}$ does not satisfy \RA\ but  $\Elned{}{}$ does satisfy \RA.\\

\noindent (i) \EWCM. First we prove this property for  $\Elneu{}{}$. Let us assume that $\gamma
\equiv_\Sigma \varepsilon_\ell(\Sigma \wedge \alpha)$ with $\varepsilon_\ell(\Sigma \wedge \alpha) = \varepsilon^n(\Sigma \wedge \alpha)$, and $\gamma \equiv_\Sigma \varepsilon_\ell(\Sigma \wedge \beta)$ with $\varepsilon_\ell(\Sigma
\wedge \beta) = \varepsilon^m(\Sigma \wedge \beta)$.
\begin{enumerate}
\item Let us first consider the case where $\varepsilon^{n+1}(\Sigma \wedge \alpha) = \bot$ and $\varepsilon^{m+1}(\Sigma \wedge \beta) = \bot$. Let us assume that $\varepsilon_\ell(\Sigma \wedge \alpha
\wedge \beta) = \varepsilon^k(\Sigma \wedge \alpha \wedge \beta)$. Since erosion commutes with infimum, we have $\varepsilon^k(\Sigma \wedge \alpha \wedge \beta) = \varepsilon^k(\Sigma \wedge \alpha) \wedge \varepsilon^k(\Sigma \wedge \beta)$. If $k >n$ or $k>m$ this conjunction would be inconsistent. Therefore we necessarily have $k\leq n$ and $k \leq m$. Without loss of generality, we take $n \leq m$. Then $\varepsilon^n(\Sigma \wedge \alpha \wedge \beta) = \varepsilon^n(\Sigma \wedge \alpha) \wedge \varepsilon^n(\Sigma \wedge \beta)$. We have $\varepsilon^n(\Sigma \wedge \alpha) \equiv_\Sigma \gamma$ and $\gamma \vdash_\Sigma \varepsilon^n(\Sigma \wedge \beta)$ since $n\leq m$. Hence $\varepsilon^n(\Sigma \wedge \alpha \wedge \beta) \equiv_\Sigma \gamma$. Moreover $\varepsilon^{n+1}(\Sigma \wedge \alpha \wedge \beta) = \bot$. Finally $\Elneu{\gamma}{(\alpha \wedge \beta)}$.

\item Let us now consider the case where $\varepsilon^n(\Sigma \wedge \alpha)$ and $\varepsilon^m(\Sigma \wedge \beta)$ are fixed points, and assume $n \leq m$. For $k=n$, we have $\varepsilon^k(\Sigma \wedge \alpha \wedge \beta) = \varepsilon^n(\Sigma\wedge \alpha) \wedge \varepsilon^n(\Sigma \wedge \beta) \equiv_\Sigma \gamma$, for the same reasons as in the first case. Similarly, $\varepsilon^{n+1}(\Sigma \wedge \alpha \wedge \beta) = \varepsilon^{n+1}(\Sigma\wedge \alpha) \wedge \varepsilon^{n+1}(\Sigma \wedge \beta) = \varepsilon^n(\Sigma\wedge \alpha) \wedge \varepsilon^{n+1}(\Sigma \wedge \beta) \equiv_\Sigma \gamma$ (since $\gamma \vdash_\Sigma \varepsilon^{n+1}(\Sigma \wedge \beta)$, or $\gamma \equiv_\Sigma \varepsilon^{n+1}(\Sigma \wedge \beta)$ if $n=m$). This means that a fixed point has been reached (for $n$ erosions or earlier), and $\Elneu{\gamma}{(\alpha \wedge \beta)}$.

\item If $\varepsilon^{n+1}(\Sigma \wedge \alpha) = \bot$ and $\varepsilon^{m+1}(\Sigma \wedge \beta) = \varepsilon^m(\Sigma \wedge \beta)$ (fixed point), then the first relation would imply $\varepsilon(\Sigma \wedge \gamma) = \bot$ and the second one $\varepsilon(\Sigma \wedge \gamma) = \varepsilon^{m+1}(\Sigma \wedge \beta)$ which is consistent. This leads to a contradiction and this case is not possible. The same reasoning applies if $\varepsilon^{n+1}(\Sigma \wedge \alpha) = \varepsilon^{n}(\Sigma \wedge \alpha)$ and $\varepsilon^{m+1}(\Sigma \wedge \beta) = \bot$.
\end{enumerate}

Now we prove the property for  $\Elned{}{}$. Thus, let us  assume that $\gamma
\vdash_\Sigma \varepsilon_\ell(\Sigma \wedge \alpha)$ with $\varepsilon_\ell(\Sigma
\wedge \alpha) = \varepsilon^n(\Sigma \wedge \alpha)$, $\gamma
\vdash_\Sigma \varepsilon_\ell(\Sigma \wedge \beta)$ with $\varepsilon_\ell(\Sigma
\wedge \beta) = \varepsilon^m(\Sigma \wedge \beta)$, and that the next erosions are empty. Let us assume that the
last non-empty erosion of $\Sigma \wedge \alpha \wedge \beta$ is
obtained for $k$. Since the erosion commutes with the conjunction, we have:
$\varepsilon_\ell(\Sigma \wedge \alpha \wedge \beta) = \varepsilon^k(\Sigma \wedge
\alpha \wedge \beta) = \varepsilon^k(\Sigma \wedge \alpha ) \wedge
\varepsilon^k(\Sigma \wedge \beta)$.

We necessarily have $k \leq n$ and $k \leq m$ since otherwise either $\varepsilon^k(\Sigma \wedge \alpha )$ or $\varepsilon^k(\Sigma \wedge \beta)$ would be inconsistent. This implies, due to the monotonicity property of erosion that:
$\vdash_\Sigma \varepsilon^n(\Sigma \wedge \alpha) \rightarrow \varepsilon^k(\Sigma \wedge \alpha)$
and
$\vdash_\Sigma \varepsilon^m(\Sigma \wedge \beta) \rightarrow \varepsilon^k(\Sigma \wedge \beta)$
from which we derive:
\[
\vdash_\Sigma \varepsilon_\ell(\Sigma \wedge \alpha) \wedge \varepsilon_\ell(\Sigma
\wedge \beta) \rightarrow \varepsilon_\ell(\Sigma \wedge \alpha \wedge
\beta).
\]
This interesting general result proves that $\gamma \vdash_\Sigma
\varepsilon_\ell(\Sigma \wedge \alpha \wedge \beta)$.

The proof for the other two cases is similar to the ones for $\Elneu{}{}$.

\medskip

\noindent (ii) \EWCC. First we prove this property for  $\Elneu{}{}$.
Let $\gamma \equiv_\Sigma \varepsilon_\ell(\Sigma \wedge \alpha \wedge \beta)$. From \Econ, for each consistent $\alpha$, there exists $\delta$ such that $\alpha \rhd \delta$. Since $\delta \equiv_\Sigma \varepsilon(\Sigma \wedge \alpha)$, $\delta$ is unique modulo $\Sigma$. We then have $\beta \rhd \delta$. From \EWCM, we have $\alpha \wedge \beta \rhd \delta$, and since the explanation is unique modulo $\Sigma$, $\delta \equiv_\Sigma \gamma$, and $\alpha \rhd \gamma$. This is a general result: if explanations are unique, then \Econ~and \EWCM~imply \EWCC.

Now, %let us consider non-unique explanations, replacing $\equiv_\Sigma$ by $\vdash_\Sigma$.
let us examine the property for  $\Elned{}{}$.
Thus assume $\gamma \vdash_\Sigma \varepsilon_\ell(\Sigma
\wedge \alpha \wedge \beta) = \varepsilon^n(\Sigma \wedge \alpha \wedge
\beta)$. For all $\delta$ such that $ \Elned \delta\alpha$, i.e. $\delta
\vdash_\Sigma \varepsilon_\ell(\Sigma \wedge \alpha) = \varepsilon^m(\Sigma \wedge
\alpha)$, we have $\Elned \delta\beta$, i.e. $\delta \vdash_\Sigma \varepsilon_\ell(\Sigma \wedge \beta) = \varepsilon^k(\Sigma \wedge
\beta)$. Let us detail in which situations we have $\gamma\vdash_\Sigma \varepsilon^m(\Sigma \wedge \alpha)$.

First we consider the case where the erosion of the last non-empty erosion is empty.
%Let $\gamma \vdash_\Sigma \varepsilon_\ell(\Sigma
%\wedge \alpha \wedge \beta) = \varepsilon^n(\Sigma \wedge \alpha \wedge
%\beta)$. For all $\delta$ such that $\alpha \rhd \delta$, $\delta
%\vdash_\Sigma \varepsilon_\ell(\Sigma \wedge \alpha) = \varepsilon^m(\Sigma \wedge
%\alpha)$.
Since $\Sigma \wedge \alpha \wedge \beta \vdash_\Sigma
\Sigma \wedge \alpha$ we have:
\[
\varepsilon^n(\Sigma \wedge \alpha \wedge \beta) \not \vdash_\Sigma \bot \Rightarrow \varepsilon^n(\Sigma \wedge \alpha) \not \vdash_\Sigma \bot.
\]
Therefore $n \leq m$.
For the same reason, we necessarily have $n
\leq k$.

Let us first assume that $n < m$.
Since the set of preferred explanations of $\alpha$ is
included in the one of $\beta$, we have:
$\varepsilon^m(\Sigma \wedge \alpha) \vdash_\Sigma \varepsilon^k(\Sigma \wedge \beta)$.
Since $m > n$, we have:
\[
\varepsilon^m(\Sigma \wedge \alpha \wedge \beta) = \varepsilon^m(\Sigma \wedge \alpha) \wedge \varepsilon^m(\Sigma \wedge \beta) \vdash_\Sigma \bot.
\]
Let us now assume $n < k$. Then similarly, we have:
\[
\varepsilon^k(\Sigma \wedge \alpha \wedge \beta) = \varepsilon^k(\Sigma \wedge \alpha) \wedge \varepsilon^k(\Sigma \wedge \beta) \vdash_\Sigma \bot.
\]

If $k > m$, we have:
$\varepsilon^m(\Sigma \wedge \beta) \not\vdash_\Sigma \bot$,
and, since the erosion is decreasing with respect to the size of the structuring element:
$\varepsilon^k(\Sigma \wedge \beta) \vdash_\Sigma \varepsilon^m(\Sigma \wedge \beta)$.
Therefore:
$\varepsilon^m(\Sigma \wedge \alpha) \vdash_\Sigma \varepsilon^k(\Sigma \wedge \beta) \vdash_\Sigma \varepsilon^m(\Sigma \wedge \beta)$,
which implies:
$\varepsilon^m(\Sigma \wedge \alpha \wedge \beta) \not\vdash_\Sigma \bot$
which leads to a contradiction.

Similarly, if $k<m$, we have:
$\varepsilon^k(\Sigma \wedge \alpha) \not\vdash_\Sigma \bot$,
and
$\varepsilon^m(\Sigma \wedge \alpha) \vdash_\Sigma \varepsilon^k(\Sigma \wedge \alpha)$.
Therefore, since we had $\varepsilon^m(\Sigma \wedge \alpha) \vdash_\Sigma \varepsilon^k(\Sigma \wedge \beta)$, we have:
\[
\varepsilon^k(\Sigma \wedge \alpha \wedge \beta) = \varepsilon^k(\Sigma \wedge \alpha) \wedge  \varepsilon^k(\Sigma \wedge \beta) \not\vdash_\Sigma \bot
\]
which also leads to a contradiction. From these two
contradictions, we can conclude that necessarily $k = m$. Then
$\varepsilon^m(\Sigma \wedge \alpha) \vdash_\Sigma \varepsilon^k(\Sigma \wedge \beta)$
becomes $\varepsilon^m(\Sigma \wedge \alpha) \vdash_\Sigma \varepsilon^m(\Sigma \wedge
\beta)$ and therefore we have:
\[
\varepsilon^m(\Sigma \wedge \alpha \wedge \beta) = \varepsilon^m(\Sigma \wedge \alpha) \not\vdash_\Sigma \bot
\]
which is in contradiction with $n < m$. Therefore the case $n<m$ and $n<k$ is not possible.

If $n = m$. In this case, we have:
\[
\varepsilon^n(\Sigma \wedge \alpha \wedge \beta) \vdash_\Sigma \varepsilon^n(\Sigma \wedge \alpha) \wedge \varepsilon^n(\Sigma \wedge \beta) = \varepsilon^m(\Sigma \wedge \alpha) \wedge \varepsilon^m(\Sigma \wedge \beta) \vdash_\Sigma \varepsilon^m(\Sigma \wedge \alpha),
\]
and therefore:
\[
\gamma \vdash_\Sigma \varepsilon^n(\Sigma \wedge \alpha \wedge \beta) \Rightarrow \gamma \vdash_\Sigma \varepsilon^m(\Sigma \wedge \alpha),
\]
i.e. $\Elned\gamma\alpha$.
This shows that in this particular case, the property holds.

\begin{figure}[htbp]
\centerline{\hbox{
\includegraphics[height=3.5cm]{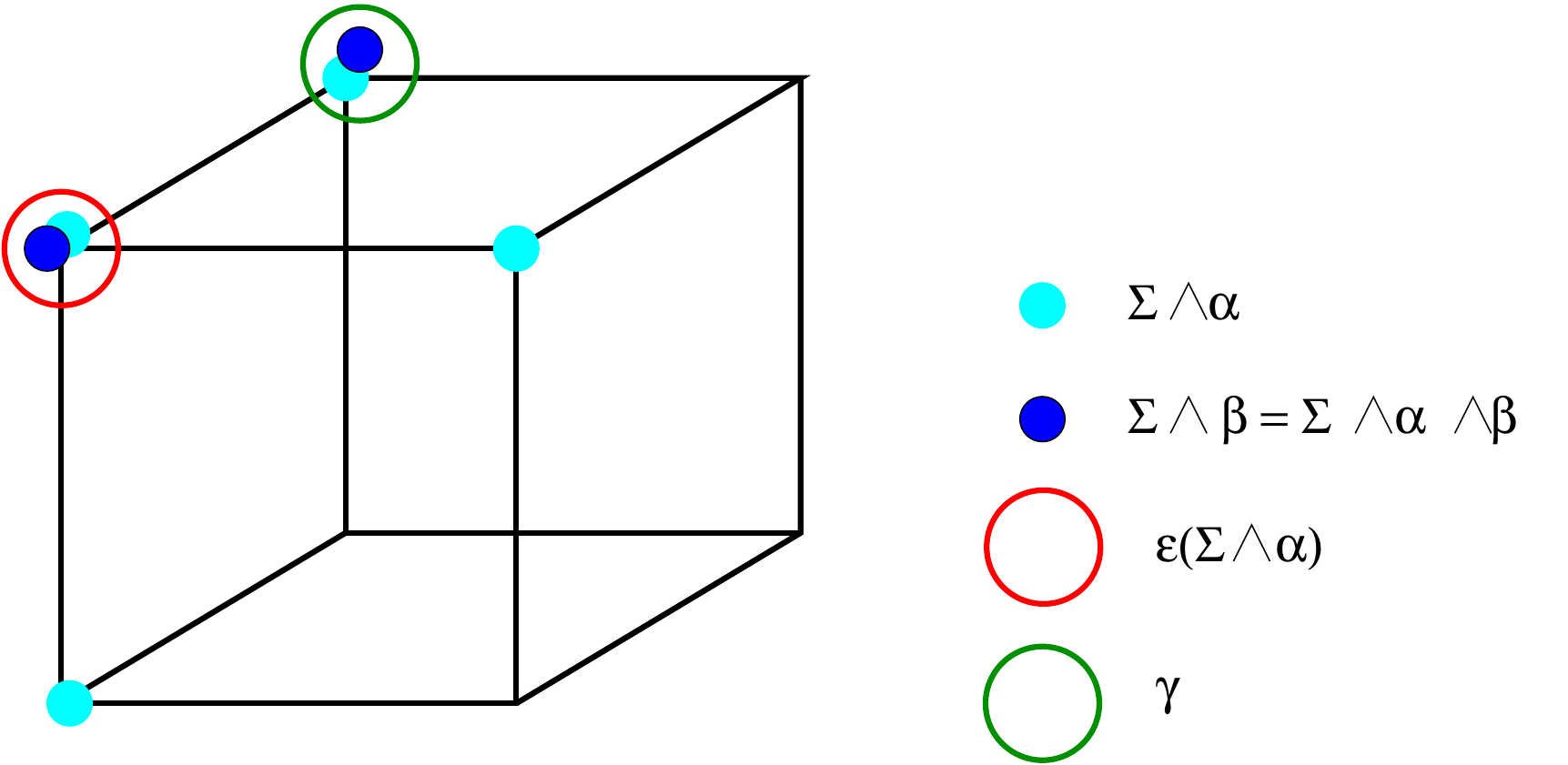}
}}
\caption{Counter-example for \EWCC\ for $\Elned{}{}$.}
\label{fig:CEx-EWCCut}
\end{figure}

Finally, in the last possibility where $n<m$ and $k=n$, the property does not hold, as shown by the following counter-example, illustrated in Figure~\ref{fig:CEx-EWCCut}: $\Sigma = \top$, $\Sigma \wedge \alpha \wedge \beta = \Sigma \wedge \beta = \varepsilon_\ell(\Sigma \wedge \alpha \wedge \beta) = \varepsilon_\ell(\Sigma \wedge \beta)$, this last erosion being obtained for $n=k=0$. For $\alpha$, $\varepsilon_\ell(\Sigma\wedge\alpha)$ is obtained for $m=1$ and has only one model. It is easy to check that for all $\delta$ such that $\delta \vdash_\Sigma \varepsilon_\ell(\Sigma \wedge \alpha)$, we have $\delta \vdash_\Sigma \varepsilon_\ell(\Sigma \wedge \beta)$. But there is a $\gamma$ such that $\gamma \vdash_\Sigma \varepsilon_\ell(\Sigma \wedge \alpha \wedge \beta)$ and $\gamma \not\vdash_\Sigma \varepsilon_\ell(\Sigma \wedge \alpha)$.

%Let us now consider again the case of non-unique explanations, but for last erosions that can be fixed points.
\medskip

Now consider the case where last erosions  can be fixed points. Actually,
several cases can occur. But before to explore the possible cases, we establish a useful claim:\\[2mm]
{\bf Claim:} Under the assumption that the premises of \EWCC\ hold, if $\varepsilon^k(\Sigma\wedge \beta)$ is a fixed point, then
$\varepsilon^m(\Sigma \wedge \alpha) \vdash_\Sigma \varepsilon^k(\Sigma \wedge \beta) \vdash_\Sigma \varepsilon^{k'}(\Sigma \wedge \beta)$ for all $k'$.\\

The reason is that we have $\varepsilon^m(\Sigma \wedge \alpha) \vdash_\Sigma \varepsilon^k(\Sigma \wedge \beta)$ by the hypothesis. And we have
$\varepsilon^k(\Sigma \wedge \beta) \vdash_\Sigma \varepsilon^{k'}(\Sigma \wedge \beta)$ for $k' < k$
because of the decreasingness of erosion with respect to $k$. Also we have
$\varepsilon^k(\Sigma \wedge \beta) \vdash_\Sigma \varepsilon^{k'}(\Sigma \wedge \beta)$ for $k\leq k'$
because of the fixed point property.\\

Now we examine the possible cases:
\begin{enumerate}
\item If the last erosion of $\Sigma \wedge \alpha \wedge \beta$ is a fixed point, i.e. $\varepsilon_\ell(\Sigma \wedge \alpha \wedge \beta)= \varepsilon^n(\Sigma \wedge \alpha \wedge \beta) = \varepsilon^{n'}(\Sigma \wedge \alpha \wedge \beta)$ for all $n'\geq n$. This implies that $\varepsilon^{n'}(\Sigma \wedge \alpha) \wedge \varepsilon^{n'} (\Sigma \wedge \beta)$ can never be inconsistent (for all $n'$). Hence the last erosions of $\Sigma \wedge \alpha$ and $\Sigma \wedge \beta$ have to be fixed points too. Let us denote by $m$ and $k$ the first size of erosions where these fixed points are reached.
    %Since for each $\delta$ being an explanation of $\alpha$, $\delta$ is also an explanation of $\beta$, we have $\varepsilon^m(\Sigma \wedge \alpha) \vdash_\Sigma \varepsilon^k(\Sigma \wedge \beta) \vdash_\Sigma \varepsilon^{k'}(\Sigma \wedge \beta)$ for all $k'$ (for $k' < k$ this comes from the decreasingness of erosion with respect to $k$, and for $k' \geq k$ from the fixed point property).
    By the Claim,
    $\varepsilon^m(\Sigma \wedge \alpha) \vdash_\Sigma \varepsilon^k(\Sigma \wedge \beta) \vdash_\Sigma \varepsilon^{k'}(\Sigma \wedge \beta)$ for all $k'$.
    If $n\geq m$ we have $\varepsilon^n(\Sigma \wedge \alpha \wedge \beta) =  \varepsilon^n(\Sigma \wedge \alpha) \wedge \varepsilon^n(\Sigma \wedge \beta) = \varepsilon^n(\Sigma \wedge \alpha) = \varepsilon^m(\Sigma \wedge \alpha)$, and $\gamma \vdash_\Sigma \varepsilon^m(\Sigma \wedge \alpha)$. If $n<m$, then similarly $\varepsilon^n(\Sigma \wedge \alpha \wedge \beta) = \varepsilon^m(\Sigma \wedge \alpha \wedge \beta) = \varepsilon^m(\Sigma \wedge \alpha)$ and $\gamma \vdash_\Sigma \varepsilon^m(\Sigma \wedge \alpha)$.

\item If the last erosion of $\Sigma \wedge \alpha$ is a fixed point. Then, $\varepsilon^m(\Sigma \wedge \alpha) \vdash_\Sigma \varepsilon^k(\Sigma \wedge \beta)$ implies that the last erosion of $\Sigma \wedge \beta$ is a fixed point too. By the Claim, $\varepsilon^m(\Sigma \wedge \alpha) \vdash_\Sigma \varepsilon^{k'}(\Sigma \wedge \beta)$ for all $k'$. This means that $\varepsilon^{n+1}(\Sigma \wedge \alpha \wedge \beta) = \varepsilon^{n+1}(\Sigma \wedge \alpha) \wedge \varepsilon^{n+1}(\Sigma \wedge \beta)$ can never be inconsistent, and the last erosion of $\Sigma \wedge \alpha \wedge \beta$ is a fixed point too. Hence this case is equivalent to the first one.

\item If the last erosion of $\Sigma \wedge \beta$ is a fixed point, and $\varepsilon^{m+1}(\Sigma \wedge \alpha) = \bot$. Then $\varepsilon^{m+1}(\Sigma \wedge \alpha\wedge \beta) = \bot$, which implies $n \leq m$ and $\varepsilon^{n+1}(\Sigma \wedge \alpha\wedge \beta) = \bot$. If $n<m$, then, by the Claim, $\varepsilon^m(\Sigma \wedge \alpha) \vdash_\Sigma \varepsilon^{n+1}(\Sigma \wedge \alpha) \wedge \varepsilon^{n+1}(\Sigma \wedge \beta) = \varepsilon^{n+1}(\Sigma \wedge \alpha\wedge \beta)$ which can therefore not be inconsistent. Hence $n=m$. Then we have $\varepsilon^n(\Sigma \wedge \alpha \wedge \beta) =  \varepsilon^m(\Sigma \wedge \alpha) \wedge \varepsilon^m(\Sigma \wedge \beta) = \varepsilon^m(\Sigma \wedge \alpha)$, and $\gamma \vdash_\Sigma \varepsilon^m(\Sigma \wedge \alpha)$.
\end{enumerate}

\medskip

\noindent (iii) \ERef. The definition of $\Elneu{}{}$ is based on
the notion of largest possible erosion, and therefore no further
erosion can be performed. More precisely, let $\Elneu\gamma\alpha$
and suppose that the last non empty erosion of $\Sigma\wedge
\alpha$ is $\varepsilon^n(\Sigma \wedge \alpha)$. Then we have $\gamma \equiv_\Sigma \varepsilon^n(\Sigma \wedge \alpha)$. Let us now consider two cases:
\begin{enumerate}
\item If $\varepsilon^{n+1}(\Sigma \wedge \alpha) = \bot$, then
$\varepsilon^0(\Sigma \wedge \gamma) = \Sigma \wedge \gamma$
and $\varepsilon^1(\Sigma \wedge \gamma) = \varepsilon^{n+1}(\Sigma \wedge \alpha) = \bot$.
Therefore $\varepsilon_\ell(\Sigma \wedge \gamma) = \Sigma \wedge \gamma$ and $\gamma \equiv_\Sigma \varepsilon_\ell(\Sigma \wedge \gamma)$. Hence $\Elneu{\gamma}{\gamma}$.
\item If $\varepsilon^{n+1}(\Sigma \wedge \alpha) = \varepsilon^n(\Sigma \wedge \alpha)$ (fixed point). Then $\varepsilon^0(\Sigma \wedge \gamma) = \varepsilon^n(\Sigma \wedge \alpha) = \Sigma \wedge \gamma$
and $\varepsilon^1(\Sigma \wedge \gamma) = \varepsilon^{n+1}(\Sigma \wedge \alpha) = \varepsilon^n(\Sigma \wedge \alpha) = \Sigma \wedge \gamma$, which is a fixed point of the erosions. Therefore $\varepsilon_\ell(\Sigma \wedge \gamma) = \Sigma \wedge \gamma$ and $\gamma \equiv_\Sigma \varepsilon_\ell(\Sigma \wedge \gamma)$. Hence $\Elneu{\gamma}{\gamma}$.
\end{enumerate}

Now, if we consider $\Elned{}{}$, the same reasoning applies in the first case (when the successive erosions end up with $\bot$).
%Note that if we define the explanations as $\gamma \vdash_\Sigma \varepsilon_\ell(\Sigma \wedge \gamma)$ (replacing $\equiv_\Sigma$ by $\vdash_\Sigma$), the same reasoning applies in the first case (when the successive erosions end up with $\bot$).
However it does not apply in the case of non-empty fixed point. Let us for instance consider erosions performed with $B^{ab}$, as in Example~\ref{ejemplo}, and let us assume that $\varepsilon_\ell(\Sigma \wedge \alpha)=c$. Let us take $\gamma = (\neg a \wedge b \wedge c) \vee (a \wedge \neg b \wedge c) \vee (a\wedge b\wedge c)$ as an explanation of $\alpha$ (we have $\gamma \vdash_\Sigma \varepsilon_\ell(\Sigma \wedge \alpha)$). Then $\varepsilon^1(\Sigma \wedge \gamma) = a \wedge b \wedge c = \varepsilon_\ell(\Sigma \wedge \gamma)$ (still with $B^{ab}$ as structuring element). However $\gamma \not\vdash_\Sigma a \wedge b \wedge c $ and therefore $\gamma$ is not an explanation of $\gamma$ in this case.

\endproof

%%%%%%%%%%%%%%%%%%%%%%%%%%%%%%%%%%%%%%%%%%%%%%%%%%%%%%%%%%%

\bibliographystyle{plain}
\bibliography{biblio}

\begin{thebibliography}{10}
%\expandafter\ifx\csname url\endcsname\relax
%  \def\url#1{\texttt{#1}}\fi
%\expandafter\ifx\csname urlprefix\endcsname\relax\def\urlprefix{URL }\fi
%
\bibitem{ARXIV}
M.~Aiguier, J.~{Atif}, I.~{Bloch}, C.~Hudelot, Belief revision, minimal change
  and relaxation: {Part I - A} general framework based on the theory of
  institutions, CoRR abs/1502.02298.

\bibitem{AABH15}
M.~Aiguier, J.~Atif, I.~Bloch, C.~Hudelot, Belief revision, minimal change and
  relaxation: Part {II} - {I}nstantiation in multiple description logics, arXiv
  CoRR CoRR abs/1502.07628.

\bibitem{MA:AIJ-17}
M.~{Aiguier}, J.~{Atif}, I.~{Bloch}, C.~{Hudelot}, Belief revision, minimal
  change and relaxation: A general framework based on satisfaction systems, and
  applications to description logics, Artificial Intelligence to appear.

\bibitem{AGM-85}
C.~E. Alchourr\'on, P.~G\"ardenfors, D.~Makinson, On the {L}ogic of {T}heory
  {C}hange: {P}artial {M}eet {C}ontraction and {R}evision {F}unctions, Journal
  of Symbolic Logic 50 (1985) 510--530.

\bibitem{bienvenu2008}
M.~Bienvenu, Complexity of abduction in the el family of lightweight
  description logics., in: International Conference on Principles of Knowledge
  Representation and Reasoning (KR), 2008.

\bibitem{JANCL-02}
I.~Bloch, Modal {L}ogics based on {M}athematical {M}orphology for {S}patial
  {R}easoning, Journal of Applied Non Classical Logics 12~(3-4) (2002)
  399--424.

\bibitem{IB:INS-11}
I.~{Bloch}, Lattices of fuzzy sets and bipolar fuzzy sets, and mathematical
  morphology, Information Sciences 181 (2011) 2002--2015.

\bibitem{IB:CVIU-13}
I.~{Bloch}, A.~{Bretto}, Mathematical morphology on hypergraphs, application to
  similarity and positive kernel, Computer Vision and Image Understanding
  117~(4) (2013) 342--354.

\bibitem{IB:LOS-07}
I.~Bloch, H.~Heijmans, C.~Ronse, Mathematical {M}orphology, in: M.~Aiello,
  I.~Pratt-Hartman, J.~van Benthem (eds.), Handbook of Spatial Logics,
  chap.~13, Springer, 2007, pp. 857--947.

\bibitem{IPMU-00b}
I.~Bloch, J.~Lang, Towards {M}athematical {M}orpho-{L}ogics, in: 8th
  International Conference on Information Processing and Management of
  Uncertainty in Knowledge based Systems IPMU 2000, vol. III, Madrid, Spain,
  2000.

\bibitem{TCIS-02}
I.~Bloch, J.~Lang, Towards {M}athematical {M}orpho-{L}ogics, in:
  B.~Bouchon-Meunier, J.~Gutierrez-Rios, L.~Magdalena, R.~Yager (eds.),
  Technologies for Constructing Intelligent Systems, Springer, 2002, pp.
  367--380.

\bibitem{ECSQARU-01}
I.~Bloch, R.~{Pino P\'erez}, C.~Uzc\'ategui, Explanatory {R}elations based on
  {M}athematical {M}orphology, in: ECSQARU 2001, Toulouse, France, 2001.

\bibitem{BFKP12}
R.~Booth, E.~Ferm{\'e}, S.~Konieczny, R.~{Pino P{\'e}rez}, Credibility-limited
  revision operators in propositional logic, in: 13th International Conference
  on Principles of Knowledge Representation and Reasoning (KR 2012), 2012.

\bibitem{DALA-88}
M.~Dalal, Investigations into a {T}heory of {K}nowledge {B}ase {R}evision:
  {P}reliminary {R}eport, in: AAAI'88, 1988.

\bibitem{DP11}
J.-P. Delgrande, P.~Peppas, Revising {H}orn theories, in: T.~Walsh (ed.), 22nd
  International Joint Conference on Artificial Intelligence (IJCAI),
  IJCAI/AAAI, 2011.

\bibitem{DP15}
J.-P. Delgrande, P.~Peppas, Belief revision in {H}orn theories, Artificial
  Intelligence 218 (2015) 1--22.

\bibitem{eiter1995}
T.~Eiter, G.~Gottlob, The complexity of logic-based abduction, Journal of the
  ACM 42~(1) (1995) 3--42.

\bibitem{Flach96}
P.~A. Flach, Rationality {P}ostulates for {I}nduction, in: Y.~Shoham (ed.),
  Sixth Conference of Theoretical Aspects of Rationality and Knowledge
  (TARK96), The Netherlands, 1996.

\bibitem{Ginsberg98}
M.~L. Ginsberg, A.~J. Parkes, A.~Roy, Supermodels and {R}obustness, in:
  Fifteenth National Conference on Artificial Intelligence AAAI'98, Madison,
  Wisconsin, 1998.

\bibitem{Gorogiannis2008b}
N.~Gorogiannis, A.~Hunter, {Implementing semantic merging operators using
  binary decision diagrams}, International Journal of Approximate Reasoning
  49~(1) (2008) 234--251.

\bibitem{Gorogiannis2008a}
N.~Gorogiannis, A.~Hunter, {Merging First-Order Knowledge using Dilation
  Operators}, in: Fifth International Symposium on Foundations of Information
  and Knowledge Systems, FoIKS'08, vol. LNCS 4932, 2008.

\bibitem{halland2012}
K.~Halland, K.~Britz, {ABox} abduction in {ALC} using a {DL} tableau, in: ACM
  South African Institute for Computer Scientists and Information Technologists
  Conference, 2012.

\bibitem{HFCF01}
S.~O. Hansson, E.~Ferm\'e, J.~Cantwell, M.~Falappa, Credibility limited
  revision, Journal of Symbolic Logic 66 (2001) 1581--1596.

\bibitem{Heijmans94}
H.~J. A.~M. Heijmans, Morphological {I}mage {O}perators, Academic Press,
  Boston, 1994.

\bibitem{HEIJ-90}
H.~J. A.~M. Heijmans, C.~Ronse, The {A}lgebraic {B}asis of {M}athematical
  {M}orphology -- {P}art~{I}: {D}ilations and {E}rosions, Computer Vision,
  Graphics and Image Processing 50 (1990) 245--295.

\bibitem{HughesCreswell68}
G.~E. Hughes, M.~J. Cresswell, An {I}ntroduction to {M}odal {L}ogic, Methuen,
  London, UK, 1968.

\bibitem{KatsunoMendelzon91}
H.~Katsuno, A.~O. Mendelzon, Propositional {K}owledge {B}ase {R}evision and
  {M}inimal {C}hange, Artificial Intelligence 52 (1991) 263--294.

\bibitem{keshet2000}
R.~Keshet, {Mathematical Morphology on Complete Semilattices and its
  Applications to Image Processing}, Fundamenta Informaticae 41 (2000) 33--56.

\bibitem{KMP10}
S.~Konieczny, M.~{Medina Grespan}, R.~{Pino P\'rez}, Taxonomy of improvement
  operators and the problem of minimal change, in: Proceedings of the 12th
  International Conference on Principles of Knowledge Representation and
  Reasoning (KR 2010), 2010.

\bibitem{KONI-98}
S.~Konieczny, R.~{Pino P\'erez}, On the {L}ogic of {M}erging, in: 6th
  International Conference on Principles of Knowledge Representation and
  Reasoning, Trento, Italy, 1998.

\bibitem{KONI-02}
S.~Konieczny, R.~{Pino P\'erez}, Merging {I}nformation: {A} {Q}ualitative
  {F}ramework, Journal of Logic and Computation 12~(5) (2002) 773--808.

\bibitem{KP11}
S.~Konieczny, R.~{Pino P{\'e}rez}, Logic based merging, Journal of
  Philosophical Logic 40~(2) (2011) 239--270.

\bibitem{LAFA-00b}
C.~Lafage, J.~Lang, Logical {R}epresentation of {P}references for {G}roup
  {D}ecision {M}aking, in: A.~G. Cohn, F.~Giunchiglia, B.~Selman (eds.), 7th
  International Conference on Principles of Knowledge Representation and
  Reasoning KR 2000, Morgan Kaufmann, San Francisco, CA, Breckenridge, CO,
  2000.

\bibitem{MATH-67}
G.~Matheron, El\'ements pour une th\'eorie des milieux poreux, Masson, Paris,
  1967.

\bibitem{MATH-75}
G.~Matheron, Random {S}ets and {I}ntegral {G}eometry, Wiley, New-York, 1975.

\bibitem{NajTal10}
L.~Najman, H.~Talbot (eds.), {Mathematical morphology: from theory to
  applications}, ISTE-Wiley, 2010.

\bibitem{PINO-99}
R.~{Pino P\'erez}, C.~Uzc\'ategui, Jumping to {E}xplanations versus jumping to
  {C}onclusions, Artificial Intelligence 111 (1999) 131--169.

\bibitem{PU03}
R.~{Pino P{\'e}rez}, C.~Uzc{\'a}tegui, Preferences and explanations, Artificial
  Intelligence 149~(1) (2003) 1--30.

\bibitem{QLB06}
G.~Qi, W.~Liu, D.-A. Bell, Knowledge base revision in description logics, in:
  M.~Fisher, W.~V. der Hoek, B.~Konev, A.~Lisitsa (eds.), European Conference
  on Logics in Artificial Intelligence (JELIA), vol. LNCS 4160,
  Springer-Verlag, 2006.

\bibitem{QY08}
G.~Qi, F.~Yang, A survey of revision approaches in description logics, in:
  D.~Calvanese, G.~Lausen (eds.), Web Reasoning and Rule Systems (RR), Second
  International Conference, vol. LNCS 5341, Springer-Verlag, 2008.

\bibitem{RW09}
M.-M. Ribeiro, R.~Wassermann, {AGM} revision in description logics, in: First
  {W}orkshop on {A}utomated {R}easoning about {C}ontext and {O}ntology
  {E}volution (ARCOE), 2009.

\bibitem{RW10}
M.-M. Ribeiro, R.~Wassermann, More about {AGM} revision in description logics,
  in: Second {W}orkshop on {A}utomated {R}easoning about {C}ontext and
  {O}ntology {E}volution (ARCOE), 2010.

\bibitem{RW14}
M.-M. Ribeiro, R.~Wassermann, Minimal change in {AGM} for non-classical logics,
  in: C.~Baral, G.~D. Giacomo, T.~Eiter (eds.), Fourteenth International
  Conference on Principles of Knowledge Representation and Reasoning (KR),
  {AAAI} Press, 2014.

\bibitem{RWFA13}
M.-M. Ribeiro, R.~Wassermann, G.~Flouris, G.~Antoniou, Minimal change:
  {R}elevance and recovery revisited, {A}rtificial {I}ntelligence 201 (2013)
  59--80.

\bibitem{Rons90}
C.~Ronse, Why {M}athematical {M}orphology {N}eeds {C}omplete {L}attices, Signal
  Processing 21~(2) (1990) 129--154.

\bibitem{RonsHeij:91}
C.~Ronse, H.~J. A.~M. Heijmans, The {A}lgebraic {B}asis of {M}athematical
  {M}orphology -- {P}art {II}: {O}penings and {C}losings, Computer Vision,
  Graphics and Image Processing 54 (1991) 74--97.

\bibitem{RYAN-91}
M.~D. Ryan, Belief {R}evision and {O}rdered {T}heory {P}resentations, in:
  A.~Fuhrmann, H.~Rott (eds.), Logic, Action and Information. Also in Eighth
  Amsterdam Colloquium on Logic, 1991, De Gruyter, 1994, pp. 129--151.

\bibitem{SERR-82}
J.~Serra, Image Analysis and Mathematical Morphology, Academic Press, London,
  1982.

\bibitem{SERR-88}
J.~Serra, Image {A}nalysis and {M}athematical {M}orphology, {P}art {II}:
  {T}heoretical {A}dvances, Academic Press (J. Serra Ed.), London, 1988.

\bibitem{TK83}
A.~Tversky, D.~Kahneman, Extension versus intuitive reasoning: The conjunction
  fallacy in probability judgment, Psychological Review 90~(4) (1983) 293--315.

\bibitem{WWT10}
Z.~Wang, K.~Wang, R.-W. Topor, Revising general knowledge bases in description
  logics, in: F.~Lin, U.~Sattler, M.~Truszczynski (eds.), Twelfth International
  Conference on Principles of Knowledge Representation and Reasoning (KR),
  {AAAI} Press, 2010.

\bibitem{ZPZ13}
Z.-Q. Zhuang, M.~Pagnucco, Y.~Zhang, Definability of {H}orn revision from
  {H}orn contraction, in: 23rd International Joint Conference on Artificial
  Intelligence (IJCAI), IJCAI/AAAI, 2013.

\end{thebibliography}

\end{document}